\documentclass[11pt,a4paper]{article}

\usepackage[margin=1in]{geometry}
\usepackage{microtype}
\usepackage{graphicx}
\usepackage{subcaption}
\usepackage{booktabs}
\usepackage{amsmath}
\usepackage{amssymb}
\usepackage{mathtools}
\usepackage{amsthm}
\usepackage{fvextra}
\usepackage{hyperref}
\usepackage[capitalize,noabbrev]{cleveref}
\usepackage{placeins}

\DeclareMathOperator*{\argmax}{\arg\!\max}

\DefineVerbatimEnvironment{ChatLog}{Verbatim}{
  breaklines=true,
  fontsize=\small
}

\title{SEISMO: LLMs can be sample-efficient molecular optimisers}

\author{%
  Fabian P. Krüger\textsuperscript{1,2,3,*} \and
  Andrea Hunklinger\textsuperscript{3,4,*} \and
  Adrian Wolny\textsuperscript{3} \and
  Tim J.\ Adler\textsuperscript{3} \and
  Igor V. Tetko\textsuperscript{2,5} \and
  Santiago David Villalba\textsuperscript{3,\dag}%
}

\date{}

\begin{document}
\maketitle

\noindent
\textsuperscript{1}Technical University of Munich, TUM School of Computation, Information and Technology, Department of Mathematics, Germany.\\
\textsuperscript{2}Helmholtz Munich -- German Research Center for Environmental Health (GmbH), Institute of Structural Biology, Molecular Targets and Therapeutics Center, 85764 Neuherberg, Germany.\\
\textsuperscript{3}Bayer AG, Machine Learning Research.\\
\textsuperscript{4}Universitat de Barcelona, Facultat de Farm\`acia I Ci\`encies de l'Alimentaci\'o, Avda. Diagonal 643, 08028 Barcelona, Spain.\\
\textsuperscript{5}BIGCHEM GmbH, Valerystr.\ 49, 85716 Unterschlei\ss{}heim, Germany\\ [0.3em]
\textsuperscript{*}Equal contribution. \quad
\textsuperscript{\dag}Corresponding authors:
\href{mailto:fabian.krueger@tum.de}{fabian.krueger@tum.de},
\href{mailto:andi.hunklinger@bayer.com}{andi.hunklinger@bayer.com},
\href{mailto:santiago.villalba@bayer.com}{santiago.villalba@bayer.com}.

\bigskip

%%%%%%%%%%%%%%%%%%%%%%%%%%%%%%%%%%%%%%%%%%%%%%%%%%%%%%%%%%%%%%%%%%%%%%%
\begin{abstract}

Optimising molecules to achieve desired properties is a central bottleneck across the chemical sciences, particularly in the pharmaceutical industry, where it underlies the discovery of new drugs. Since molecular property evaluation often relies on costly and rate-limited oracles, such as experimental assays, molecular optimisation must be highly sample-efficient. To address this, we introduce SEISMO, an LLM agent for inference-time molecular optimisation that turns information routinely available alongside the oracle score, but discarded by existing methods, into an explicit guidance signal. By conditioning each proposal on the full optimisation trajectory and on explanatory feedback derived from post-hoc explainability methods, we consistently increase sample efficiency across a wide range of drug-discovery-relevant tasks compared to existing optimisers as well as zero-shot LLM generation, with gains growing as explanatory feedback is enriched. In practice, medicinal chemists can inspect the agent’s reasoning and intervene to steer generation in natural language, keeping them central to molecular optimisation projects.

\medskip
\noindent\textbf{Scientific Contribution.}
We introduce SEISMO, a trajectory-conditioned LLM agent for molecular optimisation that, unlike prior work, requires neither populations or batches, nor fine-tuning. Rather than treating the oracle as a scalar black box, SEISMO incorporates context such as sub-score decompositions, feature attributions, and structural-similarity differences as a novel machine-readable learning signal. We show that this setup substantially improves sample efficiency over existing methods on a range of drug-discovery tasks.

\end{abstract}


\newpage
%%%%%%%%%%%%%%%%%%%%%%%%%%%%%%%%%%%%%%%%%%%%%%%%%%%%%%%%%%%%%%%%%%%%%%%
\section{Introduction}
\label{sec:intro}

Generating molecules with desired properties is a central goal in chemistry. A prominent example is the pharmaceutical industry, which requires the discovery of novel molecules in order to bring new drugs to the market. Such molecules must satisfy multiple objectives, including potency, selectivity, and a range of pharmacokinetic and physicochemical properties \cite{lipinski}, which can conflict and therefore require explicit trade-offs. At the same time, the searchable chemical space is enormous, often estimated at around 10$^{60}$ plausible small drug-like molecules \cite{chem_space}.

Because this chemical space is far too large to be explored exhaustively, practical molecular design is instead carried out as an iterative optimization process in which candidate molecules are proposed, evaluated, and refined over successive steps \cite{dmta}. For the evaluations, experimental assays provide the most reliable assessment of molecular properties, but their high cost and limited throughput make them impractical for use at every iteration. As a result, molecular optimization pipelines rely on computational methods that approximate experimental outcomes \cite{dmta}. In practice, these methods span a spectrum of cost and fidelity: lower-cost approximations often provide weaker correlations with experimental measurements, while more accurate approaches, such as physics-based simulations and protein–ligand co-folding methods, are computationally expensive and rate-limited \cite{tradeoff,passaro2025boltz2,abramson2024accurate}. Reducing the number of required evaluations therefore has direct practical value: sample-efficient optimization methods enable the use of higher-quality evaluations, which in turn leads to molecules that are more likely to perform well in lab experimentation. 
%\paragraph{Classical molecular optimisation.}

Computational molecular optimisation is dominated by three broad approaches: reinforcement-learning-based optimisation of generative models, genetic algorithms, and Bayesian optimisation \cite{pmo_benchmark}. Examples of each category are REINVENT \cite{reinvent}, Graph-GA \cite{graphga}, and Gaussian-process Bayesian optimisation (GP BO) \cite{gpbo}. REINVENT trains a generative model against a reward signal, Graph-GA evolves a population through graph-based crossover and mutation, and GP BO fits a surrogate over molecular fingerprints and selects candidates with an acquisition function. Within the reinforcement-learning family, subsequent work has recently improved sample efficiency further through experience replay \cite{augmented} and substructure-based self-conditioning \cite{beam_enumeration}. Even so, optimisation in these settings still typically requires a large number of oracle calls and therefore assumes relatively cheap, scalar-valued signal, rather than rich, structured feedback that could guide search more directly.

\begin{figure}[ht]
  \centering
  \includegraphics[width=0.7\linewidth]{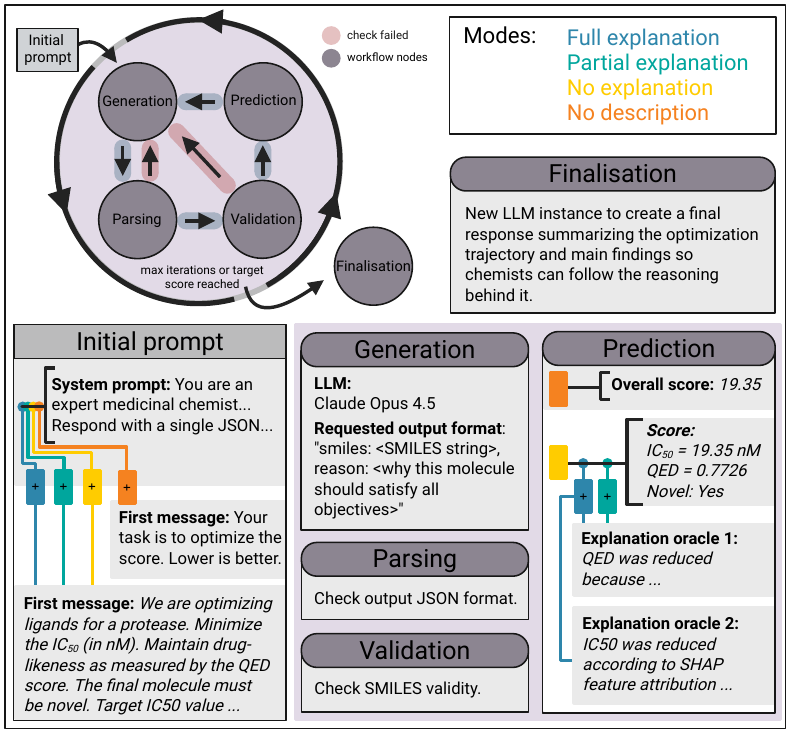}
  \caption{Overview of the iterative optimisation cycle of SEISMO. The initial message and prediction feedback are modular and vary between tasks. Colours indicate the four information levels that we ablate in our experimental setup for the LLM agent (no description, no explanation, partial explanation, full explanation). The full system prompt and an example conversation are provided in Appendix~\ref{sec:appendix:examples}.}
  \label{fig:workflow}
\end{figure}

%\paragraph{LLMs in optimisation loops.}
A more recent line of work embeds large language models (LLMs) directly into these optimisation loops. MolLEO \cite{molleo} uses an LLM as a mutation and crossover operator inside a genetic algorithm. LICO \cite{lico} uses an LLM as a surrogate that scores candidates produced by an evolutionary loop. More generally, LLM-as-optimiser frameworks such as OPRO \cite{opro} and BOPRO \cite{bopro} iterate prompts conditioned on a scalar objective. In each of these designs, however, the LLM either operates on a single scalar feedback channel or is invoked within a surrogate-driven outer loop that does not condition on the full optimisation trajectory.

%\paragraph{LLM agents in chemistry.}
Beyond these optimisation methods, a separate line of work develops LLM agents for broader chemistry workflows. These agents address tasks such as chemistry question answering (ChemAgent \cite{chemagent}, CACTUS \cite{cactus}), drug--target interaction prediction (DrugAgent \cite{drugagent}), and tool-augmented retrieval, synthesis planning, and end-to-end medicinal-chemistry assistance (ChemCrow \cite{chemcrow}, Frogent \cite{frogent}, MADD \cite{madd}). Across these systems, the LLM acts as a controller orchestrating external tools, with optimisation delegated to learned models or heuristics. Yet these systems also show that LLMs can readily ingest heterogeneous chemical information, which is precisely the kind of side-information that conventional optimisers leave unexploited.

%\paragraph{From black-box to grey-box optimisation.}
Most existing molecular optimisers treat the evaluation function as a scalar black box. In practice this function is rarely strictly black box: it is typically a weighted aggregate of multiple property scores, and substantial chemical and biological prior knowledge is available about the individual components. When these scores are obtained from specialised in-silico predictive models, explainable AI (XAI) methods such as Shapley Additive Explanations (SHAP) \cite{shap} can provide feature attributions for individual predictions. These attributions can serve as an additional feedback signal by making feature level patterns learned by specialised models accessible to the optimiser. Framing molecular optimisation as a \emph{grey-box} problem makes this side-information explicit and motivates optimisers that can consume it directly. We argue that an LLM agent is a natural way to operationalise grey-box molecular optimisation: natural-language conditioning makes it straightforward to pass task descriptions, sub-score breakdowns, and structured explanations to the generator, and the model can interpret them against the rich chemical prior knowledge it already encodes.

In contrast to previous methods, our optimiser is an LLM that reasons over the full optimisation trajectory. Rather than relying on many cheap scalar evaluations, we treat each oracle call as an informative, potentially expensive observation, and propose candidates sequentially, revising after every evaluation. This setup is best suited to an information-rich, low-oracle-budget regime, as in in silico molecular optimisation, but the same design could also support a best-of-N parallel sampling strategy for batch-oriented wet-lab evaluation. Specifically, our contributions are: 

\begin{itemize}
  \item We introduce SEISMO (Sample-Efficient
Inference-Stage Molecular-Optimization agent), an online, trajectory-conditioned LLM optimiser for molecular design that incorporates feedback beyond scalars.
  \item We show that post-hoc explainability such as sub-score decomposition, SHAP attributions, and structural-similarity diffs can serve as a machine-readable feedback signal within an agentic optimisation loop, rather than used solely for human interpretation. To our knowledge, the use of feature attribution methods as an optimisation signal is new in cheminformatics.
  \item We validate the approach across (i) medicinal-chemistry hit-identification and lead-optimisation proxies, with comparisons to classical baselines and zero-shot LLM generation; (ii) a co-folding-based oracle on a novel protein with no known small-molecule binders; and (iii) the Practical Molecular Optimization (PMO) benchmark.
  \item We provide a cost-benefit analysis that identifies the oracle-cost regime in which SEISMO is economically advantageous over classical baselines.
\end{itemize}

%%%%%%%%%%%%%%%%%%%%%%%%%%%%%%%%%%%%%%%%%%%%%%%%%%%%%%%%%%%%%%%%%%%%%%%
\section{Methods}
\label{sec:methods}

\subsection{Problem formulation}
\label{sec:methods:problem}

The molecular optimisation problem is the task of finding a molecule $m \in M$ that maximises an evaluation function $F: M \to \mathbb{R}$. Formally,
\begin{equation*}
  m^* \in \argmax_{m \in M} F(m).
\end{equation*}
Although $F$ is typically treated as a black box, this abstraction discards substantial additional information that is typically available in real world settings.

First, $F(m)$ usually encompasses several desirable properties, such as binding affinities, ADMET properties, and synthesisability, that are aggregated into a scalar score
\begin{equation*}
  F(m) = \sum_{k=1}^{K} w_k\, f_k(m),
\end{equation*}
where the individual components $f_k$ and their weights $w_k$ are predefined.

Second, for tasks of direct scientific or industrial relevance - such as optimising drug-like properties or binding affinity to a target protein - there usually exists substantial prior knowledge about the different properties $f_k$, accumulated through decades of chemical and biological research. This knowledge includes, for example, well-characterized functional-group effects, structural alerts and other medicinal-chemistry heuristics, synthetic feasibility considerations, and established structure--property relationships. Such domain knowledge can inform and constrain the search over candidate molecules even when $F$ itself is only accessible via queries.

Third, because most property scores are derived from computational surrogate predictors, these predictors often expose additional information beyond the scalar score itself that provides insight into why a candidate molecule receives a particular score, thereby informing subsequent optimization steps further. This includes, for instance, post hoc feature attribution methods such as Shapley Additive Explanations\cite{shap}, as well as auxiliary outputs attached to the oracle itself, such as 3D co-folding structures or uncertainty estimates that accompany an affinity prediction. We model this via an explainer $G$ and a natural-language verbaliser,
\begin{equation*}
  G:\mathcal{F}\times M \to \mathcal{E},
  \qquad
  \text{verbalize}:\mathcal{E}\to \Sigma_{\mathrm{NL}}^{*},
\end{equation*}
where $\mathcal{F}$ denotes the space of evaluation functions ($F\in\mathcal{F}$). Our goal is to improve sample efficiency by leveraging all these sources of information and assessing their individual contributions.

\subsection{SEISMO agent}
\label{sec:methods:agent}

A natural way to incorporate the aforementioned priors and side-information is through LLMs. As LLMs are trained on large, heterogeneous text corpora that include substantial scientific and technical content, they can serve as a practical proxy for the kind of accumulated domain knowledge described above.
While such knowledge is not guaranteed to be perfectly accurate or complete, it provides an inductive bias that can guide the search toward plausible, high-scoring regions of chemical space.
In addition, modern LLMs are capable of reasoning over the provided information before committing to an answer~\cite{deepseekr1, openai_o1}.

Crucially, LLMs operate on natural language, which makes it straightforward to provide rich, structured feedback beyond a scalar objective. In particular, we can supply the agent with the task description, the sub-score decomposition ${f_k(m)}_{k=1}^{K}$, and extra explanations produced by $G$. This turns each molecule evaluation into an informative observation rather than a single number.

Motivated by this, we construct SEISMO as a goal-directed LLM agent that performs inference-time search through iterative interaction with an evaluation environment. Neither the LLM nor the scoring models are updated during optimisation. Instead, SEISMO uses the feedback obtained during inference to guide subsequent molecular proposals.

Let $d \in \Sigma_{\mathrm{NL}}^{*}$ denote the initial natural-language problem description (task context and any available prior constraints). At iteration $t \in \{1,\dots,T\}$, the agent proposes a molecule based on the context and the history of previous trials:
\begin{equation*}
    m_t \sim \pi_{\mathrm{LLM}}(\,\cdot \mid d, H_{t-1}),
\end{equation*}
where $H_{t-1}$ is the history (see below) and $\pi_{\mathrm{LLM}}$ is the proposal policy obtained by sampling from an LLM with fixed weights and mapping the generated strings to molecules (via SMILES parsing), thereby inducing a conditional distribution over $M$ given $(d, H_{t-1})$. The environment then returns not only the aggregate score but also its decomposition and explanations,
\begin{equation*}
    y_t = \Big(F(m_t),\, (f_k(m_t))_{k=1}^{K},\, (r_{t,k})_{k=1}^{K}\Big),
\end{equation*}

where $r_{t,k} \in \Sigma_{\mathrm{NL}}^{*}$ is a textual explanation for the component score $f_k(m_t)$. The history is updated by appending the new trial and feedback,
\begin{equation*}
    H_t = H_{t-1} \cup \{(m_t, y_t)\}.
\end{equation*}
The update from $H_{t-1}$ to $H_t$ after every evaluation induces an online optimisation process with update frequency one per sample, rather than one per batch. This removes the lower bound on sample efficiency imposed by population-based or batched acquisition schemes used in many existing methods.
After a budget of $T$ evaluations, the agent outputs the best molecule encountered,
\begin{equation*}
    \hat m_T \in \argmax_{t \in \{1,\dots,T\}} F(m_t).
\end{equation*}
In this setup, optimisation happens at inference time through the evolving context $(d, H_t)$: the agent repeatedly conditions on prior knowledge (encoded in $d$ and the LLM) and on increasingly informative feedback (scores, breakdowns, and explanations) to guide subsequent proposals toward higher-scoring molecules with improved sample efficiency.

We implement SEISMO as a stateful LLM workflow that realizes the trajectory-conditioned, inference-time optimization procedure described above. Figure \ref{fig:workflow} provides an overview of the workflow. The concrete workflow graph structure, validation logic, prompting, and implementation are described in detail in Appendix \ref{sec:appendix:impl}. An ablation study of the LLM backbone is provided in Appendix \ref{sec:appendix:llm}. An example output trajectory is shown in Appendix \ref{sec:appendix:example-trajectory}.

\subsection{Tasks and oracles}
\label{sec:methods:tasks}

\subsubsection{Medicinal-chemistry oracles}
\label{sec:methods:medchem-oracles}

We evaluated the agent on medicinal chemistry tasks that represent common scenarios in pharmaceutical research. In hit-identification, the goal is to find an initial molecule that is active against a specific target. In lead-optimization, an active compound is further optimized for additional properties while remaining close to the lead structure. We implemented the oracles so that the level of information provided to the LLM could be controlled, allowing assessment of the impact of each source individually.

\paragraph{Hit-identification proxy (IC$_{50}$--QED--Novel).}
The objective is to minimise a predicted half-maximal inhibitory concentration ($\mathrm{IC}_{50}$) against SARS-CoV-2 main protease (M$^\mathrm{pro}$) while satisfying QED (Quantitative Estimate of Drug-likeness \cite{qed}) and novelty constraints. The $\mathrm{IC}_{50}$ values are predicted by an XGBoost regressor \cite{xgboost} trained on molecules featurised as 167-bit MACCS keys \cite{maccs} using the ASAP Discovery antiviral-potency-2025-unblinded dataset (training details in Appendix~\ref{sec:appendix:impl}). QED is computed with RDKit \cite{rdkit}. Novelty is defined as absence from PubChem \cite{pubchem}. Non novel molecules or molecules with QED values below $0.6$ are treated as invalid proposals and not considered towards the $\mathrm{IC}_{50}$ optimisation target. The task description refers only to a protease target, as explicitly naming SARS-CoV-2 M$^\mathrm{pro}$ triggered LLM safety-filter refusals, which we will discuss further in Section \ref{sec:discussion}.

\paragraph{Lead-optimisation proxy (Sim--QED).}
Given a reference molecule $x^\star$, the goal is to maximise Tanimoto similarity (sim) to $x^\star$ over MACCS-key fingerprints \cite{maccs} while increasing QED. The scalar objective is $s = \tfrac{1}{2}(\mathrm{sim}+\mathrm{QED})$, so $s \ge 0.8$ corresponds to average similarity and QED of at least $0.8$. Two regimes are studied: (i)~a praxis-motivated case study with $x^\star=$~Quercetin \cite{quercetin}, a molecule of known therapeutic interest but poor drug-like properties and (ii)~a randomised setting using 20 reference molecules sampled from PubChem under diversity and difficulty constraints (molecular weight $<500$, $\mathrm{QED}\le 0.5$, pairwise Tanimoto similarity $\le 0.2$; cf.~Appendix~\ref{sec:appendix:impl}).

\paragraph{Information-level conditions.}
% we should explain that no description does not get a description of the specific task but a generic minimize/maximize the score prompt (together with scalar-only feedback most similar to RF strategy); the others get task specific prompts
We evaluate four feedback variants:
\begin{itemize}
    \item \emph{no description}: Prompt only mentions a generic scalar score without any task specific information or score breakdown.
    \item \emph{no explanation}: Detailed task description and biological description of subscores $f_k$.
    \item \emph{partial explanation} Similar to no explanation but adds an explanation of the drug-likeness (QED) subscore on which molecular properties contributed how much in reducing it and what the optimal ranges are.
    \item \emph{full explanation} Similar to partial explanation, but includes explanations also for every other subscore $f_k$ in the objective. 
    
\end{itemize}

For the XGBoost $\mathrm{IC}_{50}$ model, we used SHAP \cite{shap} to compute an attribution value for each MACCS fingerprint feature. We reported the 10 highest and 10 lowest attribution values in a structured table additionally containing the fingerprint value, the MACCS key number, and the associated SMARTS pattern. The SMARTS patterns were included to provide the detailed chemical feature associated with the key to the model. We want to emphasize that feature attribution methods like SHAP do not establish the causal effects of molecular edits therefore they should be interpreted as model based empirical guidance rather than mechanistic explanations. For QED we provide a property-level decomposition (molecular weight, logP, HBA/HBD, PSA, rotatable bonds, aromatic-ring count and alerts) and indicate which properties most reduced the score and the optimal ranges. For similarity we explain which MACCS keys differ between $x^\star$ and the proposal and map these keys again to SMARTS patterns. Example explanations are given in Appendix~\ref{sec:appendix:examples}.

\subsubsection{Co-folding oracle}
\label{sec:methods:cofolding-oracle}

Co-folding models \cite{abramson2024accurate,passaro2025boltz2,corley2025accelerating} take protein sequences, small molecules (as SMILES), or other biological inputs and predict a 3D structure of the complex. They represent the high-cost end of the spectrum that motivates this work: a single call takes minutes on an A100 GPU, so the oracle is only practical when the number of calls is small. This is precisely the regime in which sample efficiency is decisive. We use Boltz-2 \cite{passaro2025boltz2}, which augments structure prediction with a binding-affinity head that returns a sigmoid probability that the small molecule binds to the protein (\texttt{affinity\_probability\_binary}) and a predicted affinity value that can be approximately interpreted as an $\mathrm{IC}_{50}$-like measure of binding strength. For identifying potential hits across diverse molecular screening scenarios, the authors recommend using the binding affinity probability value \cite{passaro2025boltz2}. We therefore use this probability as the optimisation score.

As protein target we chose MGYP001550541752, which was sampled at random from the MGnify metagenomics database \cite{richardson2023mgnify}, restricted to (i) low density regions of the ESMAtlas embedding space \cite{esm} and (ii) high confidence predicted 3D structures. The protein has no known small molecule binders, which removes LLM memorisation as a confounding factor. The high confidence structure prediction may also help preserve the reliability of Boltz-2 for this target.

In this case three information-level conditions are being evaluated:
\begin{itemize}
    \item \emph{no sequence}: generic task description and scalar binding-affinity probability only
    \item \emph{no explanation}: additionally provides the target protein sequence in the task description
    \item \emph{full explanation}: additionally provides three-letter codes of all residues within 5\,\AA{} of the ligand and Boltz-2 confidence scores
\end{itemize}

\subsubsection{PMO benchmark oracles}
\label{sec:methods:pmo-oracles}

For comparison against established methods we additionally evaluate on the Practical Molecular Optimization benchmark \cite{pmo_benchmark}, which consists of 23 molecule-optimisation tasks designed to assess sample efficiency. We invoke the benchmark oracles via the TDC package \cite{tdc} without modification. Two variants are tested: a \emph{task described} variant whose initial prompt states the objective, and a \emph{no description} variant in which the agent treats the oracle as fully black box. Importantly, PMO oracles return only a scalar score and no explanatory feedback is provided. The PMO oracles used in this benchmark are computationally inexpensive surrogate models with millisecond scale evaluation times, making millions of evaluations feasible. In this setting, LLM inference is the main computational bottleneck. This, together with the possibility that the tasks and their solutions may be represented in the training data, constrains the conclusions that can be drawn from the benchmark results (§\ref{sec:discussion}). We therefore include a zero shot LLM baseline, but limit both SEISMO and the zero shot baseline to 50 iterations. For comparison with prior work that used more iterations, our best score after 50 iterations is held fixed while the other methods are allowed to optimise further.

\subsection{Baselines}
\label{sec:methods:baselines}

\paragraph{Classical molecular optimisers.}
We compare against REINVENT \cite{reinvent}, Graph-GA \cite{graphga}, and Gaussian-process Bayesian optimisation (GP BO) \cite{gpbo, gpbo_original} on all three task families. For the PMO benchmark, classical-baseline results are obtained from the released per-iteration trajectories. For medicinal-chemistry tasks and the co-folding oracle, we run the baselines ourselves. Details and per-task results are reported in §\ref{sec:results} and Appendix~\ref{sec:appendix:impl}. For broader PMO context we additionally include published top-10 AUC values for MolLEO \cite{molleo}, LICO \cite{lico}, Genetic GFN \cite{genetic_gfn}, and Augmented Memory \cite{augmented} from Nguyen and Grover~\cite{lico}.

\paragraph{Zero-shot LLM.}
To bound the contribution of LLM priors and quantify the impact of the iterative oracle loop, we implemented two zero shot generation modes in which the LLM generates n molecules without intervening feedback at a sampling temperature of 1.0. In \emph{description} mode, the task description is similar to that used in the SEISMO runs. In \emph{no description} mode, only a generic task description is provided making this setup similar to the SEISMO no description mode, but without feedback. This batched generation can be viewed as a parallel best of N regime, as described in recent work on test time compute optimisation \cite{test-time}. The sequential sampling variant described there, where $n$ LLM instances are independently prompted to each generate one molecule, produced comparable but slightly more redundant molecules and was therefore omitted as an additional baseline.

\subsection{Metrics and evaluation protocol}
\label{sec:methods:metrics}

Consistent with the PMO benchmark \cite{pmo_benchmark}, we evaluate sample efficiency using the area under the optimisation curve (AUC). The optimisation curve is
\begin{equation*}
  s_t = \max_{i \le t} F(m_i),
\end{equation*}
the best score achieved up to oracle call $t$. AUC is the area under $s_t$ normalised by the oracle budget, so values lie in $[0,1]$ as long as $F\in[0,1]$. Higher AUC corresponds to faster discovery of high-scoring molecules. We report top-1 AUC and best-at-budget for the medicinal-chemistry and co-folding tasks. For PMO we additionally report top-10 AUC for comparison with prior work. For our Quercetin lead-optimisation case we report a goal-based metric: the number of oracle calls required to reach the optimisation goal of $s\ge 0.8$. This metric corresponds to the well-known goal of improving quercetin's poor drug-likeness while retaining its scaffold, and gives medicinal chemists a familiar, easily interpretable success criterion.

Unless stated otherwise, each setting is repeated independently three times. The main evaluation uses a budget of 50 oracle calls per run. For comparability with prior PMO work, we also report results at the 1,000 and 10,000 call horizons, holding SEISMO fixed at its best score achieved within the 50 call budget. In the framework analysis (§\ref{sec:results:framework_analysis}), we additionally report an extended run of 133 iterations, which exhausts the 200,000 token context window of Opus 4.5. For the 20 target Sim--QED setting, three optimization runs are performed for each reference molecule. Every result is reported together with the standard deviation, except the results of the PMO benchmark. For the PMO benchmark, 95\% confidence intervals are computed by bootstrap resampling, as the reported score aggregates many tasks, which yields a more stable and representative confidence interval than would be obtained for a single scalar result. 

\paragraph{Iso-quality budget.} To establish how much budget the classical methods need to reach SEISMO's quality, we reran Graph-GA, REINVENT and GP BO on the 20 SIM--QED references to a horizon of 10,000 oracle calls, with three replicates per reference. For each run we record the first oracle call at which its best-so-far score reaches the mean score SEISMO attained within 50 calls over the three repetitions for that reference molecule (Appendix~\ref{sec:appendix:iso-quality}).

\paragraph{Statistical analysis.} IC$_{50}$--QED--Novel uses 10 replicates per variant, SIM--QED 3 replicates on each of 20 reference molecules. SIM--QED is analysed as a randomised block design over the references, with Tukey's honest significant difference for the information levels and paired $t$-tests against SEISMO for the alternatives. IC$_{50}$ uses the Alexander--Govern test with Games--Howell, and Welch's $t$-test against the alternatives.

%%%%%%%%%%%%%%%%%%%%%%%%%%%%%%%%%%%%%%%%%%%%%%%%%%%%%%%%%%%%%%%%%%%%%%%
\section{Results}
\label{sec:results}

\subsection{Medicinal-chemistry tasks}
\label{sec:results:medchem}

We first evaluate SEISMO on two task families that mirror common medicinal-chemistry iteration loops: hit identification (IC$_{50}$--QED--Novel against SARS-CoV-2 M$^\mathrm{pro}$) and lead optimisation (Sim--QED on Quercetin and on 20 random PubChem reference molecules). The tasks were designed to be as unaffected as possible by the influence of training data memorization: the novelty constraint excludes PubChem molecules, and the random references are sampled to have low a-priori familiarity.

\paragraph{Comparison to classical baselines.}
SEISMO with full explanations outperforms all three classical baselines and the zero-shot LLM on both tasks, in every case statistically significant (Table \ref{tab:medchem-main}, Appendix~\ref{sec:appendix:pairwise-significance}). This includes both our proxy for hit-identification (IC$_{50}$--QED--Novel) and the lead-optimisation proxies (SIM-QED). For the hit-identification proxy, it increased AUC of pIC$_{50}$ against SARS-CoV-2 M$^\mathrm{pro}$ by 10\% from 6.57 to 7.25 compared to the best performing classical baseline. For the lead-optimisation proxy against 20 random molecules, it increased the AUC by 22\% from 0.64 to 0.78 compared to the best performing classical baseline. The optimisation curves from which the AUC was calculated, can be seen in Appendix \ref{sec:appendix:baselines_med_chem_curves}. For the lead-optimisation proxy against Quercetin, SEISMO managed to reach the target of finding a similar but more drug-like molecule within 4.6 iterations on average, while none of the classical baselines were able to reach the target within the budget of 50 oracle calls (Table \ref{tab:medchem-main}). 

Graph-GA, REINVENT and GP BO were run with the \texttt{mol-opt} defaults. Under these defaults all three only score their initial set of molecules and do not make a single optimisation step, as Graph-GA samples an initial population of 120 molecules, GP BO starts from 340 samples and REINVENT scores batches of 64, all of which are larger than the budget of 50 evaluations. We therefore also investigated reducing their hyperparameters to allow them multiple optimisation steps. We found that this did not change the results (Appendix~\ref{sec:appendix:baseline-configs}). For none of the three methods did any alternative setting significantly improve the score compared to its default. 

To see how long the baselines need to find a molecule of similar quality to the one SEISMO found in 50 calls on the SIM-QED task, we ran them for up to 10,000 oracle calls with their default parameters. In 27\% of runs they never found a molecule of similar quality, and in the remaining 73\% they needed on average more than 4,000 oracle calls to do so (Appendix~\ref{sec:appendix:iso-quality}).

\paragraph{Information-level ablation.}
A natural follow-up question is where these large improvements come from. Since we incorporated different additional sources of available information (see §\ref{sec:methods:problem}), we want to disentangle the contributions of using strong prior knowledge, our agentic optimisation loop, and explanations beyond scalar scores. Therefore, we incorporated the ablations, which can be seen in Table \ref{tab:medchem-main}. The first thing we observe is the strong gains from prior knowledge. The zero-shot LLM variant with description performs comparably to the classical baselines, exceeding all three on Sim-QED (0.70 vs.\ 0.64) while trailing the strongest baselines on IC$_{50}$. This shows that just incorporating generally available knowledge for medicinal chemistry tasks can already give a strong prior when optimising for binding affinity, similarity and drug-likeness. We also observe the importance of a task description, which allows the agent to condition its proposals on task-relevant prior knowledge encoded in the LLM. Without the task description both the zero-shot variant and the SEISMO variant perform significantly worse than all other variants and baselines. 

The next thing we observe is the benefit of our agentic optimisation loop. When comparing the SEISMO variants to their corresponding zero-shot variants, we can see a consistently better performance for SEISMO. This is true both for the no description variants (3.04 vs. 0.00 for AUC of pIC$_{50}$, and 0.58 vs. 0.49 for AUC of Sim-QED) and the no explanation SEISMO variant, which corresponds to the zero-shot LLM with descriptions (6.19 vs. 5.24 for AUC of pIC$_{50}$, and 0.76 vs. 0.70 for AUC of Sim-QED). In addition, only the variants where we have our optimisation loop were able to reach the target for the Quercetin task within 50 iterations, except for a single repetitions of the zero-shot LLM with description.

\begin{figure*}[t]
  \centering
  % --- Full-width legend strip on top ---
  \includegraphics[width=\linewidth]{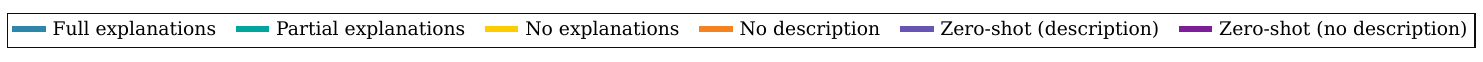}

  \vspace{0.1em}  % tune the gap between legend and plots

  % --- Two plots side by side below ---
  \begin{subfigure}[t]{0.48\textwidth}
    \includegraphics[width=\linewidth]{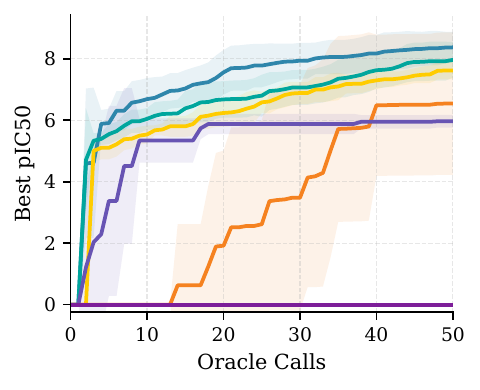}
    \caption{Comparing the four information-level of SEISMO with the zero-shot LLM baselines for the IC$_{50}$-QED-Novel task. Lines and bands denote the mean and standard deviation over 10 independent runs per SEISMO variant and 5 independent runs per zero-shot condition.}
    \label{fig:ic50_zeroshot}
  \end{subfigure}\hfill
  \begin{subfigure}[t]{0.48\textwidth}
    \includegraphics[width=\linewidth]{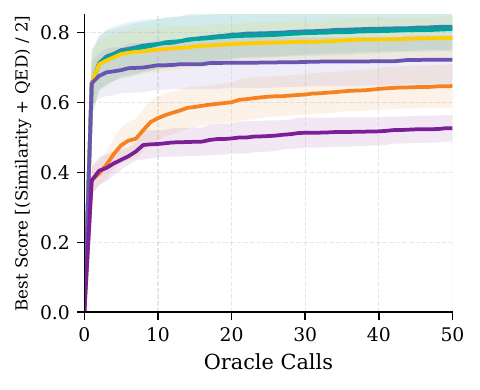}
    \caption{Comparing the four information levels of SEISMO with the zero-shot LLM baselines for the SIM--QED task. Every variant was run on the same 20 reference molecules with 3 independent replicates per molecule. Lines and bands denote the mean and standard deviation across the 20 reference molecules.}
    \label{fig:sim_qed_zeroshot}
  \end{subfigure}

  \caption{Best-so-far optimisation curves under different oracle-feedback variants for the medicinal-chemistry settings.}
  \label{fig:medchem-curves}
\end{figure*}

The last result we observe in Table \ref{tab:medchem-main} is the benefit of adding explanations of the score. Every comparison between information levels is statistically significant except full against partial explanations on Sim--QED, and partial against no explanations on either task (Appendix~\ref{sec:appendix:pairwise-significance}). Full explanations improve the outcome significantly over not having explanations on both tasks, while the marginal value of the full SHAP table over a partial explanation is supported on IC$_{50}$ only. The benefit of richer explanations also grows over the trajectory (Figure \ref{fig:medchem-curves}). Early curves are more similar across explanation levels, but variants with more feedback increasingly pull away as more oracle calls accumulate, consistent with the agent integrating explanation-derived insights into subsequent proposals. The increase over all tasks shows the importance and benefit of our use of explainability methods as a learning signal.

Finally, we ask whether conditioning on the complete trajectory is necessary, by truncating the history. Keeping only the last two steps ($p=0.004$), or only the single best and single worst ($p<10^{-4}$), significantly degrades performance, whereas for the ten most recent steps or the five best and five worst we saw no significant difference from the complete history on the Sim--QED task (Appendix~\ref{sec:appendix:history}).

All of the results of these ablations follow a similar trend: Providing more information increased the sample-efficiency of the agent. In addition, they show that each individual component of our design fulfills its purpose and provides a compound increase to our system.

% Generated by scripts/stats/medchem_stats.py, do not edit by hand.
\begin{table*}[t]
  \caption{Results of the medicinal-chemistry tasks, comparing four information levels of SEISMO with the LLM zero-shot capabilities with and without task description and three classical generative models as additional baselines. Cells report the mean and standard deviation over all runs of a condition, with 10 runs per SEISMO variant and 5 per baseline on IC$_{50}$--QED--Novel, 3 replicates on each of the same 20 random reference molecules for Sim--QED, and 5 runs per condition for Quercetin.}
  \label{tab:medchem-main}
  \centering
  \resizebox{\textwidth}{!}{%
  \begin{tabular}{lccc}
    \toprule
    Model variant
    & IC$_{50}$--QED--Novel
    & Sim--QED, 20 molecules
    & Sim--QED, Quercetin\\
    & SARS-CoV-2 M$^\mathrm{pro}$
    & top-1 AUC $\uparrow$
    & oracle calls to target $\downarrow$\\
    & top-1 AUC of pIC$_{50}$ $\uparrow$ & &\\
    \midrule
    SEISMO -- full explanation       & $\mathbf{7.25 \pm 0.42}$ & $\mathbf{0.78 \pm 0.07}$ & $\mathbf{4.6 \pm 2.1}$\\
    SEISMO -- partial explanation    & $6.62 \pm 0.42$ & $0.78 \pm 0.07$ & $5.4 \pm 1.3$\\
    SEISMO -- no explanation         & $6.19 \pm 0.21$ & $0.76 \pm 0.06$ & $14.2 \pm 10.8$\\
    SEISMO -- no description         & $3.04 \pm 1.60$ & $0.58 \pm 0.06$ & $>50$\\
    \addlinespace[0.3em]
    Zero-shot LLM (description)      & $5.24 \pm 0.44$ & $0.70 \pm 0.07$ & $26^{*}$\\
    Zero-shot LLM (no description)   & $0.00 \pm 0.00$ & $0.49 \pm 0.06$ & $>50$\\
    \addlinespace[0.3em]
    REINVENT                         & $6.57 \pm 0.21$ & $0.64 \pm 0.05$ & $>50$\\
    Graph-GA                         & $6.33 \pm 0.36$ & $0.64 \pm 0.07$ & $>50$\\
    GP BO                            & $6.18 \pm 0.65$ & $0.64 \pm 0.06$ & $>50$\\
    \bottomrule
  \end{tabular}}
  \vspace{0.35em}
  {\footnotesize\raggedleft $^*$Only 1 of 5 repetitions reached the target.\par}
\end{table*}

\subsection{Co-folding-based optimisation}
\label{sec:results:cofolding}

We next evaluate SEISMO using a co-folding model as the oracle. We select a target with no known small-molecule binders. For such a target, strong performance cannot be attributed to the LLM recalling ligands from its training data. This regime is difficult to achieve with traditional machine learning models, as they would require known binders as training data to produce reliable predictions. We exploit this to study MGYP001550541752, a metagenomic protein with no known binder and no resolved structure, for which memorisation is effectively excluded. We also include ablations to isolate the contribution of structured feedback on the novel target, and finally verify that the agent is driven by the external signal rather than its own priors (oracle ablations).

\begin{figure*}[t]
  \centering
  % --- Two plots side by side below ---
  \begin{subfigure}[t]{0.48\textwidth}
    \includegraphics[width=\linewidth]{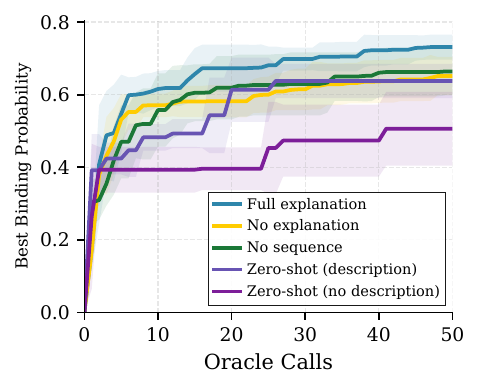}
    \caption{Comparison of the three information level conditions of SEISMO (full explanation, no explanation and no sequence) with the zero-shot LLM baselines either with task description or without.}
    \label{fig:mgyp_zeroshot}
  \end{subfigure}\hfill
  \begin{subfigure}[t]{0.48\textwidth}
    \includegraphics[width=\linewidth]{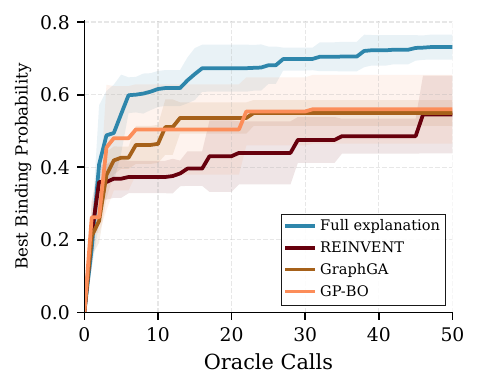}
    \caption{Comparison of the best-performing SEISMO mode (full explanation) with the three classical baseline models.}
    \label{fig:mgyp_baselines}
  \end{subfigure}

  \caption{Best-so-far optimisation curves using Boltz-2 as oracle and MGYP001550541752 as the target protein. Lines and bands denote the mean and standard deviation over 5 independent runs per SEISMO condition and 3 independent runs per zero-shot condition and per classical baseline.}
  \label{fig:mgyp_boltz-2}
\end{figure*}

\paragraph{Novel protein (MGYP001550541752).}
We test three different SEISMO variants for MGYP001550541752: a variant providing the protein sequence, which gets feedback explanations where the proposed molecules are predicted to bind; a similar variant but without explanations; and a variant that neither receives the protein sequence nor any explanations. All three variants allow SEISMO to increase the predicted binding-affinity probability from the received optimisation signal, but their performances differ (Figure~\ref{fig:mgyp_zeroshot}). The full-explanation condition, which adds three-letter codes of residues within 5\,\AA{} of the ligand together with Boltz-2 confidence scores, achieved the best performance. By contrast, the no-explanation and no-sequence conditions are nearly indistinguishable, indicating that the protein sequence alone is not enough for the agent to identify productive binding sites on an unfamiliar target. Close-residue feedback seems to be required for the LLM to extract usable structural context. Structured oracle feedback thus seems to contribute most precisely when the model's priors are weakest. We note that for MGYP001550541752 scores reach a binding probability of approximately 0.75 within 50 oracle calls, which is noticeably lower than for M$^\mathrm{pro}$, which reflect the lack of prior data. However, when we compare this value to classical optimisation baselines, we observe that SEISMO increases the binding probability by around 0.20 relative to them for the novel protein when all methods are limited to 50 iterations (Figure~\ref{fig:mgyp_baselines}.).

\paragraph{Oracle ablations.}
Finally, we investigate further how the agent weighs the external oracle signal against its own chemical priors. To this end, we perform two oracle ablations on the SARS-CoV-2 M$^\mathrm{pro}$ target (Appendix~\ref{sec:appendix:cofolding-extra}, Figure~\ref{fig:cofolding-ablations}). When the LLM is given the green fluorescent protein sequence as a decoy while the oracle continues to score against the true M$^\mathrm{pro}$ structure, the optimisation curve still trends upward but more slowly than with the correct sequence. The contextual information therefore matters, yet even with a misleading prior the agent improves on the basis of the feedback alone. Further, when the affinity probability is inverted before being returned to the agent, the curve consistently decreases, showing that the agent optimises for the score it receives however contradictory that score is to its prior information. Together these ablations indicate that the agent weights the oracle signal above its internal knowledge, a behaviour required for an optimiser intended to generalise beyond familiar targets.

\subsection{Cross-literature context: the PMO benchmark}
\label{sec:results:pmo}

For consistency with prior cheminformatics literature we additionally evaluate SEISMO on the 23 PMO tasks \cite{pmo_benchmark}. To stay consistent with the original implementations of the PMO oracles, we did not modify them to provide additional explanations of the scores.  Thus, we only provide an initial task description and then subsequently the scalar feedback scores in our optimisation loop. Since all reported metrics are bounded in $[0,1]$, the maximum achievable aggregate score is bound by 23.

\paragraph{Caveats up front.} % SANTI cool paragrapth title.
As noted above, SEISMO uses a natural-language task description, which lets it exploit prior chemical knowledge, and we provide such descriptions for the PMO tasks. The non-LLM baselines (Graph-GA, REINVENT, GP-BO) cannot make use of this text. We use descriptions comparable to those of MolLEO \cite{molleo}. Some of these descriptions make a task substantially easier. For example, when the objective is to match a molecular formula (the isomer tasks), the formula is stated directly, which is a large advantage. Likewise, when optimizing the geometric mean of similarities to two reference molecules, both references are given explicitly. And in the most extreme case, rediscovering a named molecule by structural (Tanimoto) similarity, an LLM that has memorized that molecule's structure during pretraining can solve the task almost immediately.

However, providing this information reflects how molecular optimization actually works in practice. To score a molecule at all, we must already know what we are scoring for: the substructures we require, the reference compounds we are trying to resemble, and how we weigh the individual components that make up the score. None of this appears at evaluation time out of nowhere. It is the very definition of the objective, and so in practice it is always available before optimization begins. Providing it to the model is therefore not an extra assumption but simply a matter of not discarding what we already hold. We do, however, caution the reader to keep this advantage in mind when comparing against methods that cannot incorporate a task description. To make the effect of using this available information measurable, we additionally report a no-description ablation of SEISMO, which removes this information entirely.

\paragraph{Performance.}
Table~\ref{tab:pmo-summary} compares the performance of SEISMO to the baseline models. Across all metrics, SEISMO achieves consistently better performance than all baselines. Summed over all 23 tasks, the top-1 AUC over 50 oracle calls is 18.12 for SEISMO, compared to 7.06 for the strongest classical baseline, corresponding to a 2.6$\times$ improvement. For the sum of best scores after 50 oracle calls, SEISMO achieves a 2.3$\times$ improvement (18.96 vs.\ 8.18). Even when compared using the top-1 AUC over 10{,}000 oracle calls, SEISMO achieves a 1.3$\times$ improvement, despite being restricted to 50 oracle calls compared to the full budget for the baselines.

Examining individual tasks rather than aggregate metrics reinforces this pattern. Across both metrics evaluated over 50 oracle calls, SEISMO achieves the best performance on 21 of 23 tasks (cf.~Tables~\ref{tab:si-top1-50-iterations} and \ref{tab:si-best-score-50-iterations}). Over 10,000 oracle calls, SEISMO remains best on 18 of the 23 tasks (cf.~Table~\ref{tab:si-top1-10000-iterations}). Inspecting best scores after 50 oracle calls at the task level shows that SEISMO already reaches scores within 10\% of the task maximum ($\geq 0.9$) on 16 of 23 tasks (cf.~Table ~\ref{tab:si-best-score-50-iterations}). Notably, this occurs within a number of oracle calls that is comparable to typical initial batch or population sizes used by prior methods, which have to evaluate multiple candidates before any update step can be taken.

This strong performance is partly explained by the strong performance of zero-shot LLM, which proposed 50 different molecules based on the task description, but without the feedback loop. We attribute the strong performance of the zero-shot baseline to the task difficulty mentioned in the caveat. Nevertheless, we can see that the feedback loop still improves the performance further across all 3 metrics, bringing them close to maximal score of 23. We can also again see a large difference between SEISMO with a description and no description, which as we argued before is necessary to be able to utilize the useful prior knowledge and is consistent with our results from the previous experiments.

In the extended benchmark results (Table \ref{tab:auc1000_aggregated}), we compare SEISMO also against the reported performance of more recent methods, which also include other LLM utilising methods. SEISMO again achieves the highest aggregate performance, improving the best score by a factor of~1.6. At the level of individual tasks, SEISMO is the best-performing model on 22 of the 23 tasks (see Table~\ref{tab:si-top10-1000-iterations}). We note again that most competing methods cannot incorporate available task descriptions, so this comparison should be read with that asymmetry in mind. Nonetheless, the pattern mirrors our earlier findings: SEISMO's sample-efficiency arises from effectively exploiting the prior information at hand, an effect that is further amplified by our feedback loop.

\begin{table}[t]
  \caption{Aggregate PMO performance summed over 23 tasks. Top-1 AUC at 50 calls; top-1 AUC at 10{,}000 calls (SEISMO held fixed after 50); best score at 50 calls.}
  \label{tab:pmo-summary}
  \begin{center}
  \begin{small}
  \begin{sc}
  \begin{tabular}{lccc}
    \toprule
    Model
    & \shortstack{Top-1 AUC \\ 50 ($\uparrow$)}
    & \shortstack{Top-1 AUC \\ 10k ($\uparrow$)}
    & \shortstack{Best \\ @50 ($\uparrow$)} \\
    \midrule
    SEISMO                            & $\mathbf{18.12}\,^{+0.26}_{-0.27}$ & $\mathbf{18.95}\,^{+0.32}_{-0.39}$ & $\mathbf{18.96}\,^{+0.32}_{-0.39}$\\
    SEISMO -- no description          & $8.50\,^{+0.82}_{-0.76}$           & $10.44\,^{+0.90}_{-1.09}$          & $10.45\,^{+0.90}_{-1.09}$ \\
    \addlinespace[0.3em]
    Zero-shot LLM            & $16.43\,^{+0.45}_{-0.45}$ & $17.66\,^{+0.40}_{-0.40}$ & $17.66\,^{+0.40}_{-0.40}$ \\
    Zero-shot LLM -- no description     & $4.62\,^{+0.21}_{-0.21}$ & $5.76\,^{+0.35}_{-0.35}$ & $5.77\,^{+0.35}_{-0.35}$ \\
    \addlinespace[0.3em]
    REINVENT                          & $7.06\,^{+0.50}_{-0.42}$           & $14.74\,^{+0.95}_{-0.51}$          & $8.06\,^{+0.88}_{-0.62}$  \\
    Graph-GA                          & $7.04\,^{+0.49}_{-0.42}$           & $14.38\,^{+0.45}_{-0.47}$          & $8.18\,^{+0.55}_{-0.48}$  \\
    GP BO                             & $6.95\,^{+0.34}_{-0.31}$           & $13.83\,^{+0.60}_{-0.60}$          & $8.00\,^{+0.53}_{-0.48}$   \\
    \bottomrule
  \end{tabular}
  \end{sc}
  \end{small}
  \end{center}
\end{table}

\begin{table}[t]
  \caption{Top-10 AUC over 1,000 oracle calls on the Practical Molecular Optimization (PMO) benchmark, summed over 23 tasks. Baseline results are taken from prior work \cite{lico}. SEISMO proposes molecules only during the first 50 oracle calls, after which its top-10 score is held fixed, while all baselines continue optimizing for the full 1,000 calls.}
  \label{tab:auc1000_aggregated}
  \begin{center}
    \begin{small}
      \begin{sc}
        \begin{tabular}{lcc}
          \toprule
          Model & Sum of AUC's \\
          \midrule
          SEISMO                    & \textbf{18.16} \\
          LICO                          & 11.71 \\
          MOLLEO                        & 11.65 \\
          Genetic GFN                  & 11.56  \\
          GP BO                        & 11.27  \\
          Graph GA                     & 10.90  \\
          Augmented Memory             & 10.81  \\
          REINVENT                     & 10.68  \\
          SEISMO -- no description     & 9.90   \\
          \bottomrule
        \end{tabular}
      \end{sc}
    \end{small}
  \end{center}
  \vskip -0.1in
\end{table}

\subsection{Framework Analysis and Resource Efficiency}
\label{sec:results:framework_analysis}

\begin{figure}[t]
  \centering
  \includegraphics[width=0.5\linewidth]{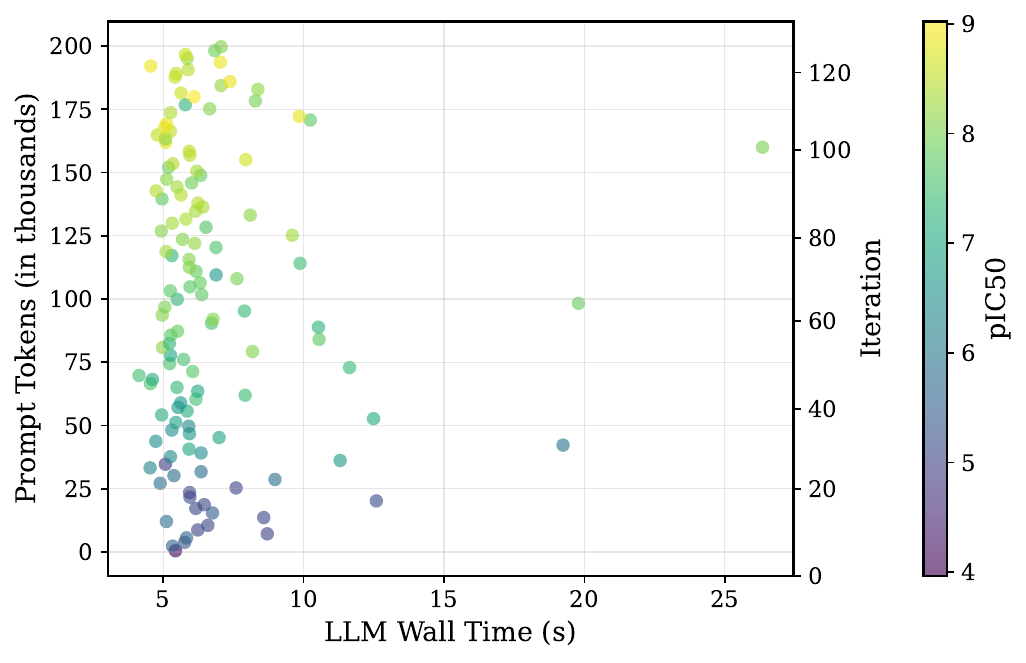}
  \caption{Extended 133-iteration SEISMO run on the IC$_{50}$--QED--Novel task in full-explanation mode. The agent identifies a novel molecule with predicted $\mathrm{IC}_{50}\approx 0.96$\,nM at iteration~119; per-call wall-clock time does not increase with longer context lengths from higher iterations.}
  \label{fig:extended-run}
\end{figure}

To assess long horizon behaviour, we additionally report an extended run beyond the main 50 iteration budget. On the IC$_{50}$–QED–Novel task in full explanation mode, the agent was run for 133 iterations until the 200,000 token context window of Opus 4.5 was reached. The best molecule was identified at iteration 119, with predicted IC$_{50}$ of 0.96 nM and QED of 0.76, improving on all molecules found in the 50 iteration runs. The optimization curve continues to improve over the extended horizon, with no evidence of degradation at longer context lengths (Figure~\ref{fig:extended-run}).

Robustness is supported not only by individual runs, but also by consistent performance across tasks. Across the medicinal chemistry tasks IC$_{50}$–QED–Novel, Sim–QED with 20 reference molecules, and Sim–QED with Quercetin, SEISMO achieved the strongest performance in terms of top-1 AUC and oracle calls to target while consistently remaining among the methods with the lowest standard deviation (Table~\ref{tab:medchem-main}). However, this robustness does not prevent the method from identifying multiple optimized solutions. Across three runs initialized from the same starting molecule generated in the first iteration, the UMAP projections show related early search behaviour followed by convergence toward different and distinct regions of chemical space (Figures~\ref{fig:appendix-exploration-exploitation-trajectory-rep1},~\ref{fig:appendix-exploration-exploitation-trajectory-rep2},~\ref{fig:appendix-exploration-exploitation-trajectory-rep3}). These results suggest that the framework is stable in outcome quality while retaining flexibility in the molecular search path and found optimized solutions.

In several runs, the optimization summaries qualitatively resemble phased medicinal chemistry refinement, with early scaffold level changes followed by more localized substituent modifications. The exploration and exploitation tradeoff is also evident in the optimization trajectories for each iteration. Figure~\ref{fig:appendix-exploration-exploitation-basins} uses colors to show the similarity of the generated molecules to one another, and in later iterations, longer contiguous patches of similar molecules become visible. This pattern is consistent with a transition from broad exploration to focused exploitation. All three runs finish within a similar pIC50 range, which again highlights the robustness of the approach. The figure also provides an intuition for how many designs did not satisfy the QED and novelty constraints for this task, with additional details provided in Appendix~\ref{sec:appendix:chemspace}.

Resource use differed substantially across information modes. The mean per iteration token addition was approximately 1,371 in full explanation mode, 485 in partial explanation mode, 305 in no explanation mode, and 137 in no description mode. Full explanation mode therefore reached the context limit fastest, although with a 1 M token context window the same workflow would support approximately 759 full explanation iterations before summarization becomes necessary. Mean LLM wall clock time in the extended run was 6.7 s per iteration and did not increase materially with context length. Based on the logged token counts and pricing assumptions described in Appendix~\ref{sec:appendix:cost-runtime}, a typical 50 iteration full explanation run incurred approximately 8.78 USD in LLM API cost, whereas the 133 iteration extended run incurred approximately 61 USD and completed in approximately 16 minutes. In co-folding settings, total wall clock cost was dominated by Boltz-2 evaluations rather than LLM inference. Detailed token accounting, runtime analysis, and cost estimates are provided in Appendix~\ref{sec:appendix:cost-runtime}.

%%%%%%%%%%%%%%%%%%%%%%%%%%%%%%%%%%%%%%%%%%%%%%%%%%%%%%%%%%%%%%%%%%%%%%%
\section{Discussion}
\label{sec:discussion}

A common theme we observed across every task family is that sample efficiency improves whenever the agent has access to more of the information already available about the optimisation problem: the task description, the sub-score decomposition, and structured post-hoc explanations of those sub-scores. While the observation that more available information increases sample efficiency is intuitive and should not come as a surprise on its own, the near-total absence of this information from existing molecular optimisers makes it worth emphasising. This is less a deliberate omission than a structural consequence of how classical optimisers are built. Genetic algorithms, Bayesian optimisation, and reinforcement-learning-based generators are all designed around a scalar fitness signal, with no natural interface for free-text task descriptions, sub-score breakdowns, or explanatory feedback \cite{reinvent, gpbo, graphga}. Such information therefore falls outside what these methods are designed to consume. LLMs remove much of that friction: natural-language conditioning makes it straightforward to hand an optimiser exactly the information a human expert would want, in a format the optimiser can act on directly \cite{opro}. The remainder of this section unpacks where the observed sample-efficiency gains exactly come from, what regime the approach is useful in, and current limitations.

\paragraph{Embracing LLM priors as a feature.}
A first source of SEISMO's performance is prior knowledge encoded in the LLM itself, and we treat this as a feature of the approach rather than a confound. Decades of medicinal-chemistry research have established broadly generalisable relationships between molecular structure and properties such as potency, selectivity, solubility, permeability, and metabolic stability \cite{lipinski, qsar}. An optimiser that ignores this accumulated knowledge must rediscover those regularities from scratch, which is inherently sample-inefficient. The relevant question is therefore not whether priors contribute to SEISMO's performance, since they should and clearly do, but whether the iterative oracle loop adds generalizability value beyond what the LLM's priors already provide zero-shot. Our zero-shot baselines confirm that it does, and most clearly on the tasks designed to limit memorisation: Sim--QED against 20 random PubChem references, novelty-constrained hit identification against SARS-CoV-2 M$^\mathrm{pro}$, and binding-affinity optimisation against a protein with no known ligands. The novel-protein result illustrates this most clearly: providing the protein sequence alone is not sufficient for the agent to identify a productive binding mode on this unfamiliar target, whereas providing close-residues feedback largely recovers the optimisation signal. This suggests that structured oracle feedback contributes most where the LLM's priors are weakest.

\paragraph{Explanation as a machine-readable control signal.}
Having established that the loop contributes beyond static priors, our second main finding is that enhancing this loop with post-hoc explainability methods increases sample efficiency and interpretability further. Across the medicinal chemistry settings we evaluated, this improvement was consistent, suggesting that the explanations provide an actionable signal for the optimiser itself. In this sense, explainability is valuable not only for human interpretation but also as a machine to machine communication tool, giving the agent richer structured information about its sub scores. We view this as an initial demonstration that explanation derived signals can shape search behaviour and open new directions for research on the role of explainable AI for molecular optimisation and other agentic systems.

\paragraph{Cost regime where SEISMO is advantageous.}
LLM inference is not free, and SEISMO is correspondingly not the right tool in every setting. It is best suited to problems where each oracle evaluation is itself expensive or rate-limited relative to a single LLM call, which is precisely the situation facing physics-based simulation, protein-ligand co-folding, and experimental assays, where a single evaluation can take minutes to weeks. In that regime, the total time spent on LLM calls is negligible compared to even a single oracle evaluation, so the sample-efficiency gains we observe translate almost directly into faster discovery. Conversely, where oracle calls are cheap and trivially parallelisable, SEISMO is not the right tool, since LLM inference itself becomes the computational bottleneck, even if SEISMO still wins on sample efficiency.

\paragraph{Limitations and future work.}
The present implementation bounds the effective optimisation horizon by the LLM's context window. Our main experiments focused on the regime of 50 oracle calls, where this limit is never reached (see estimated max. iteration number in Appendix \ref{sec:appendix:cost-runtime}). We also investigated history truncation and found that, when a sufficiently large number of iterations is retained, the full explanation mode can match the top 1 AUC performance obtained with the complete history (Appendix \ref{sec:appendix:history}), while reducing costs (Appendix \ref{sec:appendix:cost-runtime}). Further studies are needed to assess this approach in more detail, particularly at longer optimisation horizons. Most of our experiments also rely on a proprietary LLM backbone, selected because it performed best in our ablation. We found that explicitly naming the SARS-CoV-2 M$^\mathrm{pro}$ target triggered LLM safety filter refusals, so in the task description we refer only to a generic protease target. While we think it is vital that safety filters exist, they can limit research and drug discovery success, especially since we showed that giving the model more information enhances sample efficiency and generates more favored molecules. We see open-weights models as a promising future direction, provided they close the performance gap and reliably produce valid SMILES, as they allow for unrestricted generation and enable fine-tuning the backbones with domain specific data. More generally, because SEISMO places no demands on its backbone beyond structured generation and in-context reasoning, such models could be swapped in directly, and SEISMO's performance should improve as LLMs continue to advance, without requiring any change to the method itself.

Overall, the oracles used in this work are computational predictors, and improving their score more sample-efficiently does not by itself establish that this translates to better experimental performance. We demonstrate improved sample-efficiency across multiple inexpensive surrogate oracles and the Boltz-2 co-folding oracle in this work. We show that our findings hold for all of these oracles. We therefore assume that it also transfers to substantially more expensive oracles, such as physics-based binding-affinity calculations or even experimental assay. Whether this assumption holds will require validating it against these expensive evaluations, which we regard as the principal direction for future work. The framework is versatile and can be adjusted as needed, e.g., for batch-wise generation, and other molecular entities are conceivable in principle. We see particular potential in the communication that agents enable: medicinal chemists can read the LLM generated reasoning for each molecule and compare it with their own beliefs to calibrate their trust levels. This might be particularly relevant when post-hoc explanations of the oracles are supplied, since such explanations may also make it easier to exploit patterns that the oracle scores highly but that do not translate into real-world performance. <extra sentence> Additionally, through the familiar interface of LLMs, known from their widespread daily use, they allow chemists to easily and conveniently discuss the proposed molecules and express additional requirements for the next run, e.g., if they are familiar with the target or toxicological profiles and want to include or exclude specific substructure features.

\section*{Conclusions}
\label{sec:conclusions}

We introduced SEISMO, an LLM agent that reframes molecular optimisation as a grey-box problem. Rather than treating the evaluation function as a scalar black box, SEISMO conditions each proposal on the full optimisation trajectory and consumes sub-score decompositions and post-hoc explanations as a machine-readable feedback signal. SEISMO consistently outperforms classical population-based and batched optimisers as well as zero-shot LLM generation across every tested task, including settings designed to limit the LLM's reliance on memorised prior knowledge. These gains grow as explanatory feedback is enriched. SEISMO requires no training, fine-tuning, or hyperparameter tuning from the user. Because SEISMO places no demands on its backbone beyond structured generation and in-context reasoning, we expect its performance to improve as underlying LLMs continue to advance, without requiring any change to the method itself.

%%%%%%%%%%%%%%%%%%%%%%%%%%%%%%%%%%%%%%%%%%%%%%%%%%%%%%%%%%%%%%%%%%%%%%%
\section*{List of abbreviations}
\label{sec:abbreviations}

\noindent
AUC: area under the (optimisation) curve;
%BO: Bayesian optimisation;
%CI: confidence interval;
%ESM: Evolutionary Scale Modeling;
GFP: green fluorescent protein;
GP BO: Gaussian-process Bayesian optimisation;
HBA/HBD: hydrogen-bond acceptor/donor;
IC$_{50}$: half-maximal inhibitory concentration;
ipTM: interface predicted template modeling (score);
LLM: large language model;
MACCS: Molecular ACCess System (substructure keys);
MGYP: MGnify protein identifier;
M$^\mathrm{pro}$: main protease;
MPO: multi-property objective;
MSA: multiple sequence alignment;
pLDDT: predicted local distance difference test;
PMO: Practical Molecular Optimization (benchmark);
PSA: polar surface area;
QED: quantitative estimate of drug-likeness;
RDKit: open-source cheminformatics toolkit;
RMSE: root mean squared error;
SHAP: Shapley Additive Explanations;
SMARTS: SMILES Arbitrary Target Specification;
SMILES: Simplified Molecular Input Line Entry System;
TDC: Therapeutics Data Commons;
TPSA: topological polar surface area;
XAI: explainable AI;
XGBoost: extreme gradient boosting.

%%%%%%%%%%%%%%%%%%%%%%%%%%%%%%%%%%%%%%%%%%%%%%%%%%%%%%%%%%%%%%%%%%%%%%%
\section*{Declarations}
\label{sec:declarations}

\subsection*{Availability of data and materials}

The SEISMO source code, prompt templates, xgboost model weights, the raw data from all runs, and scripts to reproduce the results are available at \url{https://github.com/FabianKruger/molecule-optimization-agent}. The snapshot of the repository at the time of the paper is available at \url{https://doi.org/10.5281/zenodo.24687162}.

\subsection*{Competing interests}

Several authors (A.H., A.W., T.J.A., S.D.V.) are employed by Bayer~AG, a pharmaceutical company with commercial interests in molecular-optimisation methods. The remaining authors declare no competing interests.

\subsection*{Funding}

This work was partially funded by the Horizon Europe funding programme under the Marie Sk\l{}odowska-Curie Actions Doctoral Networks grant agreement ``Explainable AI for Molecules -- AiChemist'', no.~101120466. 

Additionally, this project was selected by Anthropic for its AI for Science program, which supported it in the form of API credits.

\subsection*{Authors' contributions}

FPK and AH: Conceptualization, Methodology, Software, Analysis, Writing - Original Draft, Writing - Review \& Editing, Visualization \\
AW and TJA: Methodology, Software, Writing - Original Draft, Writing - Review \& Editing \\
IT and SDV: Conceptualization, Writing - Review \& Editing, Supervision, Resources, Funding acquisition

\subsection*{Acknowledgements}

We thank the members of the Machine Learning Research Group at Bayer AG, especially Anastasia Pentina and Lennart Kinzel for their helpful discussions and valuable feedback that improved this work.

\subsection*{Use of Generative AI}
\label{sec:llm-use}

Generative large language models developed by Anthropic and OpenAI were investigated in this study as tools for molecular generation. These models were also used during framework development, including prompt optimization and coding assistance. During manuscript preparation, the authors used LLMs to support drafting and language editing in order to improve readability. All AI-assisted content was subsequently reviewed and edited by the authors, who take full responsibility for the final manuscript.
%%%%%%%%%%%%%%%%%%%%%%%%%%%%%%%%%%%%%%%%%%%%%%%%%%%%%%%%%%%%%%%%%%%%%%%
\bibliography{seismo}

@article{smiles,
author = {Weininger, David},
title = {SMILES, a chemical language and information system. 1. Introduction to methodology and encoding rules},
journal = {Journal of Chemical Information and Computer Sciences},
volume = {28},
number = {1},
pages = {31-36},
year = {1988},
}

@misc{rdkit,
  title        = {RDKit: Open-source cheminformatics},
  author       = {{RDKit}},
  year         = {2025},
  howpublished = {\url{https://www.rdkit.org}},
  doi          = {10.5281/zenodo.17746401},
  note         = {Release\_2025\_09\_3}
}

@misc{langgraph,
  author       = {{LangChain AI}},
  title        = {LangGraph: Stateful, Multi-Agent LLM Workflows},
  howpublished = {\url{https://github.com/langchain-ai/langgraph}},
  year         = {2025},
  note         = {Version 1.0.5. Accessed: 2026-01-08}
}

@inproceedings{tdc,
 author = {Huang, Kexin and Fu, Tianfan and Gao, Wenhao and Zhao, Yue and Roohani, Yusuf and Leskovec, Jure and Coley, Connor and Xiao, Cao and Sun, Jimeng and Zitnik, Marinka},
 booktitle = {Proceedings of the Neural Information Processing Systems Track on Datasets and Benchmarks},
 editor = {J. Vanschoren and S. Yeung},
 pages = {},
 title = {Therapeutics Data Commons: Machine Learning Datasets and Tasks for Drug Discovery and Development},
 volume = {1},
 year = {2021}
}

@article{polaris,
  title={A call for an industry-led initiative to critically assess machine learning for real-world drug discovery},
  author={Wognum, Cas and Ash, Jeremy R and Aldeghi, Matteo and Rodr{\'\i}guez-P{\'e}rez, Raquel and Fang, Cheng and Cheng, Alan C and Price, Daniel J and Clevert, Djork-Arn{\'e} and Engkvist, Ola and Walters, W Patrick},
  journal={Nature Machine Intelligence},
  volume={6},
  number={10},
  pages={1120--1121},
  year={2024},
  publisher={Nature Publishing Group UK London}
}

@article{pmo_benchmark,
  title={Sample efficiency matters: a benchmark for practical molecular optimization},
  author={Gao, Wenhao and Fu, Tianfan and Sun, Jimeng and Coley, Connor},
  journal={Advances in neural information processing systems},
  volume={35},
  pages={21342--21357},
  year={2022}
}

@article{chem_space,
  title={The art and practice of structure-based drug design: a molecular modeling perspective},
  author={Bohacek, Regine S and McMartin, Colin and Guida, Wayne C},
  journal={Medicinal research reviews},
  volume={16},
  number={1},
  pages={3--50},
  year={1996},
  publisher={Wiley Online Library}
}

@article{reinvent,
  title={Molecular de-novo design through deep reinforcement learning},
  author={Olivecrona, Marcus and Blaschke, Thomas and Engkvist, Ola and Chen, Hongming},
  journal={Journal of cheminformatics},
  volume={9},
  number={1},
  pages={48},
  year={2017},
  publisher={Springer}
}

@inproceedings{gpbo,
  title={A fresh look at de novo molecular design benchmarks},
  author={Tripp, Austin and Simm, Gregor NC and Hern{\'a}ndez-Lobato, Jos{\'e} Miguel},
  booktitle={NeurIPS 2021 AI for Science Workshop},
  year={2021}
}

@inproceedings{gpbo_original,
author = {Srinivas, Niranjan and Krause, Andreas and Kakade, Sham and Seeger, Matthias},
title = {Gaussian process optimization in the bandit setting: no regret and experimental design},
year = {2010},
isbn = {9781605589077},
publisher = {Omnipress},
address = {Madison, WI, USA},
booktitle = {Proceedings of the 27th International Conference on International Conference on Machine Learning},
pages = {1015–1022},
numpages = {8},
location = {Haifa, Israel},
series = {ICML'10}
}

@article{graphga,
  title={A graph-based genetic algorithm and generative model/Monte Carlo tree search for the exploration of chemical space},
  author={Jensen, Jan H},
  journal={Chemical science},
  volume={10},
  number={12},
  pages={3567--3572},
  year={2019},
  publisher={Royal Society of Chemistry}
}

@article{qsar,
  title={QSAR modeling: where have you been? Where are you going to?},
  author={Cherkasov, Artem and Muratov, Eugene N and Fourches, Denis and Varnek, Alexandre and Baskin, Igor I and Cronin, Mark and Dearden, John and Gramatica, Paola and Martin, Yvonne C and Todeschini, Roberto and others},
  journal={Journal of medicinal chemistry},
  volume={57},
  number={12},
  pages={4977--5010},
  year={2014},
  publisher={ACS Publications}
}

@article{molleo,
  title={Efficient evolutionary search over chemical space with large language models},
  author={Wang, Haorui and Skreta, Marta and Ser, Cher-Tian and Gao, Wenhao and Kong, Lingkai and Strieth-Kalthoff, Felix and Duan, Chenru and Zhuang, Yuchen and Yu, Yue and Zhu, Yanqiao and others},
  journal={arXiv preprint arXiv:2406.16976},
  year={2024}
}

@article{lipinski,
  title={Experimental and computational approaches to estimate solubility and permeability in drug discovery and development settings},
  author={Lipinski, Christopher A and Lombardo, Franco and Dominy, Beryl W and Feeney, Paul J},
  journal={Advanced drug delivery reviews},
  volume={64},
  pages={4--17},
  year={2012},
  publisher={Elsevier}
}

@inproceedings{chemagent,
  title={Chemagent: Self-updating memories in large language models improves chemical reasoning},
  author={Tang, Xiangru and Hu, Tianyu and Ye, Muyang and Shao, Yanjun and Yin, Xunjian and Ouyang, Siru and Zhou, Wangchunshu and Lu, Pan and Zhang, Zhuosheng and Zhao, Yilun and others},
  booktitle={The Thirteenth International Conference on Learning Representations},
  year={2025}
}

@article{frogent,
  title={Frogent: An end-to-end full-process drug design agent},
  author={Pan, Qihua and Xu, Dong and Yao, Jenna Xinyi and Ma, Lijia and Zhu, Zexuan and Ji, Junkai},
  journal={arXiv preprint arXiv:2508.10760},
  year={2025}
}

@inproceedings{madd,
  title={MADD: Multi-Agent Drug Discovery Orchestra},
  author={Solovev, Gleb V and Zhidkovskaya, Alina B and Orlova, Anastasia and Gubina, Nina and Vepreva, Anastasia and Golovinskii, Rodion and Tonkii, Ilya and Dubrovsky, Ivan and Gurev, Ivan and Gilemkhanov, Dmitry and others},
  booktitle={Findings of the Association for Computational Linguistics: EMNLP 2025},
  pages={6956--6998},
  year={2025}
}

@inproceedings{drugagent,
  title={DrugAgent: Multi-Agent Large Language Model-Based Reasoning for Drug-Target Interaction Prediction},
  author={Inoue, Yoshitaka and Song, Tianci and Wang, Xinling and Luna, Augustin and Fu, Tianfan},
  booktitle={ICLR 2025 Workshop on Machine Learning for Genomics Explorations},
  year={2025}
}

@article{cactus,
  title={Cactus: Chemistry agent connecting tool usage to science},
  author={McNaughton, Andrew D and Sankar Ramalaxmi, Gautham Krishna and Kruel, Agustin and Knutson, Carter R and Varikoti, Rohith A and Kumar, Neeraj},
  journal={ACS omega},
  volume={9},
  number={46},
  pages={46563--46573},
  year={2024},
  publisher={ACS Publications}
}

@article{chemcrow,
  title={Augmenting large language models with chemistry tools},
  author={Bran, Andres M. and Cox, Sam and Schilter, Oliver and Baldassari, Carlo and White, Andrew D and Schwaller, Philippe},
  journal={Nature Machine Intelligence},
  volume={6},
  number={5},
  pages={525--535},
  year={2024},
  publisher={Nature Publishing Group UK London}
}

@inproceedings{lico,
 author = {Nguyen, Tung and Grover, Aditya},
 booktitle = {International Conference on Learning Representations},
 editor = {Y. Yue and A. Garg and N. Peng and F. Sha and R. Yu},
 pages = {53576--53597},
 title = {LICO: Large Language Models for In-Context Molecular Optimization},
 url = {https://proceedings.iclr.cc/paper_files/paper/2025/file/84b16af3773fc5825ddb64996190289a-Paper-Conference.pdf},
 volume = {2025},
 year = {2025}
}

@article{beam_enumeration,
  title={Beam enumeration: probabilistic explainability for sample efficient self-conditioned molecular design},
  author={Guo, Jeff and Schwaller, Philippe},
  journal={arXiv preprint arXiv:2309.13957},
  year={2023}
}

@article{augmented,
  title={Augmented memory: sample-efficient generative molecular design with reinforcement learning},
  author={Guo, Jeff and Schwaller, Philippe},
  journal={Jacs Au},
  volume={4},
  number={6},
  pages={2160--2172},
  year={2024},
  publisher={ACS Publications}
}

@article{kunzmann2023biotite,
  title={Biotite: new tools for a versatile Python bioinformatics library},
  author={Kunzmann, Patrick and M{\"u}ller, Tom David and Greil, Maximilian and Krumbach, Jan Hendrik and Anter, Jacob Marcel and Bauer, Daniel and Islam, Faisal and Hamacher, Kay},
  journal={BMC bioinformatics},
  volume={24},
  number={1},
  pages={236},
  year={2023},
  publisher={Springer}
}

@article{scikit-learn,
  title={Scikit-learn: Machine Learning in {P}ython},
  author={Pedregosa, F. and Varoquaux, G. and Gramfort, A. and Michel, V.
          and Thirion, B. and Grisel, O. and Blondel, M. and Prettenhofer, P.
          and Weiss, R. and Dubourg, V. and Vanderplas, J. and Passos, A. and
          Cournapeau, D. and Brucher, M. and Perrot, M. and Duchesnay, E.},
  journal={Journal of Machine Learning Research},
  volume={12},
  pages={2825--2830},
  year={2011}
}

@article{dmta,
  title={Augmenting DMTA using predictive AI modelling at AstraZeneca},
  author={Ghiandoni, Gian Marco and Evertsson, Emma and Riley, David J and Tyrchan, Christian and Rathi, Prakash Chandra},
  journal={Drug discovery today},
  volume={29},
  number={4},
  pages={103945},
  year={2024},
  publisher={Elsevier}
}

@article{passaro2025boltz2,
  author = {Passaro, Saro and Corso, Gabriele and Wohlwend, Jeremy and Reveiz, Mateo and Thaler, Stephan and Somnath, Vignesh Ram and Getz, Noah and Portnoi, Tally and Roy, Julien and Stark, Hannes and Kwabi-Addo, David and Beaini, Dominique and Jaakkola, Tommi and Barzilay, Regina},
  title = {Boltz-2: Towards Accurate and Efficient Binding Affinity Prediction},
  year = {2025},
  doi = {10.1101/2025.06.14.659707},
  journal = {bioRxiv}
}

@article{abramson2024accurate,
  title={Accurate structure prediction of biomolecular interactions with AlphaFold 3},
  author={Abramson, Josh and Adler, Jonas and Dunger, Jack and Evans, Richard and Green, Tim and Pritzel, Alexander and Ronneberger, Olaf and Willmore, Lindsay and Ballard, Andrew J and Bambrick, Joshua and others},
  journal={Nature},
  volume={630},
  number={8016},
  pages={493--500},
  year={2024},
  publisher={Nature Publishing Group UK London}
}

@article{corley2025accelerating,
  title={Accelerating biomolecular modeling with atomworks and rf3},
  author={Corley, Nathaniel and Mathis, Simon and Krishna, Rohith and Bauer, Magnus S and Thompson, Tuscan R and Ahern, Woody and Kazman, Maxwell W and Brent, Rafael I and Didi, Kieran and Kubaney, Andrew and others},
  journal={bioRxiv},
  doi={10.1101/2025.08.14.670328},
  year={2025}
}

@article{tradeoff,
  title={Scoring functions for protein-ligand binding affinity prediction using structure-based deep learning: a review},
  author={Meli, Rocco and Morris, Garrett M and Biggin, Philip C},
  journal={Frontiers in bioinformatics},
  volume={2},
  pages={885983},
  year={2022},
  publisher={Frontiers Media SA}
}

@inproceedings{xgboost,
author = {Chen, Tianqi and Guestrin, Carlos},
title = {XGBoost: A Scalable Tree Boosting System},
year = {2016},
isbn = {9781450342322},
publisher = {Association for Computing Machinery},
address = {New York, NY, USA},
url = {https://doi.org/10.1145/2939672.2939785},
doi = {10.1145/2939672.2939785},
booktitle = {Proceedings of the 22nd ACM SIGKDD International Conference on Knowledge Discovery and Data Mining},
pages = {785–794},
numpages = {10},
location = {San Francisco, California, USA},
series = {KDD '16}
}

@article{maccs,
  title={Reoptimization of MDL keys for use in drug discovery},
  author={Durant, Joseph L and Leland, Burton A and Henry, Douglas R and Nourse, James G},
  journal={Journal of chemical information and computer sciences},
  volume={42},
  number={6},
  pages={1273--1280},
  year={2002},
  publisher={ACS Publications}
}

@article{qed,
  title={Quantifying the chemical beauty of drugs},
  author={Bickerton, G Richard and Paolini, Gaia V and Besnard, J{\'e}r{\'e}my and Muresan, Sorel and Hopkins, Andrew L},
  journal={Nature chemistry},
  volume={4},
  number={2},
  pages={90--98},
  year={2012},
  publisher={Nature Publishing Group UK London}
}

@article{pubchem,
  title={PubChem 2025 update},
  author={Kim, Sunghwan and Chen, Jie and Cheng, Tiejun and Gindulyte, Asta and He, Jia and He, Siqian and Li, Qingliang and Shoemaker, Benjamin A and Thiessen, Paul A and Yu, Bo and others},
  journal={Nucleic acids research},
  volume={53},
  number={D1},
  pages={D1516--D1525},
  year={2025},
  publisher={Oxford University Press}
}

@article{shap,
  title={A unified approach to interpreting model predictions},
  author={Lundberg, Scott M and Lee, Su-In},
  journal={Advances in neural information processing systems},
  volume={30},
  year={2017}
}

@article{richardson2023mgnify,
  title={MGnify: the microbiome sequence data analysis resource in 2023},
  author={Richardson, Lorna and Allen, Ben and Baldi, Germana and Beracochea, Martin and Bileschi, Maxwell L and Burdett, Tony and Burgin, Josephine and Caballero-P{\'e}rez, Juan and Cochrane, Guy and Colwell, Lucy J and others},
  journal={Nucleic acids research},
  volume={51},
  number={D1},
  pages={D753--D759},
  year={2023},
  publisher={Oxford University Press}
}

@article{quercetin,
  title={Disposition of quercetin in man after single oral and intravenous doses},
  author={Gugler, R and Leschik, M and Dengler, HJ},
  journal={European journal of clinical pharmacology},
  volume={9},
  number={2},
  pages={229--234},
  year={1975},
  publisher={Springer}
}

@article{esm,
  title={Evolutionary-scale prediction of atomic-level protein structure with a language model},
  author={Lin, Zeming and Akin, Halil and Rao, Roshan and Hie, Brian and Zhu, Zhongkai and Lu, Wenting and Smetanin, Nikita and Verkuil, Robert and Kabeli, Ori and Shmueli, Yaniv and others},
  journal={Science},
  volume={379},
  number={6637},
  pages={1123--1130},
  year={2023},
  publisher={American Association for the Advancement of Science}
}

@article{test-time,
  title={Scaling llm test-time compute optimally can be more effective than scaling model parameters},
  author={Snell, Charlie and Lee, Jaehoon and Xu, Kelvin and Kumar, Aviral},
  journal={arXiv preprint arXiv:2408.03314},
  year={2024}
}

@article{deepseekr1,
  title={DeepSeek-R1 incentivizes reasoning in LLMs through reinforcement learning},
  author={Guo, Daya and Yang, Dejian and Zhang, Haowei and Song, Junxiao and Wang, Peiyi and Zhu, Qihao and Xu, Runxin and Zhang, Ruoyu and Ma, Shirong and Bi, Xiao and others},
  journal={Nature},
  volume={645},
  number={8081},
  pages={633--638},
  year={2025},
  publisher={Nature Publishing Group UK London}
}

@article{openai_o1,
  title={Openai o1 system card},
  author={Jaech, Aaron and Kalai, Adam and Lerer, Adam and Richardson, Adam and El-Kishky, Ahmed and Low, Aiden and Helyar, Alec and Madry, Aleksander and Beutel, Alex and Carney, Alex and others},
  journal={arXiv preprint arXiv:2412.16720},
  year={2024}
}

@inproceedings{opro,
 author = {Yang, Chengrun and Wang, Xuezhi and Lu, Yifeng and Liu, Hanxiao and Le, Quoc V and Zhou, Denny and Chen, Xinyun},
 booktitle = {International Conference on Learning Representations},
 editor = {B. Kim and Y. Yue and S. Chaudhuri and K. Fragkiadaki and M. Khan and Y. Sun},
 pages = {12028--12068},
 title = {Large Language Models as Optimizers},
 url = {https://proceedings.iclr.cc/paper_files/paper/2024/file/3339f19c5fcee3ad74502947a32be9e6-Paper-Conference.pdf},
 volume = {2024},
 year = {2024}
}

@inproceedings{
bopro,
title={Searching for Optimal Solutions with {LLM}s via Bayesian Optimization},
author={Dhruv Agarwal and Manoj Ghuhan Arivazhagan and Rajarshi Das and Sandesh Swamy and Sopan Khosla and Rashmi Gangadharaiah},
booktitle={The Thirteenth International Conference on Learning Representations},
year={2025},
url={https://openreview.net/forum?id=aVfDrl7xDV}
}

@inproceedings{genetic_gfn,
 author = {Kim, Hyeonah and Kim, Minsu and Choi, Sanghyeok and Park, Jinkyoo},
 booktitle = {Advances in Neural Information Processing Systems},
 doi = {10.52202/079017-1350},
 editor = {A. Globerson and L. Mackey and D. Belgrave and A. Fan and U. Paquet and J. Tomczak and C. Zhang},
 pages = {42618--42648},
 publisher = {Curran Associates, Inc.},
 title = {Genetic-guided GFlowNets for Sample Efficient Molecular Optimization},
 url = {https://proceedings.neurips.cc/paper_files/paper/2024/file/4b25c000967af9036fb9b207b198a626-Paper-Conference.pdf},
 volume = {37},
 year = {2024}
}
\bibliographystyle{plain}

%%%%%%%%%%%%%%%%%%%%%%%%%%%%%%%%%%%%%%%%%%%%%%%%%%%%%%%%%%%%%%%%%%%%%%%
\newpage
\appendix
\renewcommand{\thefigure}{\thesection\arabic{figure}}
\renewcommand{\thetable}{\thesection\arabic{table}}
\setcounter{figure}{0}

%\resetlinenumber  
\section*{\huge Supplementary Information}
\bigskip
\setcounter{page}{1}

\section{Implementation details}
\label{sec:appendix:impl}

\paragraph{Agent architecture.}
We implement the SEISMO molecule-optimisation agent as a cyclic workflow using LangGraph \cite{langgraph}. The workflow consists of five nodes: generation, parsing, validation, prediction, and finalisation. The conversation history begins with a system prompt providing general instructions (Appendix~\ref{sec:appendix:examples}) followed by either a task-specific first message describing the optimisation objective or a generic message to optimise an undisclosed score. At each iteration the generation node invokes an LLM with this full history and instructs it to output structured JSON containing a SMILES string \cite{smiles} and rationale. The parsing node extracts the JSON; the validation node uses RDKit \cite{rdkit} to verify SMILES validity. Invalid outputs trigger a re-generation with corrective feedback and are counted as iterations to avoid infinite loops. Valid molecules proceed to the prediction node, which calls a pluggable oracle returning a score and optionally an explanation. Feedback (score, iteration count, explanation-derived insights) is appended to the conversation history. The loop terminates upon reaching a target score or an iteration budget. The finalisation node then generates a natural-language summary of the trajectory grounded in the structured per-step records and oracle scores, highlighting which structural modifications improved or degraded performance.

\paragraph{PMO benchmark protocol.}
To run our agent on PMO tasks we call the benchmark oracles via the TDC package \cite{tdc}, following the same oracle-access pattern used in the original PMO setup. The agent receives only the scalar oracle score without further explanation, matching prior work. We test two variants: in the \emph{task described} variant the initial message explicitly states the objective (descriptions are comparable to those used by MolLEO and LICO \cite{molleo,lico}; exemplary full prompts in Appendix~\ref{sec:appendix:examples}); in the \emph{no description} variant the initial message omits any task description and the agent treats the oracle as fully black box. For each of the 23 PMO tasks we run both variants for three independent repetitions, capped at 50 LLM calls per run due to API cost.

We follow PMO's recommended sample-efficiency evaluation via AUC over the top-$n$ optimisation curve. We focus on $n=1$ (top-1) as we are interested in finding a high-scoring molecule rather than a batch, and additionally report top-10 AUC for direct comparison with Nguyen and Grover \cite{lico}. SEISMO is allowed to propose molecules for the first 50 oracle calls; for calls 51--10{,}000 we keep its best-so-far score fixed, while baselines optimise over the full 10{,}000 calls. Baseline trajectories are obtained from the released per-iteration data and our metrics are recomputed from those raw traces to ensure identical metric definitions. For additional methods from Nguyen and Grover, we use their published top-10 AUC values directly because per-iteration SMILES/oracle traces are not released.

\paragraph{Hit-identification proxy.}
The objective is to minimise predicted $\mathrm{IC}_{50}$ against SARS-CoV-2 M$^\mathrm{pro}$ while satisfying $\mathrm{QED}\ge 0.6$ and a novelty constraint (absence from PubChem). The oracle returns a scalar optimisation score and a structured breakdown with predicted $\mathrm{IC}_{50}$ and QED. Sample efficiency is evaluated using the optimisation curve of the lowest predicted $\mathrm{IC}_{50}$ among constraint-satisfying molecules, with the curve value set to $1\,\mathrm{M}$ when no constraint-satisfying molecule has been found yet; AUC is computed over 50 oracle calls. In the task description we refer only to a generic protease target, as explicitly naming SARS-CoV-2 M$^\mathrm{pro}$ triggered LLM safety-filter refusals. We think that it is vital to have LLM safety filters in place, but especially since we showed that giving more information to the model enhances sample efficiency and generates more favored molecules, we see that the safety regulations of proprietary LLMs can limit research and drug discovery success. Running a local open weights LLM could be a good approach for the future, especially as these recently further diminished the gap on benchmarks to closed LLMs and allow additional finetuning and training on domain specific data.

\begin{figure}[t]
  \centering
  \includegraphics[width=\linewidth]{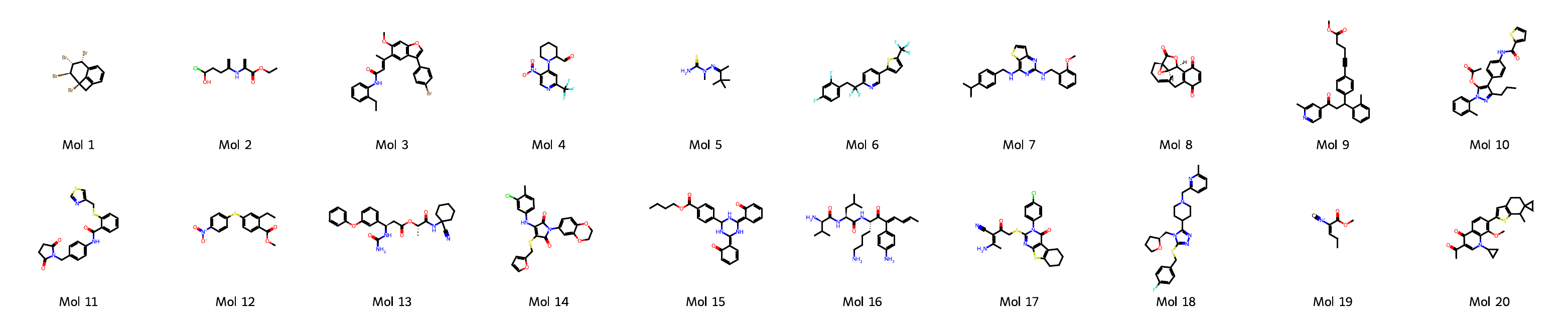}
  \caption{Visualisation of the 20 randomly sampled molecules from PubChem used as reference targets for the lead-optimisation proxy.}
  \label{fig:appendix-random-mols}
\end{figure}

\paragraph{Lead-optimisation proxy.}
Given a reference molecule $x^\star$, the objective is $s=\tfrac12(\mathrm{sim}+\mathrm{QED})$, where similarity is Tanimoto similarity over MACCS-key fingerprints \cite{maccs} and QED is computed via RDKit \cite{rdkit}. Two regimes: (i) Quercetin as $x^\star$, with metric the number of oracle calls to first reach $s\ge 0.8$; (ii) 20 PubChem references sampled uniformly under constraints designed to ensure diversity and a non-trivial QED-improvement problem --- molecular weight $<500$, $\mathrm{QED}\le 0.5$, and pairwise Tanimoto similarity $\le 0.2$ against all previously selected targets (Figure~\ref{fig:appendix-random-mols}). Metric for (ii) is top-1 AUC over 50 iterations averaged across the 20 targets.

\paragraph{Oracle explanations.}
For the XGBoost $\mathrm{IC}_{50}$ model, explanations are generated with SHAP \cite{shap} over MACCS keys. We present the 10 substructures with highest attribution (driving lower $\mathrm{IC}_{50}$) and the 10 with lowest attribution (driving higher $\mathrm{IC}_{50}$), each annotated with its presence in the current molecule and SMARTS pattern, to guide structural modifications. For QED, we provide a property-level decomposition (molecular weight, logP, HBA, HBD, PSA, rotatable bonds, aromatic-ring count, alert count) and indicate which properties most reduced QED for the current proposal and what ranges would be more favourable. For similarity, we explain which MACCS keys differ between $x^\star$ and the proposal (keys present in exactly one of the two) and map these keys to SMARTS patterns. See Appendix~\ref{sec:appendix:examples} for examples.

\paragraph{Evaluation protocol for medicinal-chemistry tasks.}
For the SARS-CoV-2 M$^\mathrm{pro}$ task we ran ten, and for the Quercetin lead-optimisation task we run five independent repetitions for each of the four information conditions, up to 50 iterations. For the 20 random PubChem targets we run three 50-iteration optimisation per target and information condition and report results averaged over the 20 targets. 

\paragraph{XGBoost training.}

The XGBoost regression model was implemented using the \texttt{xgboost} python package (version 3.4.2) and trained to predict pIC$_{50}$ values for SARS-CoV-2 M$^\mathrm{pro}$ inhibition using the ASAP Discovery antiviral-potency-2025-unblinded dataset accessed via the Polaris platform \cite{polaris}. Molecules were encoded as 167-bit MACCS fingerprints via RDKit. The XGBRegressor was configured with 2{,}000 estimators, a learning rate of 0.03, a subsample ratio of 0.8, $\alpha=0.1$, $\lambda=2$, histogram-based tree construction with a loss-guided growth policy, and a maximum of 128 leaves per tree. The four remaining hyperparameters were selected by random search over 40 distinct draws from a grid of maximum tree depth $\in\{3,4,5,6\}$, minimum child weight $\in\{1,3,5,7\}$, gamma $\in\{0,0.1,0.5,1.0\}$ and column subsampling ratio $\in\{0.4,0.6,0.8,1.0\}$, scored by five-fold cross-validation on the training split alone under a fixed seed of 42. The configuration with the lowest mean cross-validated RMSE used a maximum tree depth of 3, a minimum child weight of 5, a gamma of 0.1 and a column subsampling ratio of 0.4, reaching a cross-validated RMSE of $0.63 \pm 0.04$. The held-out split was evaluated once, after this configuration was fixed, giving an RMSE of 0.82, an MAE of 0.63 and an $R^2$ of 0.63.

\paragraph{Co-folding oracle.}
The Boltz-2 oracle takes the target protein sequence and the SMILES proposed by the LLM as inputs. We invoke Boltz-2 with default parameters with two exceptions: predicted structures are stored in PDB format, and multiple sequence alignments are not pre-computed (we use Boltz-2's default MSA server). We run the model on an NVIDIA A100 GPU with a five-minute time limit; on timeout or non-zero exit code the oracle returns NaN. On success, the oracle extracts the binding-affinity probability score (\texttt{affinity\_probability\_binary}), a sigmoid probability representing Boltz-2's belief that the compound binds. Boltz-2 also returns a binding-affinity value (\texttt{affinity\_pred\_value}) targeting the $\mathrm{IC}_{50}$ of the complex, but preliminary experiments showed the agent had difficulty optimising it, so we use the probability.

As an explanation in full-explanation mode, the oracle reports Boltz-2 confidence scores (\texttt{confidence\_score}, \texttt{[ligand\_$\mid$i]ptm}, \texttt{complex\_[i]plddt}, \texttt{complex\_[i]pde}) and the protein's close residues, defined as residues with at least one non-hydrogen atom within 5\,\AA{} of a compound non-hydrogen atom. Distances are computed with biotite \cite{kunzmann2023biotite} and scikit-learn \cite{scikit-learn}; the three-letter code and index of all qualifying residues is reported. If any of these computations fail, an empty explanation is reported.

For MGYP001550541752, the average pLDDT score obtained when predicting the structure of the isolated protein was 0.870 with ESMFold and 0.732 with Boltz-2. Across the full explanation runs, the mean Boltz-2 confidence score was 0.799. This aggregated score is defined as $0.8 *$complex pLDDT$+ 0.2 *$ipTM.

\section{LLM backbone ablation}
\label{sec:appendix:llm}

To evaluate the base performance we did an ablation study over multiple LLM backbones on the 20-random-target Sim--QED task. Tested: Claude Opus 4.5 (\texttt{claude-opus-4-5-20251101-v1:0}), Claude Sonnet 4.5 (\texttt{claude-sonnet-4-5-20250929-v1:0}), chatGPT 4.1 (\texttt{gpt-4.1-2025-04-14}), chatGPT 5.1 (\texttt{gpt-5.1-2025-11-13} without reasoning or medium reasoning effort). DeepSeek-R1 was evaluated but excluded because it frequently failed to produce syntactically valid SMILES under our JSON-constrained output format. 

We swap only the LLM, keeping prompts, JSON schema, retry/validation logic, oracle, and the 50-iteration budget fixed, and compare models by mean top-1 AUC over 50 iterations averaged across the 20 targets for $s=\tfrac12(\mathrm{sim}+\mathrm{QED})$. GPT-5.1 is run only at medium reasoning effort because higher reasoning effort produced impractical per-call latency at our experimental scale.

\begin{figure}[t]
  \centering
  \includegraphics[width=0.7\linewidth]{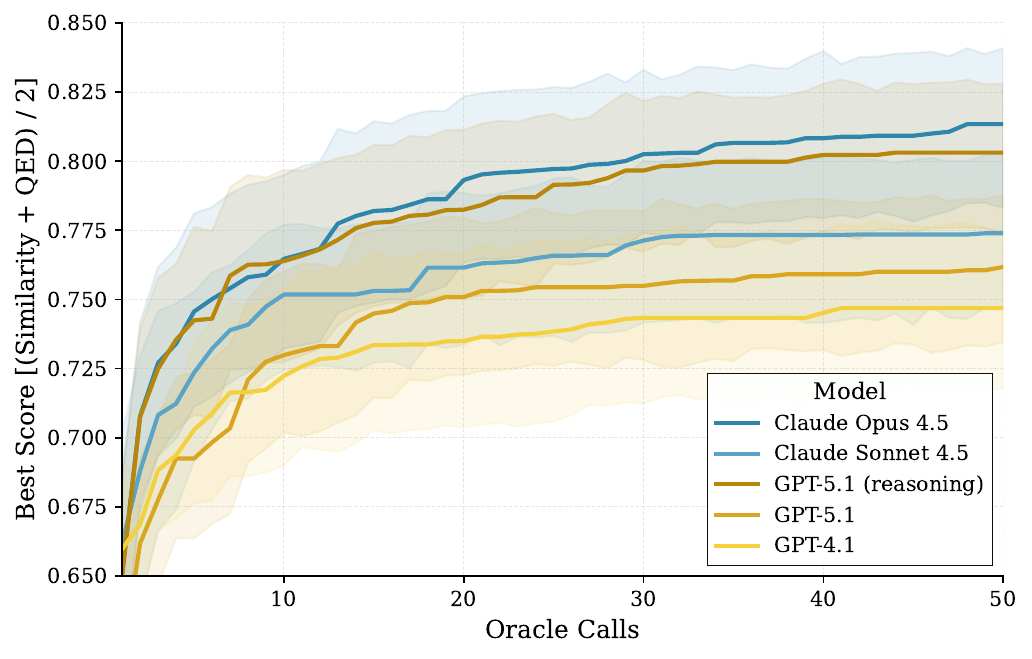}
  \caption{Comparison of multiple LLM backbones on the 20-random-target Sim--QED task. Lines and bands denote mean and standard deviation across targets.}
  \label{fig:appendix-llm-ablation}
\end{figure}

Averaged across targets, newer models achieved higher performance, with Claude Opus 4.5 best, closely followed by GPT-5.1 at medium reasoning effort (Figure~\ref{fig:appendix-llm-ablation}). In addition to its slightly better score, Opus 4.5 exhibited faster response times and lower per-iteration cost than GPT-5.1 at medium reasoning under our API provider's pricing and latency at experimentation time. Opus 4.5 outperformed Sonnet 4.5 and GPT-5.1 outperformed GPT-4.1, indicating a correlation between general model capability and molecular-optimisation performance. These results motivated our selection of Opus 4.5 for the main experiments.

Beyond the performance of the LLM itself, there is ongoing work on optimising compute at both train time and test time \cite{test-time}. Open weight models can be optimised in both regimes, whereas closed proprietary models are largely limited to test time optimisation. In molecular optimisation, however, a third compute component becomes important, namely scoring cost, which often dominates when the oracle is expensive. Improving the LLM and its surrounding framework can still indirectly reduce this cost, because better generated molecules reach target thresholds with fewer oracle calls. In practice, this means evaluating the baseline performance of currently available LLMs, potentially preferring open weight models when performance is comparable while weighing usability, cost, and sustainability parameters such as water consumption and CO$_2$ emissions, optimising the surrounding framework as done here for SEISMO, and keeping in mind the true bottleneck of the overall optimisation task.

\section{Truncated history}
\label{sec:appendix:history}

To evaluate the effect of the length and composition of the conversation history on the optimisation, we asked whether the LLM still reaches the performance of the full history (all previous molecules, scores and explanations) when only a short or selected part of it is visible. In this truncated setting, the system prompt and the task description are always retained, but only a subset of the previous iterations is kept in the prompt. Each iteration is kept or removed as a whole (the model response together with its feedback). Retained iterations keep their complete feedback including the explanations of all oracles, whereas removed iterations contribute no molecule, score or explanation. We evaluate two selection rules among the valid molecules proposed so far: last-k retains the k most recent iterations, and best/worst-k retains the k iterations with the highest and the k with the lowest combined score, re-evaluated before every generation step.

As a consequence, in best/worst-k the feedback of the most recent molecule is not necessarily shown, namely when it is neither among the k best nor among the k worst. In this case the prompt could remain identical to that of the previous step, which the LLM API may reject, and we therefore append the current iteration number to the prompt. If the most recent generation was invalid (e.g., an unparsable SMILES string), the corresponding error message is shown, so that the model can correct it. We used k = 2 and 10 for last-k and k = 1 and 5 for best/worst-k, so that a last-k and a best/worst-k setting contain the same maximum number of iterations (k and 2k, respectively). Each variant was run in three repetitions on each of the 20 PubChem molecules of the Similarity + QED benchmark, with the same model, temperature and prompts as the full-history runs.

\begin{figure}[h!]
\centering
\includegraphics[width=0.72\textwidth]{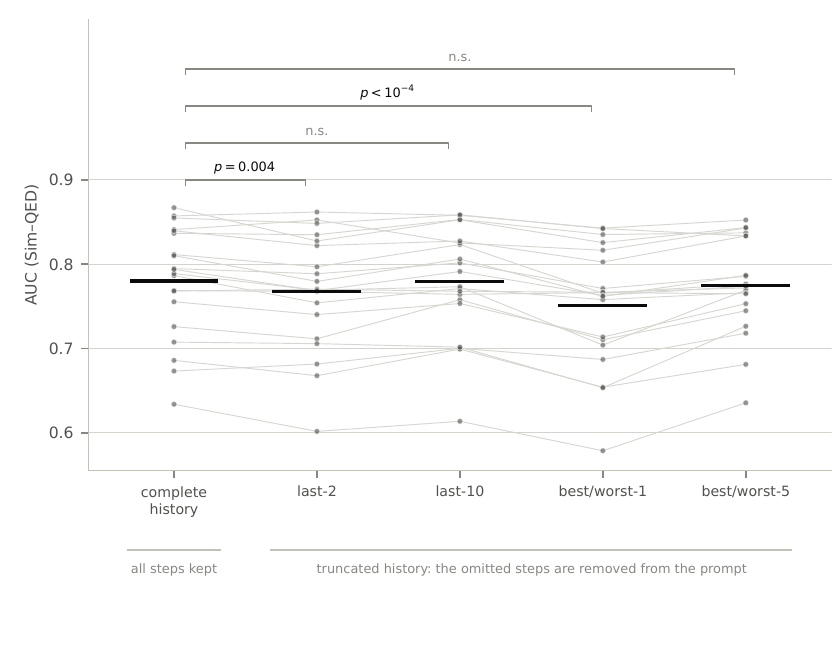}
\caption{%
Effect of truncating the optimisation history on Sim--QED. In the truncated
settings the omitted steps are removed from the prompt entirely, under
otherwise identical model and sampling settings, with two selection rules: the
last $k$ steps, and the $k$ best together with the $k$ worst. Each point is the
mean over the 3 replicates for one of the 20 reference molecules and a light
line joins the same molecule across settings. The black bar is the mean AUC over
the molecules (higher is better). Brackets carry the adjusted $p$-values and \emph{n.s.} denotes
$p>0.05$. 
}
\label{fig:appendix:history-modes}
\end{figure}

Evaluated by the top-1 AUC over 50 oracle calls, averaged over the 20 molecules (mean $\pm$ 95\% CI across molecules), the truncated variants with the larger windows are not statistically significantly different from the full history (0.780 $\pm$ 0.031): 0.780 $\pm$ 0.030 for last-k with k = 10 and 0.775 $\pm$ 0.027 for best/worst-k with k = 5 (see Figure \ref{fig:appendix:history-modes}). With the smallest windows the results are significantly lower than for the full history, with 0.767 $\pm$ 0.032 for last-k with k = 2 and 0.751 $\pm$ 0.033 for best/worst-k with k = 1, the latter being close to the result of the condition that shows the scores of all iterations but no explanations (0.756 $\pm$ 0.028). The truncation of the history also strongly reduces the prompt length and thereby the costs (see Appendix \ref{sec:appendix:cost-runtime}), which makes truncated histories with a sufficiently large k an attractive option. However, we note that the top-1 AUC metric is strongly influenced by performance in the initial iterations, especially for the SIM--QED task (Figure~\ref{fig:sim_qed_zeroshot}). As the truncated variants do not differ from the full-explanation mode at the beginning of the optimization, further work using more diverse tasks and metrics is needed to assess their influence on performance more comprehensively.

In summary, the top-1 AUC performance of the full history can be retained with a considerably shorter prompt, provided that the window is large enough (here k = 10 for last-k and k = 5 for best/worst-k), whereas very small windows lose performance.

\section{Cost and runtime analysis}
\label{sec:appendix:cost-runtime}

The token usage and the corresponding cost per run can be estimated as follows.

\begin{equation}
T_{\text{prompt},k} \approx T_1 + (k-1)\Delta T
\end{equation}

\begin{equation}
T_{\text{prompt,total}}(n)
=
\sum_{k=1}^{n} T_{\text{prompt},k}
=
nT_1 + \Delta T \frac{n(n-1)}{2}
\end{equation}

\begin{equation}
C(n)
=
\left(
\frac{T_{\text{prompt,total}}(n) + T_{\text{sum,in}}}{10^6}
\right) P_{\text{in}}
+
\left(
\frac{nT_{\text{out}}^{\text{mean}} + T_{\text{sum,out}}}{10^6}
\right) P_{\text{out}}
\end{equation}

where $T_{\text{prompt},k}$ denotes the number of input tokens in the $k$-th prompt, $T_1$ is the number of input tokens in the initial prompt, consisting of the system prompt and the first user message containing either the task description or a generic message, and $\Delta T$ is the mean increase in input tokens per iteration. Furthermore, $T_{\text{prompt,total}}(n)$ denotes the total number of prompt tokens accumulated over $n$ iterations. The quantity $T_{\text{out}}^{\text{mean}}$ denotes the mean number of output tokens generated per iteration, while $T_{\text{sum,in}}$ and $T_{\text{sum,out}}$ denote the numbers of input and output tokens used for the summarization step, respectively. Moreover, $C(n)$ denotes the total cost in USD after $n$ iterations, and $P_{\text{in}}$ and $P_{\text{out}}$ denote the prices per million input and output tokens, respectively, taken here as USD~5 and USD~20. Finally, $n_{\max}$ denotes the maximum number of iterations that can be performed before reaching Claude Opus 4.5 200k-token context window.

We logged token usage for the $IC_{50}$-QED-Novelty task, and Table~\ref{tab:appendix-token-cost} shows that token usage and cost depend on the explanation mode, with longer feedback resulting in higher costs. Cost estimates, $n_{\max}$, and $C(n_{\max})$ are approximate.

\begin{table}[ht]
\centering
\caption{Token usage and cost estimates for the four SEISMO modes on the $IC_{50}$-QED-Novelty task using Claude Opus 4.5. Costs, denoted by $C(n)$, are reported in USD and are estimated assuming prices of USD~5 and USD~20 per million input and output tokens, respectively. Uncertainties are 95\% confidence-interval half-widths across 10 runs per mode.}
\begin{tabular}{l|c|c|c|c}
\textbf{Mode} & \shortstack{\textbf{Full}\\\textbf{explanation}} & \shortstack{\textbf{Partial}\\\textbf{explanation}} & \shortstack{\textbf{No}\\\textbf{explanation}} & \shortstack{\textbf{No}\\\textbf{description}} \\ \hline
$T_1$ & 403 & 403 & 403 & 300 \\ \hline
$\Delta T$ & 1371 $\pm$ 23 & 485 $\pm$ 7 & 305 $\pm$ 5 & 137 $\pm$ 4 \\ \hline
$T_{\text{out}}^{\text{mean}}$ & 164 $\pm$ 10 & 106 $\pm$ 4 & 103 $\pm$ 3 & 68 $\pm$ 3 \\ \hline
$T_{\text{sum,in}}$ & 11601 $\pm$ 656 & 8960 $\pm$ 216 & 8788 $\pm$ 182 & 4776 $\pm$ 130 \\ \hline
$T_{\text{sum,out}}$ & 574 $\pm$ 31 & 586 $\pm$ 23 & 579 $\pm$ 15 & 546 $\pm$ 17 \\ \hline
$C(50)$ & 8.78 $\pm$ 0.15 & 3.27 $\pm$ 0.05 & 2.15 $\pm$ 0.04 & 1.03 $\pm$ 0.03 \\ \hline
$\approx n_{max}$ & 147 $\pm$ 3 & 415 $\pm$ 6 & 660 $\pm$ 13 & 1466 $\pm$ 44 \\ \hline
$\approx C(n_{max})$ & 73.98 $\pm$ 1.30 & 210.01 $\pm$ 3.18 & 334.03 $\pm$ 7.62 & 737.59 $\pm$ 23.57 \\
\end{tabular}
\label{tab:appendix-token-cost}
\end{table}

Additionally, token usage and cost are task-specific. For the SIM--QED task on 20 randomly selected PubChem molecules, full-history conditioning resulted in an estimated $C(50)$ of USD~$5.13 \pm 0.11$, with a mean input-token increase of $776 \pm 18$ tokens per iteration. History truncation, retaining only $k$ iterations including their explanations as described in Appendix~\ref{sec:appendix:history}, substantially reduced both quantities.

For policies retaining two iterations, last-$k$ ($k=2$) and best/worst-$k$ ($k=1$) yielded estimated costs of USD~$0.61 \pm 0.01$ and USD~$0.57 \pm 0.01$, respectively, corresponding to reductions of approximately 88\% relative to full-history conditioning. Their mean input-token increases were $35 \pm 1$ and $32 \pm 2$ tokens per iteration, respectively. For policies retaining ten iterations, last-$k$ ($k=10$) and best/worst-$k$ ($k=5$) both yielded an estimated $C(50)$ of USD~$1.38 \pm 0.04$, corresponding to a reduction of approximately 73\%. Their mean input-token increases were $163 \pm 7$ and $163 \pm 6$ tokens per iteration, respectively.

% add token consumption of Boltz-2 (one mode or all modes)? Only "results" of n_max and c(n_max)?

We additionally logged the wall clock LLM time as the elapsed time between each request and response. For the $IC_{50}$-QED Novelty task, the mean LLM inference time was 6.7\,s per iteration. As shown in Figure \ref{fig:extended-run}, the LLM wall clock time did not increase consistently at higher iteration counts despite the longer context length. The figure also shows that molecules with lower, and thus more favorable, $IC_{50}$ values were generated in later iterations. LLM API response times additionally depend on choosing the right timing, as the time of day could directly influences both user request volume and resource availability. The longer response times seen in the far-right outliers may be due to unfavorable conditions in these factors. Still, in our long horizon experiment the LLM inference time remained approximately constant and a run up to the maximum number of iterations for a single instance, namely 133 iterations, was completed in approximately 16 minutes of wall clock time in the full explanation mode. Considering time alone as a resource, we can perform a simple break even analysis to identify the optimization regimes in which the use of SEISMO is advantageous.

Neglecting the model-training and population-management overhead of classical baselines and counting only their oracle wall-clock time, SEISMO is faster than a baseline using $N\times$ more oracle calls whenever the per-oracle wall-clock cost $T_\mathrm{oracle}$ satisfies $T_\mathrm{oracle} > T_\mathrm{LLM}/(N-1)$, where $T_\mathrm{LLM}=6.7$\,s is the mean LLM inference time per iteration in our experiments. Under the conservative choice $N=2$ (in practice classical baselines typically need $20\times$--$200\times$ more oracle calls than SEISMO in our regime), break-even oracle cost is approximately 7\,s per call. A single Boltz-2 co-folding evaluation takes minutes; a physics-based binding-affinity calculation takes minutes to hours; an experimental assay takes days. All of these clear the break-even threshold by orders of magnitude.

\FloatBarrier

\section{Pairwise significance on the medicinal chemistry tasks}
\label{sec:appendix:pairwise-significance}

\begin{figure}[h!]
\centering
\includegraphics[width=\textwidth]{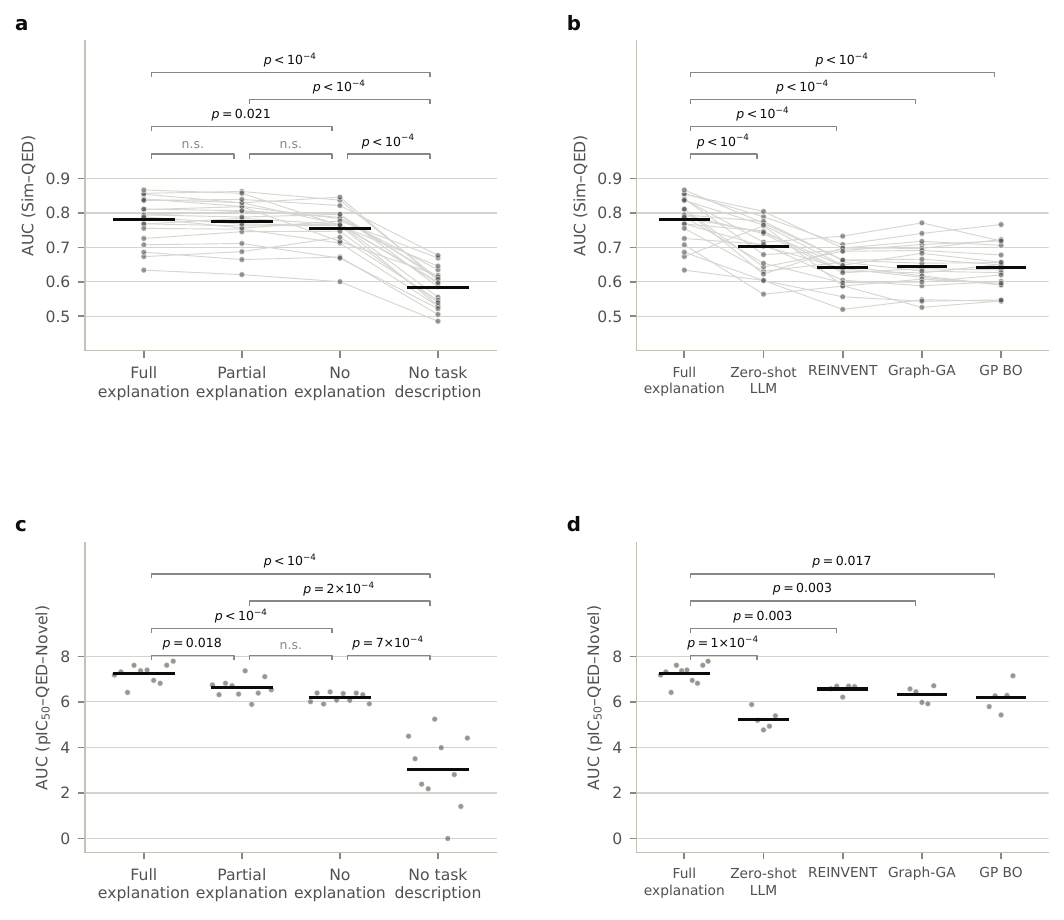}
\caption{Pairwise significance on the two medicinal-chemistry tasks. The left column compares the four information levels of SEISMO, with every pair tested. The right column compares SEISMO with full explanations against each alternative optimisation method. \textbf{(a,~b)}~Sim--QED, where each point is the AUC mean over the replicates for one of the 20 reference molecules. \textbf{(c,~d)}~pIC$_{50}$--QED--Novel, where each point is the AUC of one of the 10 replicates. The black bar is the variant mean. Brackets carry adjusted $p$-values, and \emph{n.s.} denotes $p>0.05$.}
\label{fig:appendix:pairwise-significance}
\end{figure}

\FloatBarrier

\section{Iso-quality budget for the classical baselines}
\label{sec:appendix:iso-quality}

Table~\ref{tab:appendix-iso-quality} reports how many oracle calls each classical baseline needs before it reaches the quality SEISMO attains within 50 calls. Against a measured 5.13~USD and 335~s of LLM inference per SEISMO run on this task, the three methods break even at 0.0012 to 0.0015~USD per oracle call, or 0.08 to 0.10~s per call. An oracle more expensive than that favours SEISMO outright, which covers most wet-lab assays and the slower structure-based predictors.

% Generated by scripts/baselines/iso_quality.py, do not edit by hand.
\begin{table}[ht]
\centering
\caption{Oracle calls each baseline needs before its best-so-far molecule matches the mean best score SEISMO reached within 50 calls on that reference, averaged over its three replicates, on the Sim--QED task. Baselines run at their upstream defaults with three replicates on each of the 20 references to a horizon of 10,000 oracle calls. Mean calls to match is taken over the runs that did match.}
\label{tab:appendix-iso-quality}
\begin{tabular}{lcc}
\toprule
Method & Matched & Mean calls\\
 & (of runs) & to match\\
\midrule
Graph-GA & 41/60 & 4,040\\
REINVENT & 46/60 & 4,453\\
GP BO & 44/60 & 4,154\\
\bottomrule
\end{tabular}
\end{table}

\FloatBarrier

\section{Baseline hyperparameter settings}
\label{sec:appendix:baseline-configs}

Table~\ref{tab:baseline-hyperparameters} reports each classical baseline across
its hyperparameter settings at the 50 oracle call budget, together with the
number of calls each configuration consumes before it takes a single
optimisation step. We additionally include a sequential hill-climber, which
proposes one molecule per oracle call by mutating its current best with
Graph-GA's mutation operators and accepting on improvement. It therefore updates
after every evaluation and is the most budget-compatible of the classical
methods, yet it remains below SEISMO on both tasks.

% Generated by scripts/baselines/appendix_table.py -- do not edit by hand.
\begin{table}[t]
\centering
\small
\caption{Classical baseline performance across hyperparameter settings at a 50 oracle call budget. \emph{Calls to first update} is the number of oracle calls each configuration consumes before it takes a single optimisation step. Uncertainties are standard deviations over 5 runs for IC$_{50}$-QED-Novel and over the 20 reference molecules for SIM-QED.}
\label{tab:baseline-hyperparameters}
\begin{tabular}{llrc}
\toprule
Method & Configuration & Calls to first update & Top-1 AUC \\
\midrule
\multicolumn{4}{l}{\textit{IC$_{50}$-QED-Novel (pIC$_{50}$)}} \\
Graph-GA & default \textsuperscript{\dag} & 120 & $6.33 \pm 0.32$ \\
 & pop 40 /  off 20 & 40 & $6.31 \pm 0.32$ \\
 & pop 20 /  off 10 & 20 & $6.11 \pm 0.33$ \\
 & pop 10 /  off 5 & 10 & $6.17 \pm 0.31$ \\
 & pop 20 /  off 10, mut 0.2 & 20 & $6.12 \pm 0.38$ \\
GP-BO & default \textsuperscript{\dag} & 340 & $6.18 \pm 0.58$ \\
 & init 20 /  batch 1 & 20 & $6.05 \pm 0.45$ \\
 & init 10 /  batch 5 & 10 & $6.15 \pm 0.46$ \\
REINVENT & default \textsuperscript{\dag} & 64 & $6.57 \pm 0.18$ \\
 & batch 16 & 16 & $6.60 \pm 0.63$ \\
 & batch 8 & 8 & $6.37 \pm 0.44$ \\
 & batch 4 & 4 & $6.52 \pm 0.44$ \\
 & batch 8, replay 8 & 8 & $6.36 \pm 0.43$ \\
Hill-climber & mutation only & 1 & $6.58 \pm 1.10$ \\
\midrule
\multicolumn{4}{l}{\textit{SIM-QED (score)}} \\
Graph-GA & default \textsuperscript{\dag} & 120 & $0.64 \pm 0.06$ \\
 & pop 40 /  off 20 & 40 & $0.64 \pm 0.07$ \\
 & pop 20 /  off 10 & 20 & $0.64 \pm 0.07$ \\
 & pop 10 /  off 5 & 10 & $0.63 \pm 0.07$ \\
 & pop 20 /  off 10, mut 0.2 & 20 & $0.64 \pm 0.07$ \\
GP-BO & default \textsuperscript{\dag} & 340 & $0.64 \pm 0.06$ \\
 & init 20 /  batch 1 & 20 & $0.64 \pm 0.07$ \\
 & init 10 /  batch 5 & 10 & $0.63 \pm 0.07$ \\
REINVENT & default \textsuperscript{\dag} & 64 & $0.64 \pm 0.05$ \\
 & batch 16 & 16 & $0.64 \pm 0.07$ \\
 & batch 8 & 8 & $0.64 \pm 0.06$ \\
 & batch 4 & 4 & $0.64 \pm 0.06$ \\
 & batch 8, replay 8 & 8 & $0.64 \pm 0.06$ \\
Hill-climber & mutation only & 1 & $0.60 \pm 0.08$ \\
\bottomrule
\end{tabular}
\\[2pt]
\footnotesize\textsuperscript{\dag}\,mol-opt upstream default
\end{table}

\FloatBarrier

\section{Optimisation method baselines on medicinal chemistry tasks} 
\label{sec:appendix:baselines_med_chem_curves}

In this section we show the optimisation curves corresponding to the AUC values reported in Table \ref{tab:medchem-main}. 

\FloatBarrier
\begin{figure}[h!]
  \centering
  % --- Two plots side by side below ---
  \begin{subfigure}[t]{0.48\textwidth}
    \includegraphics[width=\linewidth]{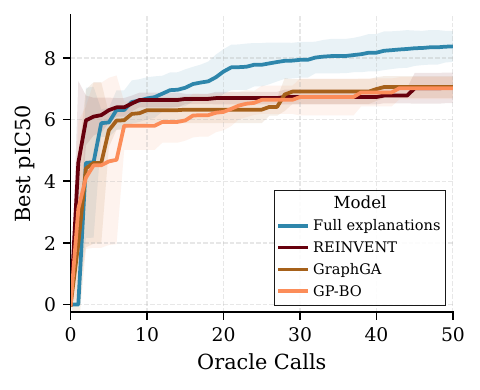}
    \caption{Comparison of SEISMO (full explanation) with the three classical optimisation baselines REINVENT, Graph-GA, and GP BO for the IC$_{50}$-QED-Novel task. Lines and bands denote the mean and standard deviation over 10 independent runs for SEISMO and 5 independent runs for each classical baseline.}
    \label{fig:appendix-ic50_baselines}
  \end{subfigure}\hfill
  \begin{subfigure}[t]{0.48\textwidth}
    \includegraphics[width=\linewidth]{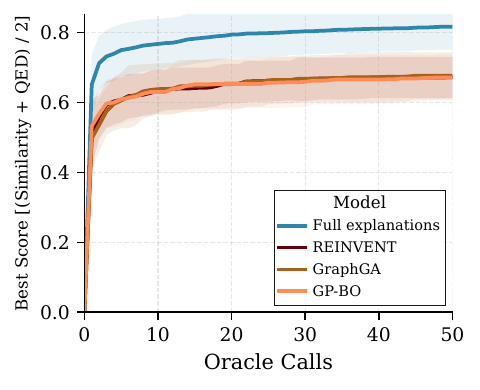}
    \caption{Comparison of SEISMO (full explanation) with the three classical optimisation baselines REINVENT, Graph-GA, and GP BO for the lead-optimisation task. All methods were run on the same 20 reference molecules, with 3 independent replicates per molecule for SEISMO. Lines and bands denote the mean and standard deviation across the 20 reference molecules.}
    \label{fig:appendix-sim_qed_random_baselines}
  \end{subfigure}

  \caption{Best-so-far optimisation curves of the classical optimisation model baseline for the medicinal-chemistry settings.}
  \label{fig:appendix-classical baselines}
\end{figure}

\FloatBarrier

\section{Further co-folding results}
\label{sec:appendix:cofolding-extra}

To validate the co-folding-based oracle, we performed an ablation study on a well-studied target (SARS-CoV-2 M$^\mathrm{pro}$) by perturbing its inputs.

\begin{figure}[t]
  \centering
  \includegraphics[width=0.5\linewidth]{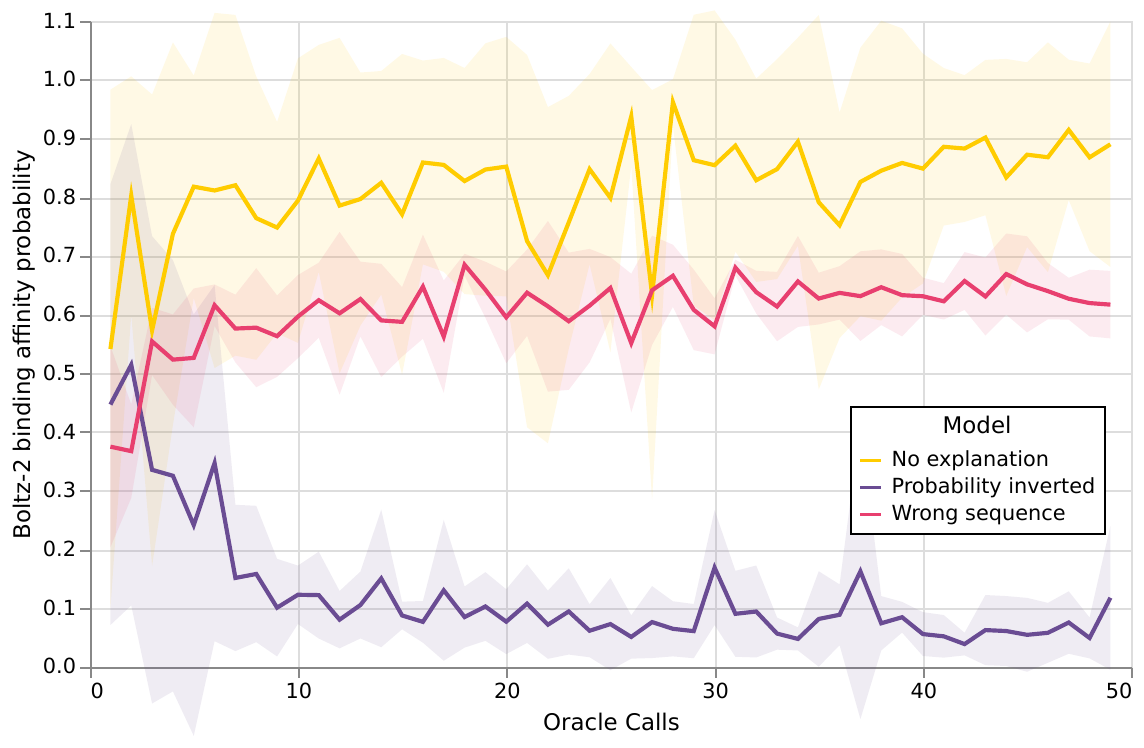}
  \caption{Oracle ablations on SARS-CoV-2 M$^\mathrm{pro}$. ``Wrong sequence'' replaces the protein sequence supplied to the LLM with GFP; ``probability inverted'' inverts the Boltz-2 affinity probability returned to the LLM. Mean and standard deviation over five repeats per condition.}
  \label{fig:cofolding-ablations}
\end{figure}

To understand how the LLM weighs its own chemical priors against external oracle signals, we perform two ablations using SARS-CoV-2 M$^\mathrm{pro}$ as the target. (i) ``Wrong sequence'': the green fluorescent protein sequence is provided to the LLM as a decoy while the oracle continues to score against the true M$^\mathrm{pro}$ structure. (ii) ``Probability inverted'': the correct sequence is provided but the Boltz-2 affinity probability is inverted before being returned to the LLM. Both use the no-explanation oracle setting. Figure~\ref{fig:cofolding-ablations} reports the per-iteration mean affinity probability (not the cumulative maximum) to highlight short-term agent behaviour.

In ``wrong sequence'' mode the curve still trends upward but never exceeds about 0.7, illustrating optimisation under contradictory contextual cues. In ``probability inverted'' mode the curve consistently decreases, indicating that the agent weights the oracle signal more than its internal knowledge.

\FloatBarrier

\section{Chemical-space analysis details}  % put that to implementation details above? contains no results
\label{sec:appendix:chemspace}

\begin{figure}[t]
  \centering
    \includegraphics[width=\linewidth]{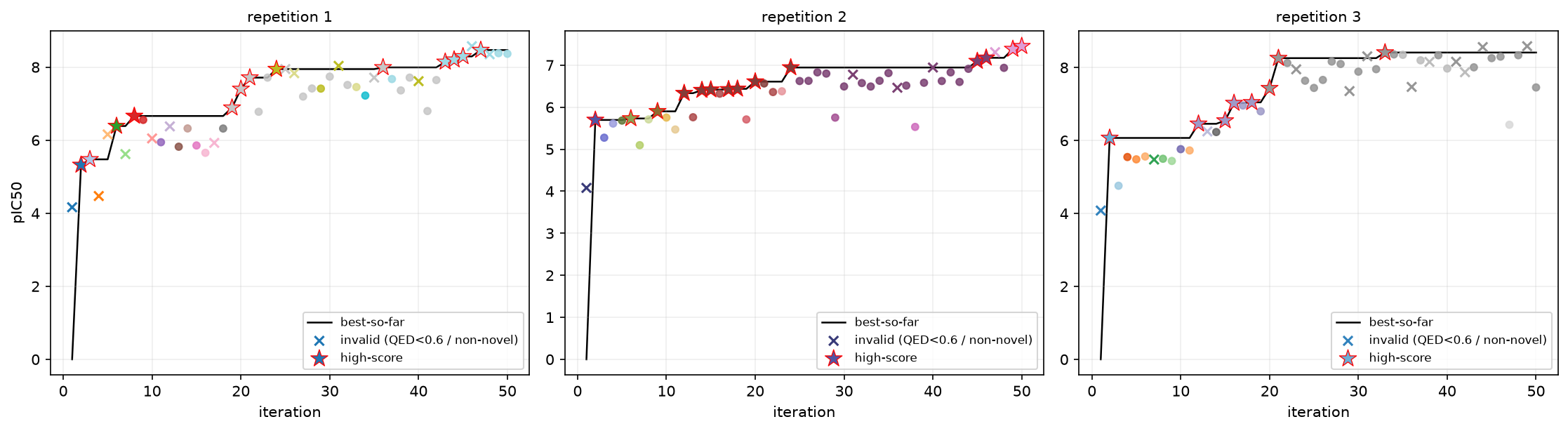}
  \caption{Exploration–exploitation trade-off of the optimisation, analysed in terms of chemical basins. Each panel shows one independent replicate of the full explanation agent on the IC$_{50}$-QED-Novel task, with molecules ordered by optimisation iteration (x-axis). The solid black line is the best-so-far potency. Validity is defined by the task's hard constraints: a molecule counts only if it is drug-like (QED $\geq 0.6$) and novel, otherwise it is disqualified (crosses). Stars mark new high-score molecules. Point colour encodes the molecule's basin. Basins are clustered per run, so colours are panel-specific and not comparable across panels.}
  \label{fig:appendix-exploration-exploitation-basins}
\end{figure}

\begin{figure}[t]
  \centering
    \includegraphics[width=\linewidth]{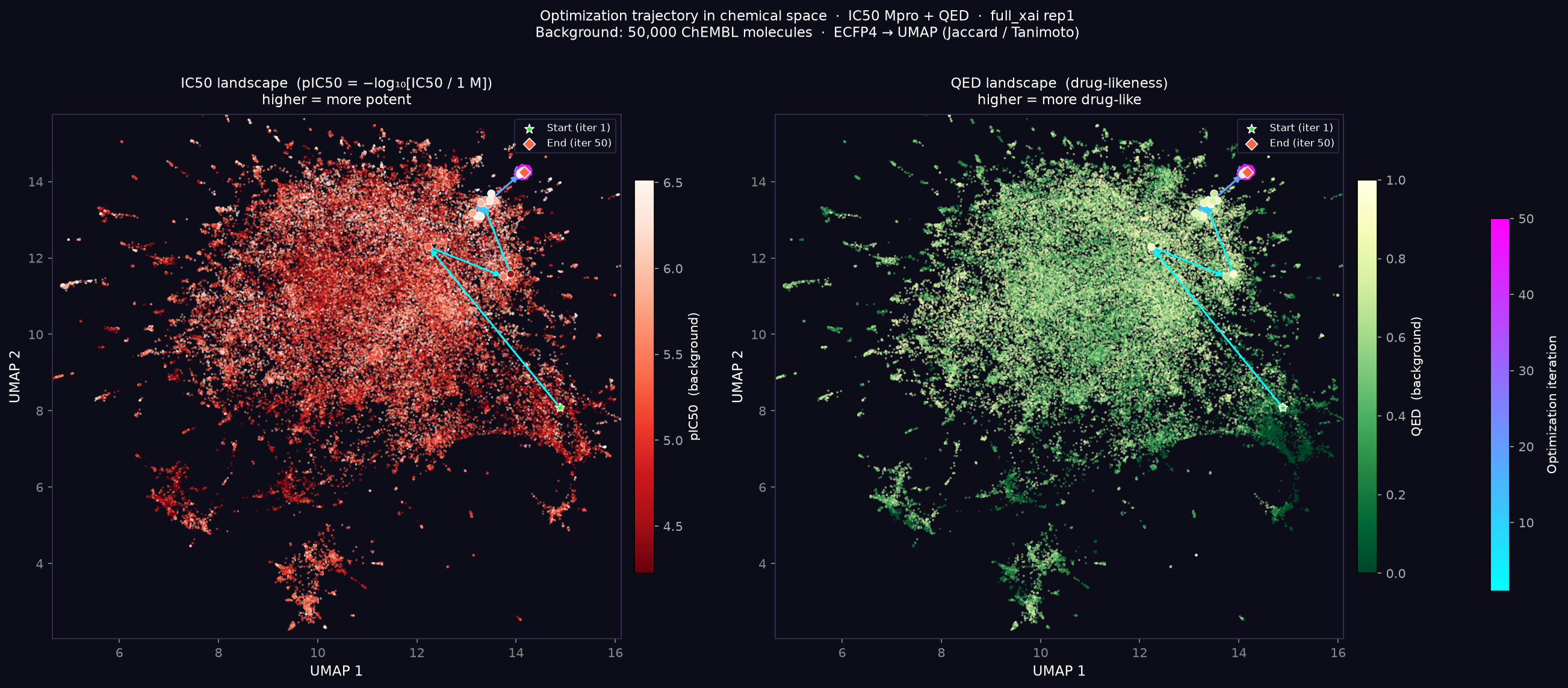}
  \caption{Molecular optimization trajectory in chemical space of the first repetition for the IC$_{50}$-QED-Novel task. Each trajectory is embedded via UMAP (ECFP4, Jaccard) with 50,000 ChEMBL molecules as chemical context. Left: Predicted IC50 landscape. Right: drug-likeness (QED) landscape. Dots represent trajectory molecules connected by directional arrows (iteration order). Stars and diamonds mark start and end points. Background coloring shows the property landscape.}
  \label{fig:appendix-exploration-exploitation-trajectory-rep1}
\end{figure}

\begin{figure}[t]
  \centering
    \includegraphics[width=\linewidth]{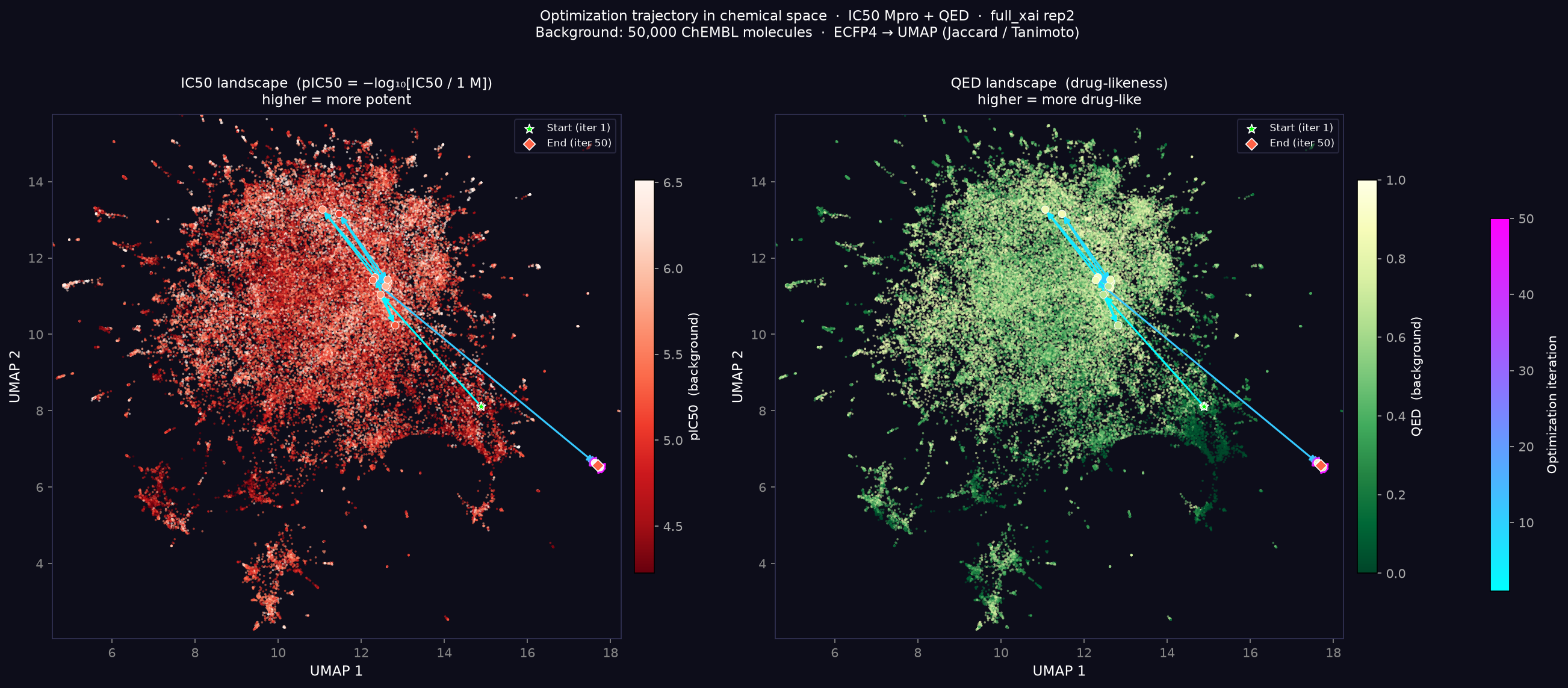}
  \caption{Molecular optimization trajectory in chemical space of the second repetition for the IC$_{50}$-QED-Novel task. Each trajectory is embedded via UMAP (ECFP4, Jaccard) with 50,000 ChEMBL molecules as chemical context. Left: Predicted IC50 landscape. Right: drug-likeness (QED) landscape. Dots represent trajectory molecules connected by directional arrows (iteration order). Stars and diamonds mark start and end points. Background coloring shows the property landscape.}
  \label{fig:appendix-exploration-exploitation-trajectory-rep2}
\end{figure}

\begin{figure}[t]
  \centering
    \includegraphics[width=\linewidth]{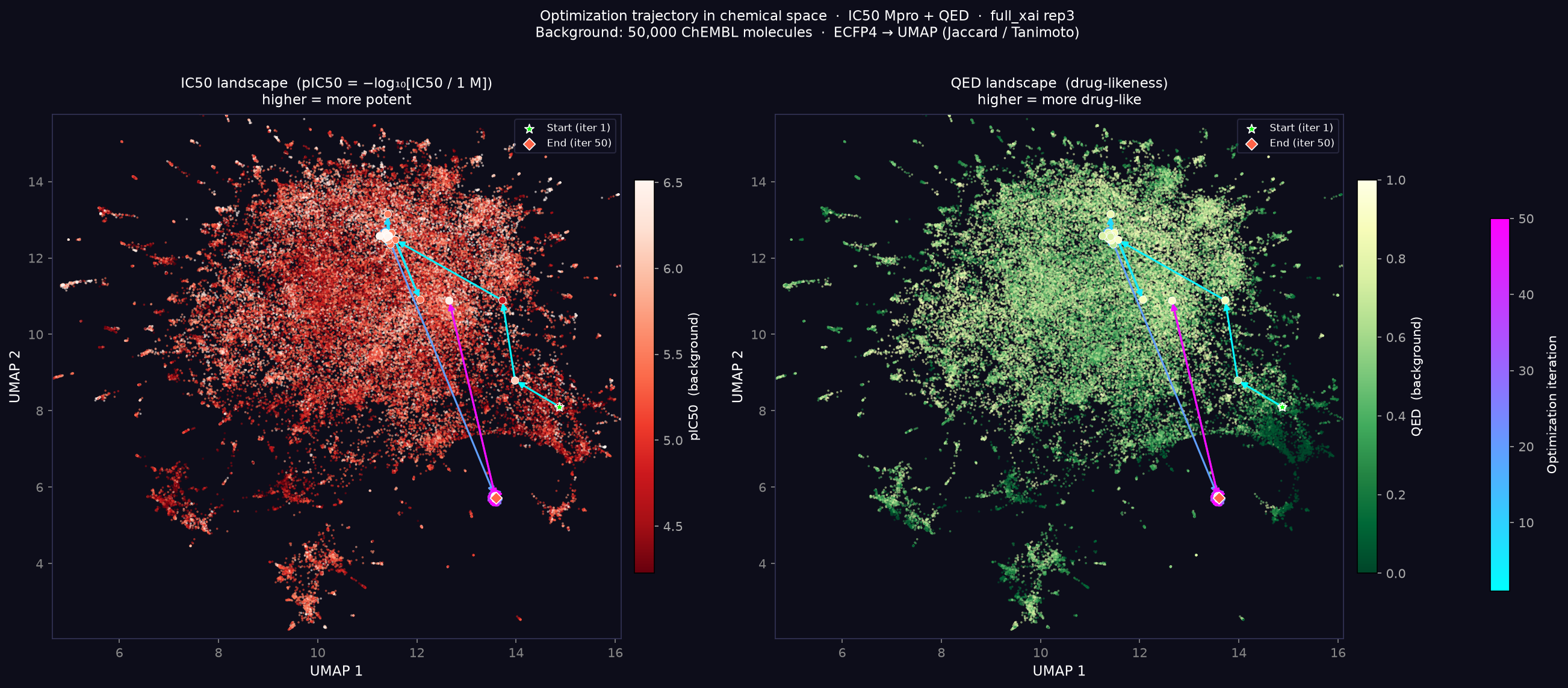}
  \caption{Molecular optimization trajectory in chemical space of the third repetition for the IC$_{50}$-QED-Novel task. Each trajectory is embedded via UMAP (ECFP4, Jaccard) with 50,000 ChEMBL molecules as chemical context. Left: Predicted IC50 landscape. Right: drug-likeness (QED) landscape. Dots represent trajectory molecules connected by directional arrows (iteration order). Stars and diamonds mark start and end points. Background coloring shows the property landscape.}
  \label{fig:appendix-exploration-exploitation-trajectory-rep3}
\end{figure}

We analyzed the molecular optimization trajectories from full explanation mode for the IC$_{50}$-QED-Novel task to investigate the exploration and exploitation behaviour of SEISMO. To achieve this, we investigated the dynamics and visualised the chemical space. For each run, molecules were extracted in iteration order.  Extended connectivity fingerprints (radius 2, 2048 bits) were generated for all trajectory molecules and clustered via Butina clustering at a Tanimoto similarity threshold of 0.65 to identify molecular basins. Across replicates, the best-so-far curve advances in discrete steps drawn from several distinct basins (3–7 high-score-holding basins per run), and the agent allocates more iterations to higher-scoring basins (Spearman $\rho$ between per-basin effort and best pIC50 $= 0.61$–$0.68$), indicating rational effort allocation that refines promising regions and quickly abandons unpromising ones (Figure \ref{fig:appendix-exploration-exploitation-basins}). The dense late-iteration sampling concentrated in high-scoring basins in each panel suggests that after initial exploration that abandons unpromising regions, the agent later exploits high-scoring regions with the remaining budget. 
For the chemical space visualisations (Figures \ref{fig:appendix-exploration-exploitation-trajectory-rep1},  \ref{fig:appendix-exploration-exploitation-trajectory-rep2}, \ref{fig:appendix-exploration-exploitation-trajectory-rep3}), we sampled 50,000 ChEMBL molecules as a chemical context background. Trajectory and background molecules were co-embedded into two dimensions using UMAP (15 neighbors, minimum distance of 0.1, Jaccard metric on ECFP4 fingerprints). The resulting chemical space map was colored by pIC50 and QED separately, with trajectory molecules annotated by iteration order and the start and end positions marked distinctively. It can be seen that SEISMO explores vastly different regions of chemical space in the beginning and, in later iterations, optimizes molecules toward regions further away from the known chemical space of the sampled ChEMBL molecules. This is consistent with the task description of finding novel molecules and shows that the model is not getting stuck in the nearest local minimum, but rather traverses broad distances in chemical space. 

\FloatBarrier

\FloatBarrier
\section{Individual PMO task tables}
\label{sec:appendix:pmo-tables}

The four tables below report per-task PMO results.

\begin{table}[ht]
\centering
\caption{Top1 AUC over 50 iterations. Performance comparison across molecular optimization tasks. Higher is better.}
\label{tab:si-top1-50-iterations}
  \begin{center}
    \begin{small}
      \begin{sc}
      \resizebox{\textwidth}{!}{
\begin{tabular}{lccccccc} 
\toprule
Task  & SEISMO  & $\text{SEISMO}_\text{No Desc.}$ & Zero-shot LLM       & $\text{Zero-shot LLM}_\text{No Desc.}$ & REINVENT  & Graph GA & GP BO \\ 
\midrule
albuterol\_similarity     & \textbf{0.990$\pm$0.000}                                                & 0.407$\pm$0.062                                               & \textbf{0.990$\pm$0.000}        & 0.361$\pm$0.003            & 0.358$\pm$0.021                                               & 0.384$\pm$0.026                                               & 0.372$\pm$0.028                                                \\
amlodipine\_mpo           & \textbf{0.882$\pm$0.001}                                                & 0.465$\pm$0.009                                               & 0.602$\pm$0.066         & 0.156$\pm$0.036            & 0.432$\pm$0.033                                               & 0.425$\pm$0.019                                               & 0.430$\pm$0.035                                                \\
celecoxib\_rediscovery    & \textbf{0.990$\pm$0.000}                                                & 0.386$\pm$0.025                                               & \textbf{0.990$\pm$0.000}         & 0.177$\pm$0.027            & 0.262$\pm$0.036                                               & 0.258$\pm$0.021                                               & 0.250$\pm$0.013                                                \\
deco\_hop                 & \textbf{0.934$\pm$0.017}                                                & 0.568$\pm$0.002                                               & 0.893$\pm$0.002         & 0.501$\pm$0.001            & 0.558$\pm$0.008                                               & 0.801$\pm$0.009                                               & 0.803$\pm$0.008                                                \\
drd2                      & \textbf{0.990$\pm$0.000}                                                & 0.432$\pm$0.418                                               & \textbf{0.990$\pm$0.000}         & 0.165$\pm$0.144            & 0.195$\pm$0.166                                               & 0.168$\pm$0.108                                               & 0.050$\pm$0.029                                                \\
fexofenadine\_mpo         & \textbf{0.971$\pm$0.006}                                                & 0.521$\pm$0.032                                               & 0.949$\pm$0.006         & 0.107$\pm$0.010            & 0.573$\pm$0.022                                               & 0.510$\pm$0.026                                               & 0.562$\pm$0.008                                                \\
gsk3b                     & \textbf{0.965$\pm$0.030}                                                & 0.314$\pm$0.148                                               & 0.719$\pm$0.054         & 0.126$\pm$0.055            & 0.141$\pm$0.023                                               & 0.134$\pm$0.045                                               & 0.163$\pm$0.040                                                \\
isomers\_c7h8n2o2         & \textbf{0.990$\pm$0.000}                                                & 0.760$\pm$0.068                                               & 0.982$\pm$0.002         & 0.640$\pm$0.019            & 0.078$\pm$0.099                                               & 0.045$\pm$0.067                                               & 0.055$\pm$0.087                                                \\
isomers\_c9h10n2o2pf2cl   & 0.985$\pm$0.002                                             & 0.383$\pm$0.144                                               & \textbf{0.990$\pm$0.000}         & 0.272$\pm$0.026            & 0.257$\pm$0.077                                               & 0.229$\pm$0.178                                               & 0.122$\pm$0.038                                                \\
jnk3                      & \textbf{0.429$\pm$0.039}                                                & 0.091$\pm$0.048                                               & 0.427$\pm$0.281         & 0.033$\pm$0.020            & 0.075$\pm$0.022                                               & 0.083$\pm$0.023                                               & 0.072$\pm$0.034                                                \\
median1                   & \textbf{0.388$\pm$0.005}                                                & 0.097$\pm$0.030                                               & 0.335$\pm$0.000         & 0.090$\pm$0.030            & 0.139$\pm$0.013                                               & 0.141$\pm$0.025                                               & 0.146$\pm$0.039                                                \\
median2                   & \textbf{0.386$\pm$0.026}                                                & 0.166$\pm$0.014                                               & 0.382$\pm$0.000         & 0.092$\pm$0.002            & 0.154$\pm$0.013                                               & 0.154$\pm$0.007                                               & 0.155$\pm$0.009                                                \\
mestranol\_similarity     & \textbf{0.990$\pm$0.000}                                                & 0.285$\pm$0.030                                               & \textbf{0.990$\pm$0.000}         & 0.083$\pm$0.004            & 0.275$\pm$0.030                                               & 0.303$\pm$0.025                                               & 0.284$\pm$0.014                                                \\
osimertinib\_mpo          & \textbf{0.827$\pm$0.046}                                                & 0.708$\pm$0.014                                               & 0.531$\pm$0.009         & 0.430$\pm$0.000            & 0.670$\pm$0.037                                               & 0.654$\pm$0.078                                               & 0.636$\pm$0.073                                                \\
perindopril\_mpo          & \textbf{0.791$\pm$0.005}                                                & 0.344$\pm$0.023                                               & 0.762$\pm$0.013         & 0.095$\pm$0.002            & 0.354$\pm$0.018                                               & 0.364$\pm$0.029                                               & 0.352$\pm$0.019                                                \\
qed                       & \textbf{0.914$\pm$0.013}                                                & 0.767$\pm$0.009                                               & 0.884$\pm$0.009         & 0.649$\pm$0.028            & 0.905$\pm$0.023                                               & 0.900$\pm$0.010                                               & 0.897$\pm$0.007                                                \\
ranolazine\_mpo           & \textbf{0.824$\pm$0.011}                                                & 0.401$\pm$0.187                                               & 0.383$\pm$0.006         & 0.009$\pm$0.005            & 0.297$\pm$0.128                                               & 0.238$\pm$0.037                                               & 0.215$\pm$0.066                                                \\
scaffold\_hop             & 0.911$\pm$0.068                                              & 0.429$\pm$0.030                                               & \textbf{0.951$\pm$0.056}         & 0.337$\pm$0.000            & 0.422$\pm$0.016                                               & 0.419$\pm$0.014                                               & 0.421$\pm$0.008                                                \\
sitagliptin\_mpo          & \textbf{0.204$\pm$0.046}                                                & 0.129$\pm$0.065                                               & 0.181$\pm$0.004         & 0.000$\pm$0.000            & 0.095$\pm$0.041                                               & 0.131$\pm$0.052                                               & 0.168$\pm$0.045                                                \\
thiothixene\_rediscovery  & \textbf{0.990$\pm$0.000}                                                & 0.281$\pm$0.032                                               & \textbf{0.990$\pm$0.000}         & 0.170$\pm$0.002            & 0.261$\pm$0.024                                               & 0.232$\pm$0.021                                               & 0.245$\pm$0.014                                                \\
troglitazone\_rediscovery & \textbf{0.962$\pm$0.025}                                                & 0.214$\pm$0.031                                               & 0.852$\pm$0.195         & 0.124$\pm$0.004            & 0.197$\pm$0.014                                               & 0.202$\pm$0.020                                               & 0.206$\pm$0.027                                                \\
valsartan\_smarts         & \textbf{0.121$\pm$0.119}                                                & 0.000$\pm$0.000                                               & 0.020$\pm$0.009         & 0.000$\pm$0.000            & 0.000$\pm$0.000                                               & 0.000$\pm$0.000                                               & 0.000$\pm$0.000                                                \\
zaleplon\_mpo             & \textbf{0.684$\pm$0.038}                                                & 0.351$\pm$0.057                                               & 0.675$\pm$0.009         & 0.000$\pm$0.001            & 0.364$\pm$0.042                                               & 0.269$\pm$0.046                                               & 0.339$\pm$0.065                                                \\ 
\midrule
Sum                       & $\mathbf{18.12}\,^{\scriptscriptstyle +0.26}_{\scriptscriptstyle -0.27}$ & $8.50\,^{\scriptscriptstyle +0.82}_{\scriptscriptstyle -0.76}$ & $16.43\,^{\scriptscriptstyle +0.45}_{\scriptscriptstyle -0.45}$ & $4.62\,^{\scriptscriptstyle +0.21}_{\scriptscriptstyle -0.21}$ & $7.06\,^{\scriptscriptstyle +0.50}_{\scriptscriptstyle -0.42}$ & $7.04\,^{\scriptscriptstyle +0.49}_{\scriptscriptstyle -0.42}$ & $6.95\,^{\scriptscriptstyle +0.34}_{\scriptscriptstyle -0.31}$  \\
Rank  & \textbf{1}    & 3  & 2   & 7 & 4   & 5 & 6  \\
\bottomrule
\end{tabular}
}
      \end{sc}
    \end{small}
  \end{center}
\end{table}

\begin{table}[ht]
\centering
\caption{Best score after 50 iterations. Performance comparison across molecular optimization tasks. Higher is better.}
\label{tab:si-best-score-50-iterations}
\begin{center}
    \begin{small}
      \begin{sc}
      \resizebox{\textwidth}{!}{
\begin{tabular}{lccccccc} 
\toprule
Task                      & SEISMO                                                                      & $\text{SEISMO}_\text{No Desc.}$                                & Zero-shot LLM        & $\text{Zero-shot LLM}_\text{No Desc.}$ & REINVENT                                                      & Graph GA                                                      & GP BO                                                          \\ 
\midrule
albuterol\_similarity     & \textbf{1.000$\pm$0.000}                                                    & 0.538$\pm$0.108                                                & \textbf{1.000$\pm$0.000}          & 0.461$\pm$0.033              & 0.390$\pm$0.028                                               & 0.422$\pm$0.034                                               & 0.412$\pm$0.039                                                \\
amlodipine\_mpo           & \textbf{0.905$\pm$0.001}                                                    & 0.532$\pm$0.032                                                & 0.885$\pm$0.000          & 0.380$\pm$0.014              & 0.471$\pm$0.030                                               & 0.463$\pm$0.020                                               & 0.464$\pm$0.028                                                \\
celecoxib\_rediscovery    & \textbf{1.000$\pm$0.000}                                                    & 0.444$\pm$0.073                                                & \textbf{1.000$\pm$0.000}          & 0.208$\pm$0.046              & 0.297$\pm$0.046                                               & 0.293$\pm$0.033                                               & 0.285$\pm$0.033                                                \\
deco\_hop                 & \textbf{0.961$\pm$0.022}                                                    & 0.598$\pm$0.003                                                & 0.915$\pm$0.013          & 0.512$\pm$0.000              & 0.572$\pm$0.012                                               & 0.818$\pm$0.008                                               & 0.820$\pm$0.011                                                \\
drd2                      & \textbf{1.000$\pm$0.000}                                                    & 0.675$\pm$0.559                                                & 1.000$\pm$0.000          & 0.192$\pm$0.146              & 0.322$\pm$0.307                                               & 0.249$\pm$0.150                                               & 0.072$\pm$0.045                                                \\
fexofenadine\_mpo         & 0.992$\pm$0.008                                                    & 0.611$\pm$0.048                                                & \textbf{0.993$\pm$0.010}          & 0.270$\pm$0.102              & 0.627$\pm$0.042                                               & 0.579$\pm$0.031                                               & 0.627$\pm$0.027                                                \\
gsk3b                     & \textbf{0.983$\pm$0.029}                                                    & 0.483$\pm$0.123                                                & 0.760$\pm$0.071          & 0.170$\pm$0.099              & 0.204$\pm$0.106                                               & 0.212$\pm$0.093                                               & 0.206$\pm$0.058                                                \\
isomers\_c7h8n2o2         & \textbf{1.000$\pm$0.000}                                                    & 0.860$\pm$0.128                                                & \textbf{1.000$\pm$0.000}          & 0.706$\pm$0.050              & 0.149$\pm$0.227                                               & 0.087$\pm$0.103                                               & 0.075$\pm$0.093                                                \\
isomers\_c9h10n2o2pf2cl   & \textbf{1.000$\pm$0.000}                                                    & 0.500$\pm$0.175                                                & \textbf{1.000$\pm$0.000}          & 0.363$\pm$0.032              & 0.411$\pm$0.077                                               & 0.422$\pm$0.162                                               & 0.354$\pm$0.181                                                \\
jnk3                      & \textbf{0.503$\pm$0.075}                                                    & 0.163$\pm$0.097                                                & 0.480$\pm$0.269          & 0.065$\pm$0.064              & 0.128$\pm$0.058                                               & 0.110$\pm$0.019                                               & 0.170$\pm$0.137                                                \\
median1                   & \textbf{0.400$\pm$0.000}                                                    & 0.135$\pm$0.024                                                & 0.340$\pm$0.000          & 0.112$\pm$0.033              & 0.162$\pm$0.024                                               & 0.168$\pm$0.030                                               & 0.166$\pm$0.041                                                \\
median2                   & \textbf{0.395$\pm$0.028}                                                    & 0.193$\pm$0.021                                                & 0.388$\pm$0.000          & 0.107$\pm$0.009              & 0.177$\pm$0.018                                               & 0.172$\pm$0.014                                               & 0.169$\pm$0.013                                                \\
mestranol\_similarity     & \textbf{1.000$\pm$0.000}                                                    & 0.379$\pm$0.033                                                & \textbf{1.000$\pm$0.000}          & 0.112$\pm$0.008              & 0.295$\pm$0.019                                               & 0.336$\pm$0.019                                               & 0.311$\pm$0.025                                                \\
osimertinib\_mpo          & \textbf{0.939$\pm$0.044}                                                    & 0.791$\pm$0.026                                                & 0.665$\pm$0.002          & 0.515$\pm$0.000              & 0.719$\pm$0.020                                               & 0.751$\pm$0.022                                               & 0.733$\pm$0.042                                                \\
perindopril\_mpo          & \textbf{0.810$\pm$0.000}                                                    & 0.432$\pm$0.024                                                & 0.793$\pm$0.005          & 0.129$\pm$0.014              & 0.391$\pm$0.011                                               & 0.404$\pm$0.021                                               & 0.394$\pm$0.033                                                \\
qed                       & \textbf{0.940$\pm$0.010}                                                    & 0.803$\pm$0.002                                                & 0.891$\pm$0.015          & 0.745$\pm$0.109              & 0.924$\pm$0.019                                               & 0.924$\pm$0.015                                               & 0.932$\pm$0.006                                                \\
ranolazine\_mpo           & \textbf{0.939$\pm$0.004}                                                    & 0.559$\pm$0.185                                                & 0.754$\pm$0.030          & 0.012$\pm$0.006              & 0.343$\pm$0.129                                               & 0.341$\pm$0.077                                               & 0.257$\pm$0.075                                                \\
scaffold\_hop             & 0.935$\pm$0.065                                                    & 0.461$\pm$0.035                                                & \textbf{1.000$\pm$0.000}          & 0.347$\pm$0.000              & 0.438$\pm$0.014                                               & 0.439$\pm$0.014                                               & 0.433$\pm$0.011                                                \\
sitagliptin\_mpo          & \textbf{0.295$\pm$0.043}                                                    & 0.219$\pm$0.108                                                & 0.206$\pm$0.029          & 0.001$\pm$0.001              & 0.136$\pm$0.063                                               & 0.157$\pm$0.052                                               & 0.207$\pm$0.054                                                \\
thiothixene\_rediscovery  & \textbf{1.000$\pm$0.000}                                                    & 0.350$\pm$0.073                                                & \textbf{1.000$\pm$0.000}          & 0.184$\pm$0.003              & 0.279$\pm$0.026                                               & 0.258$\pm$0.037                                               & 0.282$\pm$0.031                                                \\
troglitazone\_rediscovery & \textbf{1.000$\pm$0.000}                                                    & 0.311$\pm$0.019                                                & 0.869$\pm$0.186          & 0.175$\pm$0.000              & 0.214$\pm$0.018                                               & 0.217$\pm$0.018                                               & 0.237$\pm$0.028                                                \\
valsartan\_smarts         & \textbf{0.230$\pm$0.205}                                                    & 0.000$\pm$0.000                                                & 0.028$\pm$0.004          & 0.000$\pm$0.000              & 0.002$\pm$0.004                                               & 0.000$\pm$0.000                                               & 0.000$\pm$0.000                                                \\
zaleplon\_mpo             & \textbf{0.729$\pm$0.024}                                                    & 0.418$\pm$0.080                                                & 0.693$\pm$0.001          & 0.001$\pm$0.001              & 0.414$\pm$0.056                                               & 0.360$\pm$0.063                                               & 0.390$\pm$0.049                                                \\ 
\midrule
Sum                       & $\mathbf{18.96\,^{\scriptscriptstyle +0.32}_{\scriptscriptstyle -0.39}}$ & $10.45\,^{\scriptscriptstyle +0.90}_{\scriptscriptstyle -1.09}$ & $17.66\,^{\scriptscriptstyle +0.40}_{\scriptscriptstyle -0.40}$ & $5.77\,^{\scriptscriptstyle +0.35}_{\scriptscriptstyle -0.35}$ & $8.06\,^{\scriptscriptstyle +0.88}_{\scriptscriptstyle -0.62}$ & $8.18\,^{\scriptscriptstyle +0.55}_{\scriptscriptstyle -0.48}$ & $8.00\,^{\scriptscriptstyle +0.53}_{\scriptscriptstyle -0.48}$  \\
Rank                      & \textbf{1}                                                                  & 3                                                              & 2                    & 7                        & 5                                                             & 4                                                             & 6                                                              \\
\bottomrule
\end{tabular}
}
      \end{sc}
    \end{small}
  \end{center}
\end{table}

\begin{table}[ht]
\centering
\caption{Top1 AUC over 10000 iterations. Performance comparison across molecular optimization tasks. Higher is better.}
\label{tab:si-top1-10000-iterations}
\begin{center}
    \begin{small}
      \begin{sc}
      
      \resizebox{\textwidth}{!}{
\begin{tabular}{lccccccc} 
\toprule
Task                      & SEISMO                                                                      & $\text{SEISMO}_\text{No Desc.}$                                & Zero-shot LLM           & $\text{Zero-shot LLM}_\text{No Desc.}$ & REINVENT                                                       & Graph GA                                                       & GP BO                                                           \\ 
\midrule
albuterol\_similarity     & \textbf{1.000$\pm$0.000}                                                    & 0.537$\pm$0.107                                                & \textbf{1.000$\pm$0.000}            & 0.461$\pm$0.033             & 0.906$\pm$0.004                                                & 0.877$\pm$0.025                                                & 0.925$\pm$0.013                                                 \\
amlodipine\_mpo           & \textbf{0.905$\pm$0.001}                                                    & 0.532$\pm$0.032                                                & 0.884$\pm$0.000            & 0.379$\pm$0.014             & 0.655$\pm$0.042                                                & 0.688$\pm$0.023                                                & 0.609$\pm$0.049                                                 \\
celecoxib\_rediscovery    & \textbf{1.000$\pm$0.000}                                                    & 0.444$\pm$0.072                                                & \textbf{1.000$\pm$0.000}            & 0.208$\pm$0.046             & 0.803$\pm$0.110                                                & 0.684$\pm$0.137                                                & 0.809$\pm$0.084                                                 \\
deco\_hop                 & \textbf{0.961$\pm$0.022}                                                    & 0.598$\pm$0.003                                                & 0.915$\pm$0.013            & 0.512$\pm$0.000             & 0.682$\pm$0.054                                                & 0.627$\pm$0.006                                                & 0.648$\pm$0.030                                                 \\
drd2                      & \textbf{1.000$\pm$0.000}                                                    & 0.673$\pm$0.558                                                & \textbf{1.000$\pm$0.000}            & 0.192$\pm$0.146             & 0.968$\pm$0.008                                                & 0.991$\pm$0.002                                                & 0.958$\pm$0.008                                                 \\
fexofenadine\_mpo         & 0.992$\pm$0.008                                                    & 0.611$\pm$0.048                                                & \textbf{0.993$\pm$0.010}            & 0.269$\pm$0.101             & 0.804$\pm$0.008                                                & 0.777$\pm$0.013                                                & 0.743$\pm$0.008                                                 \\
gsk3b                     & \textbf{0.983$\pm$0.029}                                                    & 0.482$\pm$0.123                                                & 0.760$\pm$0.071            & 0.170$\pm$0.099             & 0.894$\pm$0.050                                                & 0.829$\pm$0.078                                                & 0.879$\pm$0.044                                                 \\
isomers\_c7h8n2o2         & \textbf{1.000$\pm$0.000}                                                    & 0.860$\pm$0.128                                                & \textbf{1.000$\pm$0.000}            & 0.705$\pm$0.050             & 0.884$\pm$0.033                                                & 0.899$\pm$0.067                                                & 0.747$\pm$0.125                                                 \\
isomers\_c9h10n2o2pf2cl   & \textbf{1.000$\pm$0.000}                                                    & 0.499$\pm$0.175                                                & \textbf{1.000$\pm$0.000}            & 0.362$\pm$0.032             & 0.673$\pm$0.066                                                & 0.766$\pm$0.052                                                & 0.514$\pm$0.192                                                 \\
jnk3                      & 0.503$\pm$0.075                                                             & 0.163$\pm$0.097                                                & 0.480$\pm$0.269            & 0.065$\pm$0.063             & \textbf{0.814$\pm$0.028}                                       & 0.598$\pm$0.159                                                & 0.593$\pm$0.179                                                 \\
median1                   & \textbf{0.400$\pm$0.000}                                                    & 0.135$\pm$0.024                                                & 0.340$\pm$0.000            & 0.112$\pm$0.033             & 0.368$\pm$0.011                                                & 0.320$\pm$0.031                                                & 0.317$\pm$0.019                                                 \\
median2                   & \textbf{0.395$\pm$0.028}                                                    & 0.193$\pm$0.021                                                & 0.387$\pm$0.000            & 0.107$\pm$0.009             & 0.290$\pm$0.010                                                & 0.289$\pm$0.009                                                & 0.310$\pm$0.011                                                 \\
mestranol\_similarity     & \textbf{1.000$\pm$0.000}                                                    & 0.378$\pm$0.033                                                & \textbf{1.000$\pm$0.000}            & 0.111$\pm$0.008             & 0.639$\pm$0.054                                                & 0.617$\pm$0.031                                                & 0.667$\pm$0.092                                                 \\
osimertinib\_mpo          & \textbf{0.938$\pm$0.044}                                                    & 0.791$\pm$0.026                                                & 0.665$\pm$0.002            & 0.515$\pm$0.000             & 0.852$\pm$0.012                                                & 0.849$\pm$0.007                                                & 0.806$\pm$0.006                                                 \\
perindopril\_mpo          & \textbf{0.810$\pm$0.000}                                                    & 0.431$\pm$0.024                                                & 0.793$\pm$0.005            & 0.129$\pm$0.014             & 0.555$\pm$0.019                                                & 0.561$\pm$0.012                                                & 0.514$\pm$0.015                                                 \\
qed                       & 0.940$\pm$0.011                                                             & 0.803$\pm$0.002                                                & 0.891$\pm$0.015            & 0.744$\pm$0.108             & \textbf{0.948$\pm$0.000}                                       & 0.947$\pm$0.000                                                & 0.946$\pm$0.001                                                 \\
ranolazine\_mpo           & \textbf{0.939$\pm$0.004}                                                    & 0.558$\pm$0.185                                                & 0.752$\pm$0.029            & 0.012$\pm$0.006             & 0.789$\pm$0.011                                                & 0.760$\pm$0.015                                                & 0.763$\pm$0.015                                                 \\
scaffold\_hop             & 0.935$\pm$0.065                                                    & 0.461$\pm$0.035                                                & \textbf{1.000$\pm$0.000}            & 0.347$\pm$0.000             & 0.575$\pm$0.025                                                & 0.528$\pm$0.010                                                & 0.565$\pm$0.026                                                 \\
sitagliptin\_mpo          & 0.294$\pm$0.043                                                             & 0.219$\pm$0.108                                                & 0.206$\pm$0.029            & 0.001$\pm$0.001             & 0.055$\pm$0.017                                                & \textbf{0.492$\pm$0.076}                                       & 0.238$\pm$0.069                                                 \\
thiothixene\_rediscovery  & \textbf{1.000$\pm$0.000}                                                    & 0.349$\pm$0.072                                                & \textbf{1.000$\pm$0.000}            & 0.184$\pm$0.003             & 0.559$\pm$0.016                                                & 0.508$\pm$0.029                                                & 0.592$\pm$0.029                                                 \\
troglitazone\_rediscovery & \textbf{1.000$\pm$0.000}                                                    & 0.311$\pm$0.019                                                & 0.869$\pm$0.186            & 0.175$\pm$0.000             & 0.460$\pm$0.038                                                & 0.411$\pm$0.019                                                & 0.432$\pm$0.018                                                 \\
valsartan\_smarts         & \textbf{0.230$\pm$0.205}                                                    & 0.000$\pm$0.000                                                & 0.028$\pm$0.004            & 0.000$\pm$0.000             & 0.187$\pm$0.418                                                & 0.000$\pm$0.000                                                & 0.000$\pm$0.000                                                 \\
zaleplon\_mpo             & \textbf{0.729$\pm$0.024}                                                    & 0.418$\pm$0.080                                                & 0.693$\pm$0.001            & 0.001$\pm$0.001             & 0.383$\pm$0.070                                                & 0.367$\pm$0.038                                                & 0.252$\pm$0.079                                                 \\ 
\midrule
Sum                       & $\mathbf{18.95\,^{\scriptscriptstyle +0.32}_{\scriptscriptstyle -0.39}}$ & $10.44\,^{\scriptscriptstyle +0.90}_{\scriptscriptstyle -1.09}$ & $17.66\,^{\scriptscriptstyle +0.40}_{\scriptscriptstyle -0.40}$ & $5.76\,^{\scriptscriptstyle +0.35}_{\scriptscriptstyle -0.35}$   & $14.74\,^{\scriptscriptstyle +0.95}_{\scriptscriptstyle -0.51}$ & $14.38\,^{\scriptscriptstyle +0.45}_{\scriptscriptstyle -0.47}$ & $13.83\,^{\scriptscriptstyle +0.60}_{\scriptscriptstyle -0.60}$  \\
Rank   & 1   & 6   & 2   &  7  & 3  & 4 & 5  \\
\bottomrule
\end{tabular}
}
      \end{sc}
    \end{small}
  \end{center}
\end{table}

\begin{table}[ht]
\centering
\caption{Top10 AUC over 1000 iterations. Performance comparison across molecular optimization tasks. Higher is better.}
\label{tab:si-top10-1000-iterations}
\begin{center}
    \begin{small}
      \begin{sc}
      
      \rotatebox{90}{
      \resizebox{\textheight}{!}{
\begin{tabular}{lccccccccccc} 
\toprule
Task                       & SEISMO                   & $\text{SEISMO}_\text{No Desc.}$ & Zero-shot LLM           & $\text{Zero-shot LLM}_\text{No Desc.}$   & GP BO           & Graph GA        & REINVENT        & LICO                     & Genetic GFN     & Augmented Memory & MOLLEO           \\ 
\midrule
albuterol\_similarity      & $\mathbf{0.999\pm0.000}$ & $0.495\pm0.093$                 & \textbf{0.999$\pm$0.000}            & 0.347$\pm$0.013           & $0.636\pm0.106$ & $0.583\pm0.065$ & $0.496\pm0.020$ & $0.656\pm0.125$          & $0.664\pm0.054$ & $0.557\pm0.048$  & $0.886\pm0.023$  \\
amlodipine\_mpo            & $\mathbf{0.898\pm0.001}$ & $0.516\pm0.027$                 & 0.674$\pm$0.130            & 0.189$\pm$0.053           & $0.519\pm0.014$ & $0.501\pm0.016$ & $0.472\pm0.008$ & $0.541\pm0.026$          & $0.534\pm0.019$ & $0.489\pm0.009$  & $0.637\pm0.023$  \\
celecoxib\_rediscovery     & $\mathbf{0.883\pm0.002}$ & $0.422\pm0.060$                 & 0.869$\pm$0.009            & 0.167$\pm$0.027           & $0.411\pm0.046$ & $0.424\pm0.049$ & $0.370\pm0.029$ & $0.447\pm0.073$          & $0.447\pm0.028$ & $0.385\pm0.027$  & $0.402\pm0.003$  \\
deco\_hop                  & $\mathbf{0.956\pm0.023}$ & $0.593\pm0.003$                 & 0.885$\pm$0.001            & 0.505$\pm$0.000           & $0.593\pm0.013$ & $0.581\pm0.006$ & $0.572\pm0.006$ & $0.596\pm0.010$          & $0.604\pm0.017$ & $0.579\pm0.010$  & $0.588\pm0.007$  \\
drd2                       & $\mathbf{0.999\pm0.000}$ & $0.652\pm0.541$                 & 0.997$\pm$0.001            & 0.059$\pm$0.046           & $0.857\pm0.080$ & $0.833\pm0.065$ & $0.775\pm0.086$ & $0.859\pm0.066$          & $0.809\pm0.045$ & $0.795\pm0.024$  & $0.910\pm0.017$  \\
fexofenadine\_mpo          & $\mathbf{0.983\pm0.003}$ & $0.595\pm0.043$                 & 0.952$\pm$0.004            & 0.066$\pm$0.024           & $0.707\pm0.021$ & $0.666\pm0.009$ & $0.650\pm0.007$ & $0.700\pm0.023$          & $0.682\pm0.021$ & $0.679\pm0.021$  & $0.674\pm0.002$  \\
gsk3b                      & $\mathbf{0.974\pm0.031}$ & $0.420\pm0.130$                 & 0.536$\pm$0.020            & 0.065$\pm$0.020           & $0.611\pm0.059$ & $0.523\pm0.047$ & $0.589\pm0.063$ & $0.617\pm0.063$          & $0.637\pm0.018$ & $0.539\pm0.097$  & $0.397\pm0.013$  \\
isomers\_c7h8n2o2          & $\mathbf{0.999\pm0.001}$ & $0.796\pm0.112$                 & 0.998$\pm$0.000            & 0.516$\pm$0.084           & $0.545\pm0.158$ & $0.735\pm0.112$ & $0.725\pm0.064$ & $0.779\pm0.099$          & $0.738\pm0.039$ & $0.661\pm0.039$  & $0.737\pm0.043$  \\
isomers\_c9h10n2o2pf2cl    & $\mathbf{0.998\pm0.000}$ & $0.443\pm0.147$                 & 0.965$\pm$0.023            & 0.233$\pm$0.041           & $0.599\pm0.059$ & $0.630\pm0.086$ & $0.630\pm0.032$ & $0.672\pm0.075$          & $0.656\pm0.075$ & $0.596\pm0.066$  & $0.635\pm0.017$  \\
jnk3                       & $\mathbf{0.464\pm0.065}$ & $0.134\pm0.077$                 & 0.209$\pm$0.104            & 0.018$\pm$0.014           & $0.346\pm0.067$ & $0.301\pm0.071$ & $0.315\pm0.042$ & $0.336\pm0.051$          & $0.409\pm0.165$ & $0.294\pm0.110$  & $0.186\pm0.076$  \\
median1                    & $\mathbf{0.343\pm0.001}$ & $0.129\pm0.023$                 & 0.315$\pm$0.003            & 0.073$\pm$0.026           & $0.213\pm0.020$ & $0.208\pm0.015$ & $0.205\pm0.012$ & $0.217\pm0.019$          & $0.219\pm0.008$ & $0.219\pm0.014$  & $0.236\pm0.021$  \\
median2                    & $\mathbf{0.365\pm0.028}$ & $0.183\pm0.018$                 & 0.326$\pm$0.003            & 0.090$\pm$0.004           & $0.203\pm0.009$ & $0.181\pm0.009$ & $0.188\pm0.010$ & $0.193\pm0.009$          & $0.204\pm0.011$ & $0.184\pm0.010$  & $0.191\pm0.009$  \\
mestranol\_similarity      & $\mathbf{0.999\pm0.000}$ & $0.353\pm0.032$                 & \textbf{0.999$\pm$0.000}            & 0.077$\pm$0.003           & $0.427\pm0.025$ & $0.362\pm0.017$ & $0.379\pm0.026$ & $0.423\pm0.016$          & $0.414\pm0.022$ & $0.393\pm0.021$  & $0.399\pm0.020$  \\
osimertinib\_mpo           & $\mathbf{0.924\pm0.047}$ & $0.775\pm0.024$                 & 0.424$\pm$0.015            & 0.092$\pm$0.002           & $0.766\pm0.006$ & $0.751\pm0.005$ & $0.737\pm0.007$ & $0.759\pm0.008$          & $0.763\pm0.008$ & $0.761\pm0.006$  & $0.779\pm0.006$  \\
perindopril\_mpo           & $\mathbf{0.809\pm0.000}$ & $0.419\pm0.028$                 & 0.777$\pm$0.004            & 0.092$\pm$0.000           & $0.458\pm0.019$ & $0.435\pm0.016$ & $0.404\pm0.009$ & $0.473\pm0.009$          & $0.462\pm0.033$ & $0.422\pm0.013$  & $0.655\pm0.054$  \\
qed                        & $\mathbf{0.937\pm0.010}$ & $0.799\pm0.001$                 & 0.850$\pm$0.001            & 0.651$\pm$0.017           & $0.912\pm0.010$ & $0.914\pm0.007$ & $0.921\pm0.002$ & $0.925\pm0.005$          & $0.928\pm0.002$ & $0.923\pm0.002$  & $0.919\pm0.006$  \\
ranolazine\_mpo            & $\mathbf{0.926\pm0.004}$ & $0.536\pm0.191$                 & 0.511$\pm$0.072            & 0.007$\pm$0.002           & $0.701\pm0.023$ & $0.620\pm0.014$ & $0.574\pm0.044$ & $0.687\pm0.029$          & $0.623\pm0.022$ & $0.614\pm0.033$  & $0.640\pm0.000$  \\
scaffold\_hop              & $\mathbf{0.931\pm0.065}$ & $0.457\pm0.035$                 & 0.908$\pm$0.099            & 0.340$\pm$0.001           & $0.478\pm0.009$ & $0.461\pm0.008$ & $0.447\pm0.010$ & $0.480\pm0.008$          & $0.485\pm0.015$ & $0.460\pm0.010$  & $0.473\pm0.000$  \\
sitagliptin\_mpo           & $0.234\pm0.059$          & $0.173\pm0.085$                 & 0.095$\pm$0.025            & 0.000$\pm$0.000           & $0.232\pm0.083$ & $0.229\pm0.053$ & $0.261\pm0.026$ & $\mathbf{0.315\pm0.097}$ & $0.227\pm0.041$ & $0.245\pm0.030$  & $0.193\pm0.073$  \\
thiothixene\_rediscovery   & $\mathbf{0.919\pm0.041}$ & $0.333\pm0.067$                 & 0.918$\pm$0.005            & 0.167$\pm$0.007           & $0.351\pm0.039$ & $0.322\pm0.023$ & $0.311\pm0.021$ & $0.343\pm0.035$          & $0.377\pm0.015$ & $0.336\pm0.033$  & $0.416\pm0.075$  \\
troglitazone\_rediscovery  & $\mathbf{0.813\pm0.083}$ & $0.287\pm0.026$                 & 0.761$\pm$0.132            & 0.124$\pm$0.005           & $0.313\pm0.018$ & $0.267\pm0.015$ & $0.246\pm0.009$ & $0.292\pm0.028$          & $0.277\pm0.015$ & $0.262\pm0.012$  & $0.302\pm0.022$  \\
valsartan\_smarts          & $\mathbf{0.137\pm0.119}$ & $0.000\pm0.000$                 & 0.004$\pm$0.002            & 0.000$\pm$0.000           & $0.000\pm0.000$ & $0.000\pm0.000$ & $0.000\pm0.000$ & $0.000\pm0.000$          & $0.000\pm0.000$ & $0.000\pm0.000$  & $0.000\pm0.000$  \\
zaleplon\_mpo              & $\mathbf{0.666\pm0.034}$ & $0.398\pm0.072$                 & 0.636$\pm$0.014            & 0.000$\pm$0.000           & $0.392\pm0.034$ & $0.374\pm0.024$ & $0.406\pm0.017$ & $0.404\pm0.022$          & $0.400\pm0.014$ & $0.415\pm0.013$  & $0.392\pm0.003$  \\ 
\midrule
Sum of scores ($\uparrow$) & $\mathbf{18.16}$         & $9.90$                          & $15.61$ & $3.88$ & $11.27$         & $10.90$         & $10.68$         & $11.71$                  & $11.56$         & $10.81$          & $11.65$          \\
\bottomrule
\end{tabular}
}}
      \end{sc}
    \end{small}
  \end{center}
\end{table}

\FloatBarrier

\section{Example prompts, explanations and optimisation trajectory}
\label{sec:appendix:examples}

\subsection{System prompt}
\label{sec:appendix:system-prompt}

The prompt below specifies a closed-loop proposal mechanism in which the language model is conditioned on the complete history of previously evaluated molecules and is required to output exactly one new candidate per iteration in a fixed JSON format.

\vspace{0.5em}
{\small
\begin{verbatim}
You are an expert medicinal chemist whose job is to propose and optimize new molecules.

You interact in a multi-step optimization loop.

In this loop you:
- Propose small organic molecules as SMILES strings.
- Receive evaluations of your proposals.
- Have access to the full conversation history, including all past proposals,
  rationales, and evaluations.

Your goal in this conversation is to iteratively propose new molecules that move
the objective toward the specified target, using the information in the
conversation history and your general chemical knowledge.

OUTPUT FORMAT:

For every proposal, you must respond with a single JSON object of the form

{
  "reason": "<short explanation>",
  "smiles": "<SMILES string>"
}

Requirements:
- "smiles" must be a single, valid SMILES string for a plausible small molecule.
- "reason" should briefly explain why, given the history of molecules and their
  scores, this new proposal might improve or further explore the objective
  landscape.
- Do not repeat any previous molecule exactly. The scores are deterministic and
  will not change if a molecule is re-evaluated.
- Do not include any text outside the JSON object. No Markdown, no comments,
  no code fences.
\end{verbatim}
}

\subsection{PMO benchmark first messages}
\label{sec:appendix:pmo-prompts}

For the PMO benchmark we include three example first messages --- Celecoxib rediscovery, Osimertinib MPO, and DRD2 --- to illustrate the prompting setup. These messages follow the level of explicitness established in prior literature such as MolLEO and LICO. For brevity, the remaining 20 first messages are available in the project's GitHub repository.

\begin{ChatLog}
Rediscovery task: The rediscovery score ([0,1], higher is better) measures the Tanimoto similarity with the SMILES of celecoxib.
Objective:
- maximize the similarity score (0-1; higher is better).
- You will have 50 iterations to optimize the molecule.
Step 1:
Propose a single initial molecule as a SMILES string that you think will achieve a high similarity score to celecoxib.
Respond with a single JSON object: {"reason": "<why this is a reasonable starting point for this objective>", "smiles": "<SMILES string>"}


Osimertinib MPO: score is the geometric mean of 4 terms (each in [0,1]):
score = gmean([sim_v1, sim_v2, tpsa_term, logp_term])
Reference molecule (Osimertinib) SMILES:
COc1cc(N(C)CCN(C)C)c(NC(=O)C=C)cc1Nc2nccc(n2)c3cn(C)c4ccccc34
Terms:
sim_v1  = clip(Tanimoto(FCFP4(test), FCFP4(ref)), t=0.8)
sim_v2  = min_gauss(Tanimoto(ECFP6(test), ECFP6(ref)); mu=0.85, sigma=0.1)
tpsa_term = max_gauss(TPSA(test); mu=100, sigma=10)
logp_term = min_gauss(MolLogP(test); mu=1, sigma=1)
Modifiers:
clip(x,t) = 1 if x>=t else x/t
min_gauss(x;mu,s) = 1 if x<=mu else exp(-0.5*((x-mu)/s)^2)
max_gauss(x;mu,s) = 1 if x>=mu else exp(-0.5*((x-mu)/s)^2)
Objective:
- maximize the Osimertinib MPO score (0-1; higher is better).
- You will have 50 iterations to optimize the molecule.
Step 1:
Propose a single initial molecule as a SMILES string that you think will have a high MPO score.
Respond with a single JSON object: {"reason": "<...>", "smiles": "<SMILES string>"}


Optimize predicted bioactivity against Dopamine Receptor D2 (DRD2). Scores are in [0, 1] with higher being better.
Objective:
- maximize the predicted DRD2 activity score (0-1; higher is better).
- You will have 50 iterations to optimize the molecule.
Step 1:
Propose a single initial molecule as a SMILES string that you think has good predicted bioactivity against DRD2.
Respond with a single JSON object: {"reason": "<...>", "smiles": "<SMILES string>"}
\end{ChatLog}

\subsection{Example explanations}
\label{sec:appendix:example-explanations}

For brevity, only one example explanation per oracle is shown here. Example trajectories and additional explanations are available in the project repository.

\paragraph{QED.}
{\small
\begin{verbatim}
Weighted log contributions (closer to 0 = better, negative values hurt score):

MW:     -0.786 (value=454.61, optimal=281-332 Da)
ALOGP:  -0.112 (value=4.26, optimal=1.95-3.57)
HBA:    -0.006 (value=4, optimal=2-3)
HBD:    -0.592 (value=3, optimal=1)
PSA:    -0.013 (value=87.66, optimal=40-69)
ROTB:   -1.009 (value=10, optimal=2-4)
AROM:   -0.031 (value=2, optimal=1-2)
ALERTS: -0.163 (value=0, optimal=0)

\end{verbatim}
}

\paragraph{IC$_{50}$ via SHAP.}
{\small
\begin{ChatLog}
The following attribution values are calculated using SHAP. These are dependent on the individual input:

error of SHAP = prediction - (expected value + total SHAP attribution)
prediction (pIC50) = 4.079466
expected value    = 5.320687
total SHAP attribution = -1.241216
error = -4.76837158203125e-06

Top 10: highest attribution values (pushing towards higher pIC50 -> lower IC50):

| Attribution | MACCS value | MACCS key | SMARTS substructure             |
| 0.216906    | 0           | 65        | c:n                             |
| 0.120592    | 0           | 153       | [!#6;!#1]~[CH2]~*               |
| 0.078208    | 1           | 54        | [!#6;!#1;!H0]~*~*~[!#6;!#1;!H0] |
| ...

Bottom 10: lowest attribution values (pushing towards lower pIC50 -> higher IC50):
| ...
| -0.200778 | 1 | 123 | [#8]~[#6]~[#8]   |
| -0.293260 | 1 | 139 | [O;!H0]          |

\end{ChatLog}
}

\paragraph{Similarity (Quercetin target vs.\ proposal).}
{\small
\begin{ChatLog}
Target: O=C1c3c(O/C(=C1/O)c2ccc(O)c(O)c2)cc(O)cc3O   (Quercetin)
Query:  O=C1c2c(OC)cc(O)cc2OC(=C1OC)c3ccc(OC)cc3      (agent proposal)

Features present in TARGET but not in QUERY (reducing similarity):
  Key 53:  [!#6;!#1;!H0]~*~*~*~[!#6;!#1;!H0]
  Key 54:  [!#6;!#1;!H0]~*~*~[!#6;!#1;!H0]
  Key 131: [!#6;!#1;!H0] (count > 1)

Features present in QUERY but not in TARGET (reducing similarity):
  Key 93:  [!#6;!#1]~[CH3]
  Key 126: *!@[#8]!@*
  Key 141: [CH3] (count > 2)
  Key 149: [C;H3,H4] (count > 1)
  Key 160: [C;H3,H4]
\end{ChatLog}
}

\paragraph{Co-folding close residues.}
The Boltz-2 explanation reports confidence scores (\texttt{confidence\_score}, ipTM, complex pLDDT, complex pIDE) together with three-letter codes and indices of all protein residues with a non-hydrogen atom within 5\,\AA{} of any non-hydrogen atom of the proposed ligand. An example for SARS-CoV-2 M$^\mathrm{pro}$ is provided in the project repository.

\clearpage

\subsection{Example optimisation trajectory}
\label{sec:appendix:example-trajectory}

% data/results/ic50mpro_qed/claude-opus-4.5_full_xai/ic50mpro_qed_novel_full_rep3_conversation_20261118_180757.json

\noindent
The conversation below was generated using the Claude Opus 4.5 model. The goal was to reduce the XGBoost-predicted IC\textsubscript{50} to below 1 nM while achieving a QED of at least 0.6 and ensuring the final molecule is novel (not listed in PubChem). Because these thresholds were highly ambitious, the generative agent did not meet all objectives and ran for the maximum of 50 iterations. The “full explanation” mode was enabled, providing QED and IC\textsubscript{50} oracle explanations at each iteration like shown in Appendix \ref{sec:appendix:example-explanations} for iteration 1. The rest is available on our GitHub.

\begin{ChatLog}

Iteration 1: "reason": "Starting with a peptidomimetic scaffold commonly found in protease inhibitors. This molecule features a hydroxyl group that can act as a transition state mimic, aromatic groups for hydrophobic interactions with the S1/S2 pockets, and an amide backbone for hydrogen bonding. The structure is based on successful HIV protease inhibitor scaffolds but modified to be potentially novel while maintaining drug-like properties.", "smiles": "CC(C)CC(NC(=O)C(Cc1ccccc1)NC(=O)OC(C)(C)C)C(O)Cc1ccccc1"

Scores: ”IC50”: 83278.73, ”QED”: 0.5006, ”Novelty”: Yes

Iteration 2: "reason": "Based on SHAP analysis, I need to: 1) Add c:n (aromatic nitrogen) which has highest positive attribution, 2) Remove the free hydroxyl group (OH) which has strong negative attribution, 3) Add halogen on aromatic ring for better IC50, 4) Reduce molecular weight and rotatable bonds for better QED. I'm designing a more compact molecule with a pyridine ring and chloro substituent.", "smiles": "Cc1ncc(Cl)cc1C(=O)NC(Cc2ccccc2)C(=O)NCc3ccccc3"

Scores: "IC50": 850.53, "QED": 0.6269, "Novelty": Yes

Iteration 3: "reason": "The previous molecule improved IC50 significantly. SHAP shows positive contributions from Cl, halogens on aromatic systems, and multiple C=O groups. Negative contributions come from missing *@*!@[#7] and [#7]!:*:* patterns. I'll add a nitrogen in the ring system connected to non-ring atoms, include a sulfonamide group for additional binding, keep the chloro substituent, and reduce aromatic rings to 2 for better QED.", "smiles": "Cc1nc(Cl)ccc1C(=O)N(C)Cc2ccc(S(=O)(=O)N)cc2"

Scores: ”IC50”: 17396.75, ”QED”: 0.8501, ”Novelty”: Yes

Iteration 4: "reason": "The sulfonamide triggered a structural alert and IC50 worsened. Going back to the better scaffold from iteration 2 (IC50=850nM). SHAP shows I need: 1) *@*!@[#7] pattern (nitrogen attached to ring but not in ring), 2) [#7]!:*:* pattern, 3) Keep Cl and multiple C=O. I'll add a piperidine ring which provides *!@[#7]@* pattern, keep the chloropyridine, and add a second amide for better binding.", "smiles": "Clc1ccc(C(=O)N2CCCCC2)nc1C(=O)NCc1ccccc1"

Scores: ”IC50”: 2835.83, ”QED”: 0.9128, ”Novelty”: Yes

Iteration 5: "reason": "Looking at SHAP analysis, I need to add: 1) *@*!@[#7] pattern (ring connected to non-ring N), 2) [#7]!:*:* pattern (N not in aromatic ring connected to aromatic), 3) [#7]~*~*~*~[#8] pattern, 4) Keep Cl and multiple C=O which help. I'll modify to include a secondary amine connecting to the aromatic system and add an ether oxygen at proper distance from nitrogen.", "smiles": "Clc1ccc(C(=O)NCc2ccccc2OC)nc1C(=O)NC"

Scores: ”IC50”: 3282.32, ”QED”: 0.8771, ”Novelty”: Yes

Iteration 6: "reason": "Going back to iteration 2 scaffold (IC50=850nM) which was best. SHAP shows I need: 1) Add pattern [#7]!:*:* (non-aromatic N connected to aromatic), 2) Add *@*!@[#7] (ring connected to external N), 3) Add longer CH2 chains for pattern 128, 4) Keep Cl and dual C=O. I'll add a morpholine ring (provides *!@[#7]@* and heteroatom ring) and extend the benzyl with additional CH2.", "smiles": "Clc1ccc(C(=O)N2CCOCC2)nc1C(=O)NCCc1ccccc1"

Scores: "IC50": 2745.12, "QED": 0.8716, "Novelty": Yes

Iteration 7: "reason": "Best IC50 was iteration 2 (850nM). Key missing patterns with high negative attribution: [#7]!:*:* (-0.127), *@*!@[#7] (-0.110), and pattern 91 (-0.105). I need a non-aromatic nitrogen connected to aromatic carbons. I'll use a secondary amine (NH) linking two aromatic systems, keep Cl and dual carbonyls. Adding a hydroxyl group on the chain to satisfy pattern 91 (heteroatom-H near CH2 chain).", "smiles": "Clc1ccc(C(=O)Nc2ccc(F)cc2)nc1C(=O)NCC(O)c1ccccc1"

Scores: "IC50": 3324.45, "QED": 0.5756, "Novelty": Yes

Iteration 8: "reason": "The hydroxyl group (O;!H0) has very negative attribution (-0.327). I achieved positive contributions from [#7]!:*:* and *@*!@[#7] patterns. Now I need to: 1) Remove the OH group, 2) Add pattern 91 and 128 (CH2 chains near heteroatoms), 3) Add *!@[#7]@* pattern (non-ring bond to N in ring), 4) Keep Cl, F, and dual carbonyls. I'll use a piperazine ring which provides the needed N patterns.", "smiles": "Clc1ccc(C(=O)N2CCN(C(=O)c3ccc(F)cc3)CC2)nc1C"

Scores: "IC50": 3168.52, "QED": 0.8261, "Novelty": Yes

Iteration 9: "reason": "Key missing patterns with negative attribution: [#7]!:*:* (-0.138), *@*!@[#7] (-0.129), *!@[CH2]!@* (-0.141). I need non-ring CH2 groups and a nitrogen connected to aromatic but not in aromatic ring. Best IC50 was iteration 2 (850nM) with benzyl amide. I'll combine: keep Cl and F halogens, add a secondary amine linking to phenyl (for [#7]!:*:*), add non-ring CH2 linker, and include an NH for HBD (QED needs HBD=1).", "smiles": "Clc1ccc(C(=O)NCc2ccc(F)cc2)nc1CNc1ccccc1"

Scores: "IC50": 3637.08, "QED": 0.6760, "Novelty": Yes

Iteration 10: "reason": "Key insight: Missing [#8]=* count>1 has large negative attribution (-0.163). Need multiple carbonyls. Also missing patterns 91 and 128 (CH2 chains near heteroatoms). Best was iteration 2 (850nM). I'll add: dual carbonyls, keep Cl and F, add CH2 chain pattern with heteroatom nearby, reduce to 2 aromatic rings for QED, and include pattern 91 by having NH near CH2 chain.", "smiles": "Clc1ccc(C(=O)NCC(=O)Nc2ccc(F)cc2)nc1C"

Scores: "IC50": 1727.36, "QED": 0.9088, "Novelty": Yes

Iteration 11: "reason": "IC50 improved to 1727nM. Key missing patterns: pattern 128 (-0.176, need CH2~*~*~*~CH2), pattern 91 (-0.109, need heteroatom-H near CH2 chain), *!@[#7]@* (-0.119, need non-ring bond to ring N). I'll add a piperidine ring (provides *!@[#7]@*), extend with CH2 chains to satisfy pattern 128, keep Cl, F, and dual carbonyls which all help.", "smiles": "Clc1ccc(C(=O)N2CCCCC2)nc1C(=O)NCCc1ccc(F)cc1"

Scores: "IC50": 1873.99, "QED": 0.8512, "Novelty": Yes

Iteration 12: "reason": "Best IC50 was iteration 10 (1727nM). Key missing patterns: [#7]!:*:* (-0.135), *@*!@[#7] (-0.115), pattern 91 (-0.118). I need a non-aromatic nitrogen connected to aromatic carbon. I'll use an anilide (NHPh) which provides [#7]!:*:* and *@*!@[#7]. Keep Cl, F, dual carbonyls. Add pattern 91 by having NH group near CH2 chain.", "smiles": "Clc1ccc(C(=O)Nc2ccc(F)cc2)nc1C(=O)NCCC1CCCCC1"

Scores: "IC50": 348.31, "QED": 0.7254, "Novelty": Yes

Iteration 13: "reason": "Excellent improvement to 348nM! Key positive features: pattern 91 (+0.169), Cl (+0.235), F (+0.147), [#7]!:*:* (+0.120), *@*!@[#7] (+0.107). Missing patterns with negative attribution: *!@[#7]@* (-0.118, need non-ring bond to N in ring), [#7]~*~*~*~[#8] (-0.108). I'll add a morpholine or piperazine to provide *!@[#7]@* pattern, keep the cyclohexyl-CH2 chain for pattern 91, and maintain Cl, F, dual carbonyls.", "smiles": "Clc1ccc(C(=O)Nc2ccc(F)cc2)nc1C(=O)N1CCCCC1CCC1CCCCC1"

Scores: "IC50": 563.91, "QED": 0.5239, "Novelty": Yes

Iteration 14: "reason": "QED failed due to high MW and ALOGP. Going back to iteration 12 scaffold (348nM, QED 0.73) which was best. Key: keep pattern 91 (NH near CH2 chain to cyclohexyl), Cl, F, dual carbonyls, [#7]!:*:*, *@*!@[#7]. Need to add [#7]~*~*~*~[#8] pattern and reduce lipophilicity. I'll replace cyclohexyl with a smaller cyclopropyl and add an oxygen in the chain.", "smiles": "Clc1ccc(C(=O)Nc2ccc(F)cc2)nc1C(=O)NCCOC1CC1"

Scores: "IC50": 586.18, "QED": 0.7267, "Novelty": Yes

Iteration 15: "reason": "Going back to iteration 12 (348nM, QED 0.73) which was best. Key positive features: pattern 91 (NH~*~*~*~CH2), Cl, F, [#7]!:*:*, *@*!@[#7], dual C=O. Missing: *!@[#7]@* (-0.118), [#7]~*~*~*~[#8] (-0.108). I'll keep the cyclohexylpropyl chain for pattern 91, but add a morpholine oxygen to satisfy [#7]~*~*~*~[#8] pattern while avoiding structural alerts.", "smiles": "Clc1ccc(C(=O)Nc2ccc(F)cc2)nc1C(=O)NCCC1CCCC1"

Scores: "IC50": 284.35, "QED": 0.7669, "Novelty": Yes

Iteration 16: "reason": "Best IC50 now 284nM with cyclopentyl chain. Key positive: pattern 91 (+0.169), Cl (+0.227), F (+0.151), [#7]!:*:* (+0.116). Key missing: *!@[#7]@* (-0.118), [#7]~*~*~*~[#8] (-0.109). I need to add a nitrogen in a ring connected via non-ring bond. I'll replace the cyclopentyl with a piperidine to provide *!@[#7]@* pattern while keeping the beneficial chain length for pattern 91.", "smiles": "Clc1ccc(C(=O)Nc2ccc(F)cc2)nc1C(=O)NCCN1CCCCC1"

Scores: "IC50": 94.12, "QED": 0.7739, "Novelty": Yes

Iteration 17: "reason": "Excellent progress to 94nM! Key positive features: pattern 91 (+0.204), Cl (+0.311), F (+0.145), pattern 82 (+0.138), [#7]!:*:* (+0.129). Main missing pattern: [#7]~*~*~*~[#8] (-0.119). I need to add an oxygen 3 bonds away from nitrogen. I'll add a morpholine instead of piperidine to provide this pattern while keeping all other beneficial features.", "smiles": "Clc1ccc(C(=O)Nc2ccc(F)cc2)nc1C(=O)NCCN1CCOCC1"

Scores: "IC50": 109.56, "QED": 0.7679, "Novelty": Yes

Iteration 18: "reason": "Piperidine version (94nM) was better than morpholine (110nM). The [#7]~*~*~*~[#8] pattern is still missing (-0.121). I'll try adding a sulfur atom which has positive attribution when present as [#16]~*(~*)~*. Also, I'll add a methyl group to satisfy [C;H3,H4] count>1 pattern. Keep the piperidine chain that gave best results.", "smiles": "Clc1ccc(C(=O)Nc2ccc(F)c(C)c2)nc1C(=O)NCCN1CCCCC1"

Scores: "IC50": 89.87, "QED": 0.7505, "Novelty": Yes

Iteration 19: "reason": "Improved to 90nM. Key missing pattern: [#7]~*~*~*~[#8] (-0.117). I need N and O separated by 3 atoms. The carbonyl oxygen is there but not matching the pattern correctly. I'll try adding an ether oxygen in the chain between the amide and piperidine, positioned 3 bonds from the piperidine nitrogen. Also try adding a second Cl for stronger halogen effect.", "smiles": "Clc1ccc(C(=O)Nc2ccc(Cl)cc2)nc1C(=O)NCCN1CCCCC1"

Scores: "IC50": 158.42, "QED": 0.7432, "Novelty": Yes

Iteration 20: "reason": ""F was better than second Cl (90nM vs 158nM). Missing F has negative attribution (-0.035). Going back to best scaffold with F. Key missing: [#7]~*~*~*~[#8] (-0.123). I need to create this pattern - N connected to O through 3 atoms. I'll add an ester or ether group positioned correctly. Try adding a methoxy group on the fluorophenyl ring to create the N~*~*~*~O pattern from the anilide NH.", "smiles": "Clc1ccc(C(=O)Nc2ccc(F)c(OC)c2)nc1C(=O)NCCN1CCCCC1"

Scores: "IC50": 36.86, "QED": 0.6978, "Novelty": Yes

Iteration 21: "reason": "Excellent progress to 37nM! The methoxy group helped significantly. Key positive features all present. Missing patterns with negative attribution: [C;H3,H4] count>1 (-0.046), pattern 116 CH3~*~*~CH2 (-0.072). I'll add a second methyl group to satisfy these patterns. Also try adding pattern 90 by adjusting chain length. Keep all beneficial features: Cl, F, OMe, piperidine chain.", "smiles": "Clc1ccc(C(=O)Nc2ccc(F)c(OC)c2)nc1C(=O)NCC(C)N1CCCCC1"

Scores: "IC50": 5.52, "QED": 0.6738, "Novelty": Yes

Iteration 22: "reason": "Excellent progress to 5.5nM! Very close to target. Key positive features: Cl (+0.342), pattern 91 (+0.233), CH3 count>1 (+0.217), pattern 116 (+0.204). Negative: [!#6;!#1]~[CH3] (-0.060), [#7;!H0] (-0.060), pattern 90 (-0.042). I'll try to add pattern 90 (heteroatom-H near shorter CH2 chain) and reduce the NH groups. Replace one NH with N-methyl to reduce [#7;!H0] penalty.", "smiles": "Clc1ccc(C(=O)N(C)c2ccc(F)c(OC)c2)nc1C(=O)NCC(C)N1CCCCC1"

Scores: "IC50": 7.51, "QED": 0.6777, "Novelty": Yes

Iteration 23: "reason": "The NH version (5.5nM) was better than N-methyl (7.5nM). Going back to that scaffold. Key negative patterns: [!#6;!#1]~[CH3] (-0.067), pattern 90 (-0.042), [#7;!H0] (-0.060). I'll try adding pattern 90 by having NH closer to CH2 chain. Also try adding a sulfur atom which has positive attribution when present. Replace methoxy with methylthio (SCH3) to add sulfur pattern.", "smiles": "Clc1ccc(C(=O)Nc2ccc(F)c(SC)c2)nc1C(=O)NCC(C)N1CCCCC1"

Scores: "IC50": 11.18, "QED": 0.5919, "Novelty": Yes

Iteration 24: "reason": "Sulfur triggered structural alert and QED failed. Going back to best molecule (5.5nM, iteration 21). Key issue: [!#6;!#1]~[CH3] has negative attribution (-0.137). The methoxy group contributes to this. I'll try replacing OMe with a fluorine to avoid this penalty while keeping the beneficial halogen effects. Also try adding pattern 90 by adjusting chain structure.", "smiles": "Clc1ccc(C(=O)Nc2cc(F)c(F)cc2)nc1C(=O)NCC(C)N1CCCCC1"

Scores: "IC50": 22.95, "QED": 0.7205, "Novelty": Yes

Iteration 25: "reason": "The OMe version (5.5nM) was much better than difluoro (23nM). Going back to best scaffold. Key missing: [C;H3,H4] count>1 (-0.038) and [#7]~*~*~*~[#8] (-0.104). The methyl on the chain helps with CH3 count. I need to add the N~*~*~*~O pattern. The OMe provides oxygen but need proper distance from N. I'll try adding an ethyl ester or adjusting the chain to create this pattern.", "smiles": "Clc1ccc(C(=O)Nc2ccc(F)c(OC)c2)nc1C(=O)N(C)CC(C)N1CCCCC1"

Scores: "IC50": 36.19, "QED": 0.6706, "Novelty": Yes

Iteration 26: "reason": "The N-methyl amide version (36nM) is worse than NH version (5.5nM). Key difference: pattern 91 is now missing (-0.099) and pattern 82 is missing (-0.040). The NH is crucial for pattern 91 (heteroatom-H near CH2 chain). Going back to best scaffold with NH. I'll try to add pattern 90 by having the NH closer to a shorter CH2 chain while keeping all beneficial features.", "smiles": "Clc1ccc(C(=O)Nc2ccc(F)c(OC)c2)nc1C(=O)NCCC(C)N1CCCCC1"

Scores: "IC50": 21.85, "QED": 0.6159, "Novelty": Yes

Iteration 27: "reason": "Pattern 91 is now missing (-0.137) which is critical. Best molecule (5.5nM) had NCC(C)N chain which satisfies pattern 91 (NH~*~*~*~CH2 in piperidine). The NCCC chain breaks this. Going back to exact best scaffold but trying to add pattern 90 (shorter chain) by modifying the piperidine to have NH closer to CH2. Try a smaller ring or different position.", "smiles": "Clc1ccc(C(=O)Nc2ccc(F)c(OC)c2)nc1C(=O)NCC(C)N1CCCC1"

Scores: "IC50": 6.85, "QED": 0.6978, "Novelty": Yes

Iteration 28: "reason": "Pyrrolidine (6.85nM) is close to piperidine (5.5nM). Key negative patterns: [!#6;!#1]1~*~*~*~*~1 (-0.069) for 5-membered heteroatom ring, pattern 90 (-0.042). The 6-membered piperidine was slightly better. Let me try adding pattern 90 by having a secondary amine (NH) closer to a CH2 chain. I'll try an azetidine (4-membered ring) which might satisfy pattern 90 better.", "smiles": "Clc1ccc(C(=O)Nc2ccc(F)c(OC)c2)nc1C(=O)NCC(C)N1CCC1"

Scores: "IC50": 7.98, "QED": 0.7191, "Novelty": Yes

Iteration 29: "reason": "Azetidine (4-membered ring) has large negative attribution for [!#6;!#1]1~*~*~*~1 (-0.178). Piperidine (6-membered, 5.5nM) was best. Key insight: need to satisfy pattern 90 (heteroatom-H near shorter CH2 chain). I'll try adding a hydroxyl group on the chain which provides NH pattern 90 without the ring penalty, while keeping the piperidine.", "smiles": "Clc1ccc(C(=O)Nc2ccc(F)c(OC)c2)nc1C(=O)NC(CO)CN1CCCCC1"

Scores: "IC50": 44.13, "QED": 0.5547, "Novelty": Yes

Iteration 30: "reason": "Hydroxyl group has very negative attribution [O;!H0] (-0.297) and QED failed. Going back to best scaffold (5.5nM, iteration 21). Key features: Cl, F, OMe, NCC(C)N-piperidine chain, dual carbonyls. Missing patterns 90 and 116 have negative attribution. I'll try adding a second methyl group to the chain to better satisfy pattern 116 (CH3~*~*~CH2) without adding problematic groups.", "smiles": "Clc1ccc(C(=O)Nc2ccc(F)c(OC)c2)nc1C(=O)NC(C)C(C)N1CCCCC1"

Scores: "IC50": 12.95, "QED": 0.6461, "Novelty": Yes

Iteration 31: "reason": "The dimethyl chain (13nM) is worse than single methyl (5.5nM). Key issue: *!@[CH2]!@* pattern is now missing (-0.124) and pattern 82 is missing (-0.054). The original NCC(C)N chain had a non-ring CH2 next to NH. Going back to best scaffold. I'll try adding a bromine instead of fluorine for stronger halogen effect, as Cl has highest positive attribution.", "smiles": "Clc1ccc(C(=O)Nc2ccc(Br)c(OC)c2)nc1C(=O)NCC(C)N1CCCCC1"

Scores: "IC50": 4.92, "QED": 0.5793, "Novelty": Yes

Iteration 32: "reason": "Bromine improved IC50 to 4.92nM but QED failed due to high MW (509). Need to reduce MW while keeping beneficial features. I'll remove the methoxy group (saves 31 Da) and keep Br and Cl. Also try to satisfy pattern 90 by adjusting chain. The F was contributing negatively when absent (-0.036), so I'll add F back instead of OMe.", "smiles": "Clc1ccc(C(=O)Nc2ccc(Br)cc2F)nc1C(=O)NCC(C)N1CCCCC1"

Scores: "IC50": 11.07, "QED": 0.6165, "Novelty": Yes

Iteration 33: "reason": "Best IC50 was 4.92nM with Br but QED failed due to MW. The F+OMe version (5.5nM) had good QED. Key missing: [C;H3,H4] count>1 (-0.038). I need to add a second methyl while keeping MW low. I'll use the F+OMe scaffold but add a methyl to the piperidine ring instead of the chain to satisfy CH3 count without adding too much MW.", "smiles": "Clc1ccc(C(=O)Nc2ccc(F)c(OC)c2)nc1C(=O)NCC(C)N1CCCCC1C"

Scores: "IC50": 3.87, "QED": 0.6461, "Novelty": Yes

Iteration 34: "reason": "Excellent progress to 3.87nM! The methylpiperidine helped. Key negative patterns: pattern 90 (-0.042), [#7;!H0] (-0.058), [!#6;!#1]~[CH3] (-0.059). I'll try to satisfy pattern 90 by having NH closer to a shorter CH2 chain. Also try a dimethylpiperidine to further boost CH3 count while potentially satisfying pattern 90.", "smiles": "Clc1ccc(C(=O)Nc2ccc(F)c(OC)c2)nc1C(=O)NCC(C)N1CCC(C)(C)CC1"

Scores: "IC50": 4.37, "QED": 0.6194, "Novelty": Yes

Iteration 35: "reason": "The 2-methylpiperidine (3.87nM) was better than 4,4-dimethylpiperidine (4.37nM). Key negative: [!#6;!#1]~[CH3] (-0.073), [#7;!H0] (-0.076). The OMe contributes to [!#6;!#1]~[CH3]. I'll try replacing OMe with ethyl (Et) to remove this penalty while keeping lipophilicity. Keep the 2-methylpiperidine that worked well.", "smiles": "Clc1ccc(C(=O)Nc2ccc(F)c(CC)c2)nc1C(=O)NCC(C)N1CCCCC1C"

Scores: "IC50": 4.48, "QED": 0.6286, "Novelty": Yes

Iteration 36: "reason": "OMe version (3.87nM) was better than Et version (4.48nM). Key missing pattern: [#7]~*~*~*~[#8] (-0.087). Need N and O separated by 3 atoms. The OMe provides oxygen but need proper connectivity. I'll try adding an ether oxygen in the chain between amide and piperidine to create this pattern: N-C-C-O-C-N(piperidine).", "smiles": "Clc1ccc(C(=O)Nc2ccc(F)c(OC)c2)nc1C(=O)NCCOC(C)N1CCCCC1"

Scores: "IC50": 33.95, "QED": 0.5336, "Novelty": Yes

Iteration 37: "reason": "The ether chain triggered structural alert and broke pattern 91 (-0.130). Going back to best scaffold (3.87nM with 2-methylpiperidine). Key insight: need to reduce [#7;!H0] penalty and [!#6;!#1]~[CH3] penalty. I'll try N-methylating the anilide NH to remove one NH while keeping the chain NH for pattern 91.", "smiles": "Clc1ccc(C(=O)N(C)c2ccc(F)c(OC)c2)nc1C(=O)NCC(C)N1CCCCC1C"

Scores: "IC50": 6.34, "QED": 0.6493, "Novelty": Yes

Iteration 38: "reason": "N-methyl anilide (6.34nM) is worse than NH anilide (3.87nM). The NH version has better pattern 91 contribution. Key negative patterns: [!#6;!#1]~[CH3] (-0.057), pattern 90 (-0.042). The OMe contributes to [!#6;!#1]~[CH3]. I'll try replacing OMe with a cyclopropyl group which adds lipophilicity without the heteroatom-CH3 penalty.", "smiles": "Clc1ccc(C(=O)Nc2ccc(F)c(C3CC3)c2)nc1C(=O)NCC(C)N1CCCCC1C"

Scores: "IC50": 7.00, "QED": 0.5945, "Novelty": Yes

Iteration 39: "reason": "Cyclopropyl (7nM) is worse than OMe (3.87nM) and QED failed due to high ALOGP. Going back to best scaffold with OMe. Key missing: [#7]~*~*~*~[#8] (-0.094). The OMe oxygen is there but not at right distance from N. I'll try a smaller piperidine substituent to reduce MW while keeping the beneficial features. Try 3-methylpiperidine instead of 2-methyl.", "smiles": "Clc1ccc(C(=O)Nc2ccc(F)c(OC)c2)nc1C(=O)NCC(C)N1CCCC(C)C1"

Scores: "IC50": 4.59, "QED": 0.6498, "Novelty": Yes

Iteration 40: "reason": "3-methylpiperidine (4.59nM) is slightly worse than 2-methylpiperidine (3.87nM). The 2-methyl position was optimal. Key negative: [!#6;!#1]~[CH3] (-0.069) from OMe. I'll try replacing OMe with F to remove this penalty while keeping halogen benefits. This also reduces MW for better QED.", "smiles": "Clc1ccc(C(=O)Nc2ccc(F)c(F)c2)nc1C(=O)NCC(C)N1CCCCC1C"

Scores: "IC50": 10.63, "QED": 0.6893, "Novelty": Yes

Iteration 41: "reason": "Difluoro (10.6nM) is worse than F+OMe (3.87nM). The OMe is beneficial despite [!#6;!#1]~[CH3] penalty. Key missing: [#7]~*~*~*~[#8] (-0.099). I need to create this pattern. The amide carbonyl O is 2 bonds from N. I'll try adding a hydroxymethyl group on the piperidine to create N~C~C~O pattern without adding OH directly to chain.", "smiles": "Clc1ccc(C(=O)Nc2ccc(F)c(OC)c2)nc1C(=O)NCC(C)N1CCCC(CO)C1"

Scores: "IC50": 6.93, "QED": 0.5387, "Novelty": Yes

Iteration 42: "reason": "Hydroxyl group has very negative attribution [O;!H0] (-0.288) and QED failed. Going back to best scaffold (3.87nM). The key is to avoid OH groups. I'll try adding a trifluoromethyl group instead of OMe to remove [!#6;!#1]~[CH3] penalty while adding more halogen character.", "smiles": "Clc1ccc(C(=O)Nc2ccc(F)c(C(F)(F)F)c2)nc1C(=O)NCC(C)N1CCCCC1C"

Scores: "IC50": 13.40, "QED": 0.5387, "Novelty": Yes

Iteration 43: "reason": "CF3 triggered structural alert (F.F.F.F) and QED failed. Going back to best scaffold with OMe (3.87nM). The OMe version works well. I need to reduce MW while keeping beneficial features. I'll try removing the methyl from piperidine (saves 14 Da) and use unsubstituted piperidine, which gave 5.5nM before. Then try adding a second Cl instead of F for stronger halogen effect.", "smiles": "Clc1ccc(C(=O)Nc2ccc(Cl)c(OC)c2)nc1C(=O)NCC(C)N1CCCCC1"

Scores: "IC50": 9.86, "QED": 0.6389, "Novelty": Yes

Iteration 44: "reason": "DiCl (9.86nM) is worse than Cl+F (3.87nM). Missing F has negative attribution (-0.038). Going back to best scaffold with F+OMe and 2-methylpiperidine. I'll try adding a cyano group which might help without triggering alerts, as it adds polarity and potential H-bond acceptor.", "smiles": "Clc1ccc(C(=O)Nc2ccc(F)c(OC)c2)nc1C(=O)NCC(C)N1CCCCC1CC#N"

Scores: "IC50": 2.74, "QED": 0.5820, "Novelty": Yes

Iteration 45: "reason": "Cyano improved IC50 to 2.74nM but QED failed due to high MW and PSA. Need to reduce MW while keeping beneficial features. I'll go back to the best QED-passing scaffold (3.87nM) and try a smaller modification. Adding a methyl to the pyridine ring instead of piperidine to boost CH3 count without adding much MW.", "smiles": "Clc1ccc(C(=O)Nc2ccc(F)c(OC)c2)nc1C(=O)NCC(C)N1CCCCC1"

Scores: "IC50": 5.52, "QED": 0.6738, "Novelty": Yes

Iteration 46: "reason": "This is the same as iteration 21 (5.52nM). Best was 3.87nM with 2-methylpiperidine. I need to try something new. Key insight: pattern 90 is missing (-0.042). I'll try a 3,3-dimethylpiperidine which might satisfy pattern 90 better by having CH2 groups closer to the ring nitrogen, while keeping the beneficial CH3 count.", "smiles": "Clc1ccc(C(=O)Nc2ccc(F)c(OC)c2)nc1C(=O)NCC(C)N1CCC(C)(C)C1"

Scores: "IC50": 5.02, "QED": 0.6498, "Novelty": Yes

Iteration 47: "reason": "The 5-membered ring pyrrolidine with gem-dimethyl (5.02nM) is close to 2-methylpiperidine (3.87nM). The [!#6;!#1]1~*~*~*~*~1 pattern has negative attribution (-0.067) for 5-membered heteroatom rings. Best was 2-methylpiperidine (6-membered). I'll try adding a methyl on the pyridine core to boost CH3 count while keeping the 2-methylpiperidine.", "smiles": "Cc1cc(Cl)ccc1C(=O)Nc2ccc(F)c(OC)c2"

Scores: "IC50": 369.75, "QED": 0.9270, "Novelty": Yes

Iteration 48: "reason": "I accidentally removed the pyridine and piperidine chain which are critical. Missing patterns 91, 128, 136 have large negative attributions. Going back to best scaffold (3.87nM with 2-methylpiperidine). I'll try adding a methyl on the chloropyridine ring to boost CH3 count without breaking the beneficial chain.", "smiles": "Clc1ccc(C(=O)Nc2ccc(F)c(OC)c2)nc1C(=O)NCC(C)N1CCCC(C)C1"

Scores: "IC50": 4.59, "QED": 0.6498, "Novelty": Yes

Iteration 49: "reason": "This is the same as iteration 39 (4.59nM). Best was 3.87nM with 2-methylpiperidine (iteration 33). Key negative: [!#6;!#1]~[CH3] (-0.069) from OMe. I'll try replacing OMe with ethoxy (OEt) which might reduce this penalty while keeping the oxygen for beneficial effects, and use the 2-methylpiperidine.", "smiles": "Clc1ccc(C(=O)Nc2ccc(F)c(OCC)c2)nc1C(=O)NCC(C)N1CCCCC1C"

Scores: "IC50": 2.60, "QED": 0.5850, "Novelty": Yes

Iteration 50: "reason": "Ethoxy improved IC50 to 2.60nM but QED failed due to high MW and rotatable bonds. Need to reduce MW while keeping beneficial features. I'll go back to OMe (smaller than OEt) and try removing the methyl from piperidine to reduce MW, but add a methyl on the chain carbon adjacent to piperidine nitrogen for pattern 116.", "smiles": "Clc1ccc(C(=O)Nc2ccc(F)c(OC)c2)nc1C(=O)NC(C)CN1CCCCC1"

Scores: "IC50": 35.11, "QED": 0.6738, "Novelty": Yes

Finalisation:

## Molecular Optimization Summary for Protease Inhibitor Development

### Objective
Optimize a ligand to achieve IC50 < 1.00 nM against a protease target while maintaining drug-likeness (QED >= 0.60) and structural novelty.

### Starting Point
The optimization began with a peptidomimetic scaffold (iteration 1) featuring a hydroxyl transition-state mimic and aromatic groups for hydrophobic pocket interactions. This initial compound showed poor activity (IC50 = 83,278 nM) and suboptimal QED (0.50).

### Key Optimization Strategies

**Phase 1: Scaffold Redesign (Iterations 1-12)**
- Replaced the peptidomimetic with a chloropyridine-based scaffold containing dual amide groups
- Removed the hydroxyl group, which showed strong negative attribution for IC50
- Added aromatic nitrogen (pyridine) and halogen substituents (Cl, F)
- Introduced piperidine and morpholine rings to improve binding
- Progress: IC50 improved from 83,278 nM to 348 nM

**Phase 2: Chain Optimization (Iterations 12-21)**
- Identified that the NCC(C)N-piperidine chain satisfied critical pattern 91 (NH near CH2 chain)
- Added methoxy group on the fluorophenyl ring, dramatically improving IC50 to 37 nM
- Introduced branched methyl group on the linker chain
- Progress: IC50 improved to 5.5 nM (iteration 21)

**Phase 3: Fine-Tuning (Iterations 21-50)**
- Explored piperidine substitution patterns; 2-methylpiperidine proved optimal
- Best result achieved: IC50 = 2.60 nM (iteration 49) with ethoxy substitution, though QED failed
- Best QED-compliant result: IC50 = 3.87 nM (iteration 33) with 2-methylpiperidine

### Critical Structure-Activity Relationships
- **Beneficial features**: Chlorine on pyridine, fluorine on phenyl, methoxy substituent, dual carbonyls, piperidine ring
- **Detrimental features**: Free hydroxyl groups, sulfonamides, trifluoromethyl groups, small heterocyclic rings (4-5 membered)

### Outcome
The optimization achieved a 21,500-fold improvement in potency (from 83,278 nM to 3.87 nM) while maintaining QED >= 0.60. The target of IC50 < 1.00 nM was not reached, though compounds approaching 2.6 nM were identified but failed QED requirements due to increased molecular weight.

\end{ChatLog}

\end{document}